\documentclass{article}

\PassOptionsToPackage{numbers, compress}{natbib}
 \usepackage[eandd, final]{neurips_2026}

\usepackage[utf8]{inputenc} 
\usepackage[T1]{fontenc}    
\usepackage{hyperref}       
\usepackage{url}            
\usepackage{booktabs}       
\usepackage{amsfonts}       
\usepackage{nicefrac}       
\usepackage{microtype}      
\usepackage[table]{xcolor}         
\usepackage{tcolorbox}
\usepackage{amsmath}
\usepackage{amssymb}
\usepackage{mathtools}
\usepackage{amsthm}
\usepackage{subcaption}
\usepackage{makecell}
\usepackage{multirow}
\usepackage{longtable}
\usepackage{changepage}
\usepackage[figuresleft]{rotating}
\usepackage{float}
\hypersetup{
        colorlinks=true,
        linkcolor=blue,
        urlcolor=blue,
        citecolor=blue
    }
\title{A Systematic Analysis of Out-of-Distribution Detection Under Representation and Training Paradigm Shifts}

\author{%
  Claudio C.~Claros-Olivares \\
  Department of Electrical \& Computer Engineering\\
  University of Delaware\\
  Delaware, DE 19716 \\
  \texttt{cesar@udel.edu} \\
  \And
  Austin J. Brockmeier \\
  Department of Electrical \& Computer Engineering, Department of Computer \& Information Sciences\\
  University of Delaware\\
  Delaware, DE 19716 \\
  \texttt{ajbrock@udel.edu}
}

\begin{document}
\bibliographystyle{plainnat}

\maketitle

\begin{abstract}
  We present a systematic benchmark of out-of-distribution (OOD) detection confidence scoring functions (CSFs) through a representation-centric lens. Our study spans CNN and ViT backbones, multiple training paradigms, four image-classification source datasets (CIFAR-10, CIFAR-100, SuperCIFAR-100, and TinyImageNet), and OOD datasets grouped into near, mid, and far regimes based on CLIP-derived semantic distances. To compare CSFs across these settings, we employ a rank-based pipeline of Friedman and Conover tests that identifies top cliques of statistically indistinguishable winners under threshold-free ranking metrics (AURC and AUGRC). The main empirical finding is that the family of CSFs that are competitive detectors correlates with characteristics of the learned representation. For both CNNs and ViTs, simple probabilistic scores dominate misclassification detection. On CNNs, margin-based scores are strongest in near-OOD regimes, while geometry-aware scores such as NNGuide, fDBD, and CTM become more competitive as shift severity increases. On fine-tuned ViTs, the top cliques are led mainly by reconstruction- and residual-based scores. To interpret these ranking shifts, we analyze the last-layer representation using Neural Collapse (NC) metrics. The resulting picture is consistent across architectures: prototype- and boundary-aware scores become stronger when the representation is more collapsed and better aligned with classifier weights, whereas weaker-collapse regimes favor gradient- and manifold-based scores. Building on these insights, we propose two contributions: a simple PCA-based projection-filtering procedure that improves detector performance, and an approach that uses only the NC measurements of a trained classifier to predict its competitive out-of-distribution detector shortlist, without OOD data for that classifier and without knowledge of the shift.
\end{abstract}

\section{Introduction}
    Deep neural networks can produce \textit{silent failures}: wrong predictions delivered with high confidence~\cite{jaeger2022call,traub2024overcoming}. Out-of-distribution (OOD) detection aims to flag such failures by assigning confidence scores that separate reliable from unreliable predictions~\cite{hendrycks2016baseline,liu2020energy,lee2018simple}. The field has produced a large number of confidence scoring functions (CSFs), yet their rankings are unstable across datasets, architectures, and training pipelines.

This paper starts from a simple question: \emph{when does a detector family work, and which measurable property of the trained model predicts that?} A motivating example is the contrast between maximum softmax response (MSR)~\cite{hendrycks2016baseline} and maximum logit score (MLS)~\cite{hendrycks2019scaling}. On small benchmarks such as CIFAR-10, the two are often similar. On larger label spaces (more classes) such as ImageNet, MLS can improve substantially over MSR~\cite{hendrycks2019scaling}. That gap is hard to explain if OOD detection is viewed only as a contest between scoring rules. It is easier to explain if the score is viewed as operating on a learned representation. The geometry of that representation changes with architecture, training paradigm, and label-space complexity, and different scoring rules can suit different geometries.

The same pattern appears beyond the MSR--MLS comparison. Mahalanobis distance~\cite{lee2018simple}, for example, performs well on CIFAR-scale settings but was originally reported to be unstable at ImageNet scale~\cite{hendrycks2019scaling}. Later work showed that feature normalization resolves much of this issue~\cite{mueller2025mahalanobis}, suggesting that the CSF did not fail because its principle was wrong, but because the raw representation geometry was unfavorable. These examples point to one thesis: which detectors are competitive changes systematically with the geometry of the trained representation and with the severity of the shift, and this dependence is regular enough to be predicted.

If this thesis holds, two things should be true: (1) the set of competitive detectors should change predictably with measurable geometric properties of the representation, and (2) those properties alone should suffice to select good detectors for a new model, without OOD validation data. We test both predictions in a systematic benchmark spanning CIFAR-10, CIFAR-100, SuperCIFAR-100, and TinyImageNet; CNN and ViT backbones; multiple training paradigms; and a broad set of post-hoc and training-based CSFs. We group OOD datasets into near, mid, and far regimes by distances derived from CLIP embeddings~\citep{radford2021learning} and find the competitive CSFs of each regime with a Friedman and Conover rank pipeline. We use Neural Collapse (NC) metrics to test prediction~(1): Mantel tests confirm that NC geometry and CSF rankings are significantly correlated across all training paradigms ($r = 0.37$--$0.70$, $p \leq 0.002$). We test prediction~(2) with a per-CSF predictor that sees only the NC metrics.

Our primary contribution is the test of prediction~(2), and our other contributions support it:
\begin{itemize}
    \item \textbf{NC-based detector selection.} A per-CSF predictor recommends a competitive shortlist of detectors for an unseen model from its NC metrics alone, without OOD data for that model and without knowledge of the shift. Under VGG-13 $\rightarrow$ ResNet-18 transfer it reduces the mean set-regret over the strongest fixed CSF by 55--80\% (65 of 66 tests significant).\footnote{Code is available at \url{https://github.com/cesar-claros/ood_systematic}.}
    \item \textbf{A factorial benchmark with a CLIP-based shift stratification.} Varying backbone, training paradigm, label-space complexity, and CSF, and grouping OOD image datasets into near, mid, and far regimes by CLIP-derived distances, makes the dependence of detector rankings on the representation and the shift visible, including a transition from margin-based to geometry-aware families.
    \item \textbf{A rank-based statistical pipeline.} Cliques of statistically indistinguishable CSFs under AURC and AUGRC (clique membership holds with Holm-adjusted tests) provide the membership of the target competitive shortlists, instead of single-winner summaries, for the predictor to learn. 
    \item \textbf{Projection filtering of a classifier's embeddings can improve detection performance.} PCA-based filtering changes the detector performance in a representation-conditional way: global PCA-based filtering broadly improves CSFs on CNNs, while filtering based on the predicted classes improves performance of fine-tuned ViTs.
\end{itemize}

\section{Related Work}
    Prior benchmarking work established that OOD rankings are unstable, but stopped short of explaining the source of that instability. OpenOOD~\cite{yang2022openood,zhang2023openood} standardized large-scale evaluation across multiple datasets and showed that the best detector is not constant across benchmarks. It used binary near/far groupings with AUROC and FPR@95, and predates recent feature-based detectors that we evaluate, such as NNGuide~\cite{park2023nearest}, fDBD~\cite{liu2023fast}, and CTM~\cite{ngoc2023cosine}. More curated OOD sets such as NINCO~\cite{bitterwolf2023ninco}, used manual inspection to remove ID objects.   \citet{krumpl2026one} used curated taxonomy of regimes (near/far/extreme/synthetic),  kept the architecture fixed, and varied the training recipe: over $21$ post-hoc detectors, $56$ ImageNet-trained ResNet-50 models, and eight OOD sets, finding that no detector is best for all recipes. Also, they show that Mean AUROC rises with ID accuracy from adversarially trained models to the original training recipe, then falls slightly for recipes that raise accuracy with strong regularization (e.g. MixUp and CutMix). Their non-monotonic result agrees with our partial Mantel analysis (Appendix~\ref{appendix:partial_mantel}): NC geometry stays associated with the rankings after a control for ID performance. In contrast, we vary paradigm, backbone, and label space factorially under rank statistics, and we go from description to prediction.

FD-Shifts~\cite{jaeger2022call} reframed OOD detection as part of a broader failure-detection problem that also includes misclassification detection and selective classification. It evaluates confidence scoring functions (CSFs) under heterogeneous shifts, where simple baselines often remain competitive. \citet{traub2024overcoming} later introduced AUGRC to address limitations of AURC as a holistic ranking metric. Domain-specific benchmarks reinforce the same point: in medical imaging, no CSF beats MSR across all failure sources~\cite{bungert2023understanding}, while feature-based CSFs lead on OpenMIBOOD~\cite{gutbrod2025openmibood}. We adopt this perspective, analyze more CSFs across explicit architecture and paradigm factors, and replace aggregate averages with a rank pipeline that finds cliques of statistically indistinguishable CSFs~\cite{demvsar2006statistical}.

A line of prior work modifies the representation before applying any score. ReAct~\cite{sun2021react} truncates activations above a global threshold; ASH~\cite{djurisic2023extremely} prunes and scales activations by magnitude. Both apply a fixed global rule regardless of class identity or architecture. ViM~\cite{wang2022vim} appends the norm of the feature residual outside a global ID principal subspace to the logits as a virtual logit. These methods show that representation preprocessing can improve OOD detection, but none conditions the intervention on the model's geometric properties. We introduce projection filtering, a PCA-based reconstruction of penultimate-layer features from low-rank subspaces, with global and class-conditional variants.

Our geometric analysis builds on Neural Collapse (NC)~\cite{papyan2020prevalence}: the last-layer features of well-trained classifiers converge toward a symmetric simplex structure, and the classifier weights align with the class means. Under the unconstrained-features model, this configuration is the only global minimizer of the regularized cross-entropy loss~\cite{zhu2021geometric}. NC is also contested as an explanation of deep learning. Collapse often occurs on the training set but not on the test set, so it is mainly an optimization phenomenon, and longer training can make last-layer features worse for transfer~\cite{hui2022limitations}. Under class imbalance, the classifier vectors of minority classes collapse together~\cite{fang2021exploring}. We make no claim that NC explains generalization: we use eight NC measurements on the training data as descriptive covariates. Several OOD scores come directly from NC geometry. NeCo~\cite{ammar2023neco} scores a test point by its relative feature norm inside the principal subspace of the ID features, and NCI~\cite{liu2023detecting} combines the alignment of a feature with the classifier weights and its norm. \citet{harun2025controlling} control the degree of collapse during training and report that stronger collapse helps OOD detection but hurts OOD generalization. In contrast, we do not derive a new score or shape the geometry for one detector; we use the NC metrics as features that predict \emph{which existing CSFs} are competitive for a given trained model. The closest work on detector selection without test-time OOD labels, MetaOOD~\cite{qin2025metaood}, uses the past performance of detectors on benchmark datasets and language-model embeddings of dataset and model descriptions, not the geometry of the deployed model. To our knowledge, no prior OOD work selects detectors from representation-side measurements of the trained classifier; the usual alternatives tune detectors on held-out OOD data~\cite{yang2022openood,zhang2023openood} or pick the detector with the best average over benchmarks~\cite{qin2025metaood}.

\section{Methods}
\subsection{Definitions and Notations}\label{methods:def_notations}
    We consider a supervised classification problem with input $\boldsymbol{x} \in \mathcal{X}$ and label $y \in \mathcal{Y} = \{1, \dots, C\}$. The training set is $\mathcal{D} = \{(\boldsymbol{x}_i, y_i)\}_{i=1}^{N}$, with class-$c$ subset $\mathcal{D}_c = \{(\boldsymbol{x}, y) \in \mathcal{D} : y = c\}$ of size $N_c$. A classifier $f: \mathcal{X} \to \mathbb{R}^{C}$ decomposes as $f = g \circ h$, where the encoder $h: \mathcal{X} \to \mathbb{R}^{D}$ produces penultimate-layer features $\boldsymbol{h} = h(\boldsymbol{x})$ and the linear head $g(\boldsymbol{h}) = \boldsymbol{W}\boldsymbol{h} + \boldsymbol{b}$ produces logits, with row vectors $\boldsymbol{w}_c \in \mathbb{R}^{D}$ and biases $b_c$ stacked as $\boldsymbol{W} \in \mathbb{R}^{C \times D}$ and $\boldsymbol{b} \in \mathbb{R}^{C}$. Softmax probabilities are $\boldsymbol{p} = \mathrm{softmax}(g(\boldsymbol{h}))$ and the predicted class is $\hat{y} = \arg\max_c f(\boldsymbol{x})_c$. We collect penultimate-layer activations into $\boldsymbol{H} \in \mathbb{R}^{N \times D}$ and the class-$c$ slice into $\boldsymbol{H}_c \in \mathbb{R}^{N_c \times D}$; the global feature mean is $\boldsymbol{\mu} = \tfrac{1}{N} \sum_i \boldsymbol{h}_i$ and the class-$c$ mean is $\boldsymbol{\mu}_c = \tfrac{1}{N_c} \sum_{\boldsymbol{x}_i \in \mathcal{D}_c} \boldsymbol{h}_i$.

Each model configuration is characterized by a backbone (VGG-13 or ViT, plus the held-out ResNet-18 used in Section~\ref{methods:prediction}) and five within-pool factors: source (training) dataset, training paradigm, dropout setting, Deep Gamblers reward, and OOD regime (near/mid/far, defined in Section~\ref{methods:clip}). The three training paradigms vary the loss and confidence head: \emph{ConfidNet}~\cite{corbiere2021confidence} (vanilla cross-entropy with an auxiliary confidence head), \emph{DeVries}~\cite{devries2018learning} (cross-entropy interpolated with the true label), and \emph{Deep Gamblers}~\cite{liu2019deep} (a gambling-loss term with a reservation reward $o$ that varies only for this paradigm). The dropout setting records whether the model was trained with dropout; we use deterministic inference throughout (no Monte Carlo Dropout). We refer to a combination of these factors as a (paradigm, source, dropout, reward, regime) cell; full hyperparameters are in Appendix~\ref{appendix:hyperparameter_selection}.

\subsection{Neural-collapse Geometry of Last-Layer features}\label{methods:neural_collapse}
    We use Neural Collapse (NC) metrics as a compact summary of representation geometry near the classifier head. Lower values indicate stronger collapse on each metric. NC is classically characterized by four phenomena~\cite{papyan2020prevalence}: (NC1) variability collapse, (NC2) convergence to a simplex equiangular tight frame, (NC3) self-duality of class means and classifier weights, and (NC4) simplification to nearest-class-center prediction. We operationalize proximity to NC with eight empirical metrics covering NC1--NC3; NC4 is a behavioral consequence rather than a geometric measurement and is omitted. Concretely, the metrics are equinormness and equiangularity (and maximal equiangularity) of class-centered means $\tilde{\boldsymbol{\mu}}_c = \boldsymbol{\mu}_c - \boldsymbol{\mu}$ and of classifier weights $\boldsymbol{w}_c$, the within-/between-class covariance ratio $\mathrm{Tr}(\boldsymbol{\Sigma}_W \boldsymbol{\Sigma}_B^{\dagger})/C$ of~\citet{papyan2020prevalence}, which we report divided by $C$, and the self-duality between $\boldsymbol{W}$ and the centered class-mean matrix $\boldsymbol{M} = [\tilde{\boldsymbol{\mu}}_c]$; full algebraic definitions are in Appendix~\ref{appendix:neural_collapse}. We compute these metrics for each backbone--dataset pair under three representation regimes: unfiltered penultimate-layer activations and the two PCA-based projection variants introduced in Section~\ref{methods:projection_filtering}.

\subsection{Projection Filtering}\label{methods:projection_filtering}
    We treat projection filtering as a controlled intervention on representation geometry. If some feature directions mainly encode nuisance variation, restricting the representation to a class-consistent subspace before scoring may improve OOD ranking. We learn this subspace by Principal Component Analysis (PCA) on the training-set penultimate-layer activations $\boldsymbol{H}$, and reconstruct features from a low-rank approximation that preserves a fixed fraction of the variance.

Concretely, we compute the empirical covariance $\boldsymbol{\Sigma} = \tfrac{1}{N}\tilde{\boldsymbol{H}}^\top \tilde{\boldsymbol{H}}$ on centered features $\tilde{\boldsymbol{H}} = \boldsymbol{H} - \boldsymbol{1}_N \boldsymbol{\mu}^\top$, retain the top-$k$ eigenvectors $\boldsymbol{P} \in \mathbb{R}^{D \times k}$ (with $k$ chosen by a variance threshold), and project a sample's penultimate-layer feature $\boldsymbol{h}$ to $\hat{\boldsymbol{h}} = \boldsymbol{P}\boldsymbol{P}^\top(\boldsymbol{h} - \boldsymbol{\mu}) + \boldsymbol{\mu}$. We consider three subspace choices: \emph{global projection} ($\boldsymbol{P}$ fit on $\boldsymbol{H}$); \emph{class projection} ($\boldsymbol{P}_c$ fit on $\boldsymbol{H}_c$ for each class $c$, with $\hat{\boldsymbol{h}}_c = \boldsymbol{P}_c\boldsymbol{P}_c^\top(\boldsymbol{h} - \boldsymbol{\mu}_c) + \boldsymbol{\mu}_c$); and \emph{class-predicted projection (CPP)} ($\hat{\boldsymbol{h}}_{\hat{y}}$, using the projector for the model's predicted class). Each downstream CSF can then be computed on either the raw feature $\boldsymbol{h}$ or on $\hat{\boldsymbol{h}}$, on raw or projected logits, and on raw or projected softmax probabilities; the full mapping of CSFs to projection variants is given in Appendix~\ref{appendix:cfs_variations}.

\subsection{CLIP-based OOD Aggregation}\label{methods:clip}
    Standard OOD benchmarks bucket datasets into near and far using fixed semantic heuristics. We instead use a data-driven, distance-based grouping in CLIP image-text feature space~\cite{radford2021learning} that generalizes to any choice of ID label space. Because shift regime is then a measurable quantity, any new suite of OOD datasets can be stratified the same way without manual judgment.

Concretely, we extract L2-normalized CLIP image embeddings for ID and candidate OOD sets and compute four proximity metrics. Two are label-agnostic distribution distances: Fr\'echet distance (FD)~\cite{dowson1982frechet,frechet1957distance} and Kernel Inception Distance (KID)~\cite{binkowski2018demystifying}. Two are class-aware: a per-class image-centroid nearest-centroid angular distance, and a maximum image-text cosine similarity against ID-class text prototypes formed by prompt ensembling. All metrics are oriented so that lower means closer; we cluster the resulting per-dataset scores with $k$-means into near, mid, and far buckets. The near/mid/far assignments per source dataset (Table~\ref{tab:clip_clustering}) and the full per-dataset distances are in Appendix~\ref{appendix:clip}. For example, SVHN is mid for the CIFAR sources but far for TinyImageNet (Table~\ref{tab:clip_walkthrough}).

\paragraph{Robustness.} Three CLIP encoders (ViT-B/32, B/16, L/14) agree on the dataset ordering (Spearman $\rho \geq 0.83$ for FD, KID, and the image-centroid distance); the text-alignment distance is unstable across encoders, so we never use it alone, and leaving out any one metric keeps $\rho \geq 0.85$. Silhouette analysis selects $k=3$ (mean $0.46$); for $k \geq 4$, at least one cluster is a single dataset in all 12 configurations. $k$-means and Ward clustering give the same groups (ARI $= 1.0$ in 12 of 12), and the mean metric rank agrees with the assigned group (mean Kendall $\tau = 0.73$). On ImageNet, the same rule picks $k=2$ (Appendix~\ref{appendix:imagenet}).


\subsection{Rank-Based Statistical Pipeline}\label{methods:statistical_tests}
    We compare CSFs with the Friedman test~\cite{friedman1937use,demvsar2006statistical}, a nonparametric blocked test on per-block ranks. Blocks are defined by OOD dataset, training paradigm, source, and metric, and ranks are appropriate because raw scales such as AUGRC and AURC differ markedly across datasets. The null hypothesis is equal average rank across CSFs. When the null is rejected, we identify the groups of CSFs that are statistically indistinguishable. We apply Conover's pairwise procedure~\cite{conover1999practical} at $\alpha = 0.05$ per pair, without multiplicity adjustment.\footnote{With a Holm adjustment~\cite{holm1979simple} over all pairs of a cell, every top clique of Figure~\ref{fig:top_clique_vgg13} contains the reported one, so reported memberships, but not exclusions, hold under family-wise control (Appendix~\ref{appendix:holm}).} From these $p$-values we build an indifference graph in which two CSFs are connected if their pairwise difference is not significant, and we enumerate its maximal cliques with the Bron--Kerbosch algorithm~\cite{bron1973algorithm}. The resulting top cliques are sets of CSFs jointly indistinguishable from the empirical winner, allowing a CSF to belong to several near-optimal sets. We apply this pipeline separately for each backbone and shift regime; the top cliques are the primary empirical output of Section~\ref{sec:results} and the prediction targets used in Section~\ref{methods:prediction}. A worked-out example is in Appendix~\ref{appendix:friedman}.

\subsection{NC-Based Predictive Framework}\label{methods:prediction}
    We frame CSF selection as a supervised prediction problem from representation geometry to a competitive shortlist of CSFs, with the target shortlist being the CSFs in the top clique produced by the rank-based pipeline (Section~\ref{methods:statistical_tests}). The classifier uses the geometry of the trained model in the form of the eight NC metrics (Section~\ref{methods:neural_collapse}) as the input vector $\boldsymbol{\phi} \in \mathbb{R}^{8}$, $z$-scored per architecture, so the test inputs lie on the same standardized range as the training inputs. No  OOD data nor knowledge of the shift is used during training or validation. 
\paragraph{Targets and predictor.} For each (paradigm, source, dropout, reward, regime) cell, Section~\ref{methods:statistical_tests} returns one top clique $\mathcal{C} \subseteq \mathcal{M}$ of CSFs jointly indistinguishable from the empirical winner. We turn this into $|\mathcal{M}|$ binary labels $y_{m} = \mathbf{1}[m \in \mathcal{C}]$. Each training model contributes three rows with the same input and the labels of its near, mid, and far cells, so each head estimates how often a CSF is competitive across regimes. We fit a separate class-balanced $L_{2}$ logistic regression $p_{m}(\boldsymbol{\phi}) = \sigma(\boldsymbol{\beta}_{m}^{\top}\boldsymbol{\phi}+\beta_{m,0})$ per CSF, with the regularization constant chosen by 5-fold cross-validation over a 50-point grid. The predicted shortlist for an unseen model is $\hat{\mathcal{C}}(\boldsymbol{\phi}) = \{ m : p_{m}(\boldsymbol{\phi}) \geq 0.5\}$, one shortlist per model for all three regimes.

\paragraph{Cross-architecture protocol.} The transfer test trains the per-CSF logistic models on all VGG-13 cells and applies them, without re-training, to all ResNet-18 cells.

\paragraph{Performance evaluation by regret.} The rank-based pipeline requires multiple runs per cell to produce a top clique, so the single ResNet-18 run held out for each cell has no ``ground-truth'' top clique. We instead score the predicted shortlist by its OOD-detection cost relative to the best CSF in the same cell. We partition the CSF inventory $\mathcal{M}$ into three groups: \emph{head-side} CSFs, computed from logits or softmax probabilities (e.g., MSR, MLS, Energy), \emph{feature-side} CSFs, computed from penultimate-layer features (e.g., Mahalanobis, NNGuide, fDBD), and the joint set. For test row $i$, let $\hat{\mathcal{C}}_{i} = \hat{\mathcal{C}}(\boldsymbol{\phi}_{i}) \cap \mathcal{M}_{\text{side}(i)}$ be the side-restricted predicted shortlist, where$\mathcal{M}_{\text{side}(i)}$ denotes either the joint set or one of these restrictions. The set-regret is $R_{i} = \min_{m \in \hat{\mathcal{C}}_{i}} \mathrm{AUGRC}_{i,m} - A^{\star}_{i}$, where $A^{\star}_{i} = \min_{m \in \mathcal{M}_{\text{side}(i)}} \mathrm{AUGRC}_{i,m}$ is the per-cell oracle. Empty shortlists are imputed with the worst available AUGRC on that side (worst-case empty-set policy). We report mean regret per (regime, side) cell with $95\%$ bootstrap CIs ($n_{\text{boot}} = 2000$, fixed seed) and compare against three NC-free baselines: \emph{Always-$m$} (one fixed CSF for each frequent top-clique member, plus Maha++ and NCI), \emph{Oracle-on-train} (the lowest training-AUGRC CSF used universally), \emph{Random-CSF} (uniform draw).

\paragraph{Statistical tests.} For each (regime, side) cell we test whether the predictor's per-row regret is smaller than each baseline using the one-sided paired Wilcoxon signed-rank test ($H_{1}: R^{\text{pred}} < R^{\text{baseline}}$); $p$-values within each cell are corrected for multiple comparisons across baselines with the Holm--Bonferroni procedure at $\alpha = 0.05$. Appendix~\ref{appendix:mean_regret} adds a test on the mean difference. We report the joint head$+$feature regret as the headline number, plus the head-only and feature-only restrictions to verify that the gains hold for both detector families. 

We also test whether the competitive CSFs are source dependent and regime dependent: a one-hot source-dataset encoding (NC + source) and a one-hot regime indicator, which assumes the regime is known (Appendix~\ref{appendix:prediction}). Unlike these ``ablation tests'', our proposed approach has no OOD-regime input: the regime of a deployment shift is not known in advance, and inferring it from a batch would assume a known mix of ID and OOD samples.

\section{Results}\label{sec:results}
\subsection{Experimental Setup}\label{results:experimental_setup}
    We follow the FD-Shifts protocol~\cite{jaeger2022call,traub2024overcoming}, using the cell factors defined in Section~\ref{methods:def_notations}. Source datasets are CIFAR-10, CIFAR-100, SuperCIFAR-100~\cite{krizhevsky2009learning} (CIFAR-100 grouped into its 20 coarse superclasses), and TinyImageNet~\cite{le2015tiny}. The OOD evaluation sets, accessed through the FD-Shifts data pipeline, are iSUN~\cite{xu2015turkergaze}, LSUN-cropped and LSUN-resize~\cite{yu2015lsun}, SVHN~\cite{netzer2011reading}, Places365~\cite{zhou2017places}, and Textures (DTD)~\cite{cimpoi2014describing}. Backbones are VGG-13 trained from scratch (the FD-Shifts CNN) and a ViT fine-tuned from pretrained weights, with 5 random seeds per (backbone, paradigm) combination; the cross-architecture transfer test of Section~\ref{methods:prediction} additionally holds out a ResNet-18 pool (one seed per cell). We evaluate 22 CSFs grouped into two sides. \emph{Head-side} scores are computed from logits or softmax probabilities (MSR, MLS, Energy, GEN, REN, PCE, GE, PE, GradNorm, pNML, plus the learned Confidence readout produced by the chosen training paradigm of Section~\ref{methods:def_notations})~\cite{hendrycks2016baseline,hendrycks2019scaling,liu2020energy,liu2023gen,huang2021importance,bibas2021single,corbiere2021confidence,devries2018learning,liu2019deep}. \emph{Feature-side} scores are computed from penultimate-layer features (CTM, fDBD, NNGuide, NeCo, Maha, Maha++, NCI, ViM, Residual, PCA RecError, KPCA RecError)~\cite{ngoc2023cosine,liu2023fast,park2023nearest,ammar2023neco,lee2018simple,mueller2025mahalanobis,liu2023detecting,wang2022vim,guan2023revisit,fang2025kernel}. ReAct and ASH-S wrappers are evaluated on the CNNs in Appendix~\ref{appendix:react_ash}. Performance is measured with two threshold-free ranking metrics: the Area Under the Risk-Coverage Curve (AURC)~\cite{geifman2018bias} and the Area Under the Generalized Risk-Coverage Curve (AUGRC)~\cite{traub2024overcoming}, which aggregates misclassification risk across coverage levels and corrects for the baseline dependence of AURC; lower is better. An ID-only validation split is used for early stopping, CSF hyperparameter tuning (e.g., temperature scaling, PCA dimensionality), and model selection (full hyperparameter values in Appendix~\ref{appendix:hyperparameter_selection}). OOD results are reported separately for the CLIP-derived near, mid, and far groups (Table~\ref{tab:clip_clustering}). When evaluating OOD detection, we compare OOD samples only against correctly classified ID samples, so a CSF is not rewarded for flagging ordinary ID errors; misclassification detection is evaluated separately on the ID test set by ranking correct versus incorrect ID predictions. Appendix~\ref{appendix:compute} reports the hardware and the compute.The appendix also analyzes projection filtering, matched pairs, FPR@95, and binary grouping use the 20 CSFs without Maha++ and NCI.

\subsection{Top Cliques}\label{results:top_cliques}
    \begin{figure*}[htb]
    \centering
    \includegraphics[width=\textwidth]{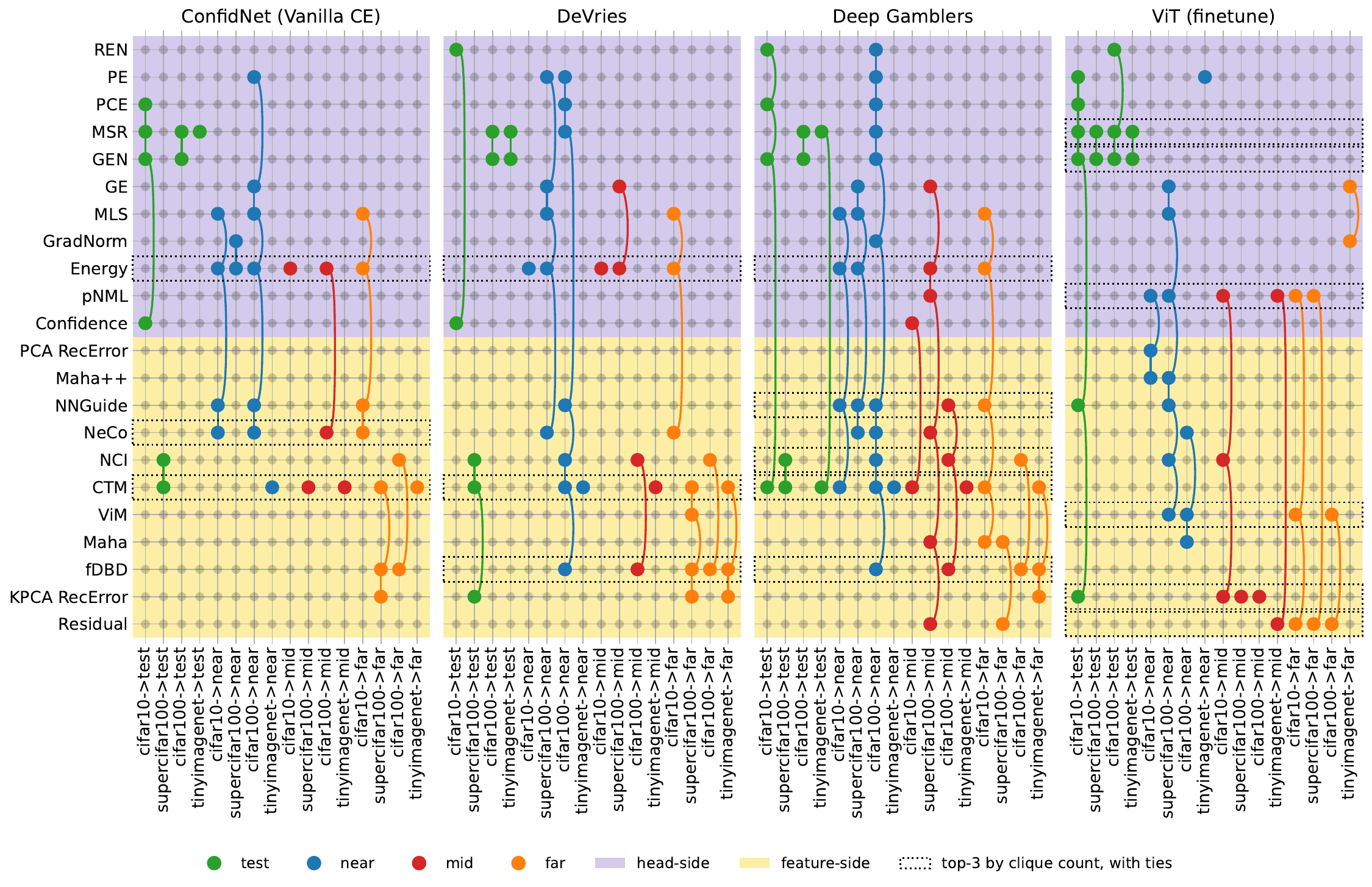}
    \caption{Top-clique maps across four training settings (AURC/AUGRC, $\alpha=0.05$, 22 CSFs). Panels, left to right: VGG-13 with vanilla cross-entropy (the ConfidNet paradigm of Section~\ref{methods:def_notations}), VGG-13 with DeVries, VGG-13 with Deep Gamblers, and a fine-tuned ViT. Columns within each panel are the four source datasets crossed with the four evaluation regimes (\emph{test}, \emph{near}, \emph{mid}, \emph{far}); connected colored dots within a column show the Conover top clique. Rows are the filtered set of CSFs, grouped by where the score reads from the network: head-side scores (logit/softmax statistics, blue band) on top and feature-side scores (penultimate-feature geometry, peach band) on the bottom. Within each band, rows are ordered by regime centre-of-mass (near-dominant at top, far-dominant at bottom). Dotted rectangles mark the CSFs with the three highest clique counts per paradigm, including ties.}
    \label{fig:top_clique_vgg13}
\end{figure*}

The main empirical finding is that the competitive detector family transitions systematically with shift severity, and that this transition differs between architectures. On CNNs, the progression runs from probabilistic scores (ID) to margin-based scores (near-OOD) to geometry-aware scores (mid/far-OOD). On ViTs, the non-ID winners are more heterogeneous, with reconstruction- and residual-based CSFs replacing geometry-aware ones in mid and far regimes. We interpret these patterns geometrically in Section~\ref{results:neural_collapse}.

\paragraph{VGG-13.} The three left panels of Figure~\ref{fig:top_clique_vgg13} show Conover top cliques ($\alpha=0.05$) for the three VGG-13 training paradigms (vanilla CE / ConfidNet head, DeVries, Deep Gamblers); the broad transition is consistent across all three. In the \emph{source$\rightarrow$test} columns (misclassification detection), probabilistic scores dominate (\textsc{GEN}, \textsc{MSR}, learned confidence heads). Across \emph{near} shifts the persistent winners are \textsc{Energy} and \textsc{MLS}, with \textsc{NNGuide} also appearing repeatedly. As semantic distance grows, the top cliques shift to geometry-aware scores: \textsc{CTM}, \textsc{fDBD}, and \textsc{Energy} on \emph{mid}, and \textsc{CTM}, \textsc{fDBD}, \textsc{NNGuide} on \emph{far}. The dotted rectangles in each panel (marking the CSFs with the three highest clique counts per paradigm, including ties) reinforce this: \textsc{CTM} and \textsc{Energy} are marked in all three panels and \textsc{fDBD} in two, followed by \textsc{NeCo}, \textsc{NNGuide}, and \textsc{NCI}.

\paragraph{ViT.} The rightmost panel of Figure~\ref{fig:top_clique_vgg13} shows the same analysis on a fine-tuned ViT. The misclassification regime again favors probabilistic scores (\textsc{GEN}, \textsc{MSR}), but the non-ID story differs in three ways. (i) \emph{Near} is more heterogeneous: no single family dominates, with \textsc{PCA RecError}, \textsc{ViM}, \textsc{pNML}, \textsc{NNGuide}, \textsc{MLS}, and \textsc{Maha++} all appearing across source datasets. (ii) \emph{Mid} is sharply focused: \textsc{KPCA RecError} anchors three of the four mid-OOD columns. (iii) \emph{Far} shifts again toward \textsc{pNML}, \textsc{Residual}, and \textsc{ViM}, with \textsc{GradNorm} and \textsc{GE} appearing for TinyImageNet. The ViT transition is more abrupt than VGG-13's: each regime has a distinct winning family rather than a gradual shift. The fine-tuned ViTs lie in a narrow band of weak collapse (mean self-duality per source $0.85$--$1.27$, against $0.03$--$0.13$ for VGG-13; Appendix~\ref{appendix:neural_collapse}); there, the margin-to-geometry transition of the CNNs does not appear, and reconstruction and residual scores are in the top cliques of every OOD regime. Feature normalization, the motivating example of Section~1, helps but does not change the competitive set at CIFAR scale: \textsc{Maha++} beats \textsc{Maha} on $93\%$ of VGG-13 and $61\%$ of ViT rows but rarely reaches a top clique, whereas on ImageNet it is the most frequent oracle CSF ($34\%$ of rows; \textsc{Maha} $6\%$; Appendix~\ref{appendix:imagenet}).

\paragraph{Three-group stratification matters.} Under OpenOOD's binary near/far grouping (Appendix~\ref{appendix:openood_comparison}), mid-OOD winners are absorbed into the far group, inflating the set of statistically tied CSFs. For example, on CIFAR-100 with ViT, the binary far clique contains 8 CSFs, whereas our three-group analysis isolates a focused mid clique (\textsc{KPCA RecError}) and a distinct far clique (\textsc{Residual}, \textsc{ViM}). Per-paradigm breakouts preserve the broad regime-level transition but shift specific winners (e.g., Deep Gamblers on VGG-13 promotes Mahalanobis and Residual into far cliques otherwise dominated by CTM and fDBD). Figure~\ref{fig:top_clique_vgg13} is restricted to base CSFs; next, Section~\ref{results:projection_gains} quantifies how projection filtering reshapes this competitive landscape.

\paragraph{Threshold metric.} Under FPR@95TPR, the AUGRC winner stays in the top clique in $83\%$ of the $48$ OOD cells. FPR@95 cliques are larger ($4.1$ vs $2.5$ members), so it mostly declares more winners, not different ones; Deep Gamblers is the most metric-sensitive panel (Appendix~\ref{appendix:fpr95}).

\subsection{Projection-Filtering Gains}\label{results:projection_gains}
    We now ask whether projection filtering significantly improves specific CSFs and whether the useful filter type depends on architecture. We compute paired AUGRC differences between each base CSF and its projection variants across all source datasets, OOD datasets, training paradigms, and runs, and test significance with a Wilcoxon signed-rank test (positive $\Delta$ = variant better). Per-CSF $\Delta$AUGRC, win/loss counts, $p$-values, and clique-membership changes are tabulated in Appendix~\ref{appendix:projection_full}. The gains are not uniform; the finding is that the useful filter follows the representation.

\paragraph{Largest gains are on distance- and reconstruction-based CSFs.} \textsc{PCA RecError} improves by $+38.6$ AUGRC under class-predicted projection on CNNs (32/0) and by $+11.6$ on ViTs (26/6); \textsc{Maha} improves by $+24.1$ under global projection on CNNs (50/14). These scores measure distances or residuals in a subspace; removing nuisance dimensions concentrates the signal.

\paragraph{Benefits of global vs.\ class-specific filtering depend on architecture.} On CNNs, global projection lifts \textsc{CTM}, \textsc{fDBD}, and \textsc{NNGuide} (win rates 70--80\%), and class-conditional variants either do not help or hurt (e.g., \textsc{CTM class pred} is significantly worse than the base, $p < 0.001$). On ViTs, the pattern flips: class-predicted projection (CPP) drives the largest gain (\textsc{GradNorm} $+5.91$, 25/7), and global projection delivers a broad but modest lift to logit-based scores (\textsc{Energy}, \textsc{MLS}, \textsc{MSR}, \textsc{GEN}, \textsc{GE}, \textsc{PE}, \textsc{PCE} each gain $+0.50$--$0.65$ AUGRC, win rates 66--88\%). This matches the NC analysis (Section~\ref{results:neural_collapse}): CNN representations are close to collapse and a shared low-rank subspace suffices, while fine-tuned ViTs carry pretraining residuals that class-conditional filtering removes better.

\paragraph{AUGRC gains translate into clique membership, but not always.} Replacing each base CSF with its best significant variant and re-running the rank pipeline yields the clearest gains on ViTs: \textsc{GradNorm} jumps from 1 to 4 cliques with CPP and \textsc{PCA RecError} from 1 to 3; on CNNs, \textsc{pNML} and \textsc{PCA RecError} each enter 2 new cliques. Some CSFs with large $\Delta$AUGRC still fail to enter any clique (\textsc{Maha} on CNNs; \textsc{REN}, \textsc{Energy} on ViTs), and others (\textsc{NNGuide} on CNNs; \textsc{MLS}, \textsc{PE} on ViTs) lose clique membership after competing variants of CSFs outperform them.  The base versus variant results(Appendix~\ref{appendix:projection_full}) detail the gains.

To summarize, use of projection-filtering depends on architecture and CSF:  On CNNs trained from scratch, use global projection for distance- and boundary-based CSFs (\textsc{Maha}, \textsc{CTM}, \textsc{fDBD}, \textsc{NNGuide}) and class-predicted projection only for \textsc{PCA RecError}. On fine-tuned ViTs, use class-predicted projection for \textsc{GradNorm} and \textsc{PCA RecError}, and global projection for a small, consistent gain on logit-based CSFs. We explore possible  mechanistic explanations of these gains in Appendices~\ref{appendix:neural_collapse} and~\ref{appendix:projection_full}.
\subsection{Predicting Detector Selection from NC Geometry}\label{results:neural_collapse}
    We now turn from \emph{which} CSFs are competitive to \emph{whether} representation geometry, summarized by NC metrics, can predict competitive CSFs. Qualitatively, prototype- and boundary-aware scores become more competitive as the representation collapses and aligns with classifier weights, while gradient- and reconstruction-based scores remain stronger when collapse is weaker. Appendix~\ref{appendix:neural_collapse} gives an exploratory reading per dataset: strong variability collapse and self-duality put the CTM score of ID samples near $1$ (TinyImageNet CNNs); weak self-duality favors the boundary margin of fDBD (CIFAR-100 CNNs); and norm-invariant scores replace norm-sensitive logit scores at far OOD.

\paragraph{NC ranks correlate with CSF ranks.} We validate the qualitative link with a Mantel test~\cite{mantel1967detection} that compares pairwise NC distance and pairwise rank-vector distance across model configurations (full details in Appendix~\ref{appendix:mantel}). Using 9{,}999 permutations with Spearman rank distance, the correlation is significant for all four training paradigms (Conv/ConfidNet $r=0.63$, Conv/DeVries $r=0.62$, Conv/Deep~Gamblers $r=0.37$, ViT $r=0.70$; all $p \leq 0.002$). It is strongest at mid OOD on ConfidNet and ViT ($r=0.67$, $0.84$) and at far OOD on DeVries and Deep~Gamblers ($r=0.70$, $0.50$). The association is not only a proxy for training convergence: a partial Mantel test that controls for ID performance is significant in all four strata ($r = 0.32$--$0.59$), and when a control for source identity is added it remains significant for three of four (these results also hold with Papyan's NC1, see Appendix~\ref{appendix:partial_mantel}). Matched pairs show that it follows the training paradigm, not the dropout setting (Appendix~\ref{appendix:matched_pairs}). Maximal equiangularity of classifier weights is the strongest single metric in 3 of 4 paradigms ($r=0.68$--$0.77$; on Deep Gamblers, equiangularity of classifier weights at $r=0.88$).

\paragraph{From correlation to prediction.} Correlation alone is not actionable, so we test whether the same NC measurements, and nothing else, predict a useful detector shortlist for a model the predictor has never seen. Per-CSF $L_2$ logistic models are fit on the VGG-13 pool and applied, without re-training, to a held-out ResNet-18 pool with the same paradigms and source datasets; each shortlist is scored by its per-row set-regret against the per-cell oracle CSF (Section~\ref{methods:prediction}).

\begin{figure}[t]
  \centering
  \includegraphics[width=0.85\textwidth]{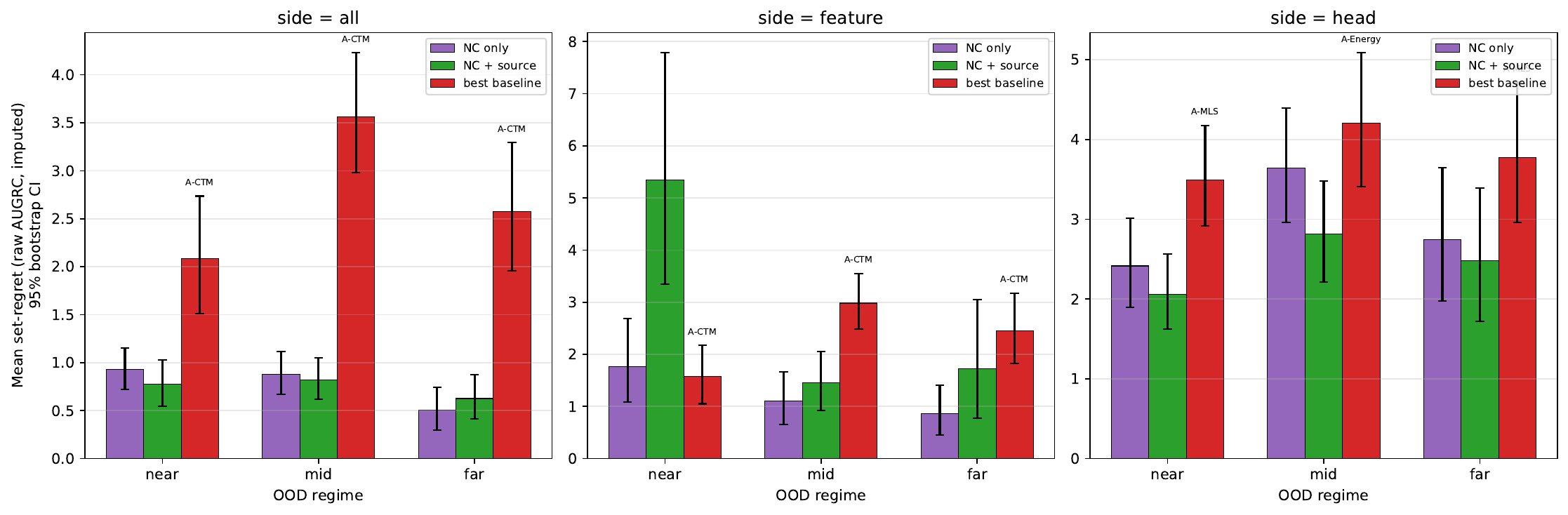}
  \caption{Cross-architecture transfer (VGG-13 $\rightarrow$ ResNet-18, 22 CSFs). Mean set-regret (raw AUGRC, empty-set imputed) with $95\%$ bootstrap CIs for the regime-free NC-based predictor (NC only, the headline; NC + source, an ablation) and the strongest fixed-CSF baseline of each cell (named above its bar; candidates include Always-Maha++ and Always-NCI). Panels: joint / feature-only / head-only side; bar groups: near, mid, far. Lower is better.}
  \label{fig:regret_xarch}
\end{figure}

On the joint head$+$feature side, the NC-only predictor cuts mean set-regret relative to the strongest fixed CSF (Always-CTM) by $55\%$ near ($0.93$ vs $2.08$ AUGRC), $75\%$ mid ($0.88$ vs $3.56$), and $80\%$ far ($0.50$ vs $2.57$). It also improves on the feature-only side at mid ($1.10$ vs $2.98$) and far ($0.87$ vs $2.45$) and on the head-only side in all three regimes ($2.41$/$3.64$/$2.75$ vs $3.49$/$4.20$/$3.77$); its mean regret is higher only at feature-only near ($1.76$ vs $1.58$, Always-CTM). Each model gets one shortlist for all regimes ($8.1$--$8.4$ of $22$ CSFs, never empty). Random shortlists of the same size have higher joint-side regret ($1.46$/$1.38$/$1.43$; significant in every regime) but not higher head-side regret (Appendix~\ref{appendix:mean_regret}).

To interpret the classifier we examine its coefficients (Appendix~\ref{appendix:prediction}), which indicate that CTM is associated with tight collapse and high self-duality, and fDBD and NCI are associated to cases where the class means differ from the classifier weights.


\paragraph{Statistical confirmation.} Holm-corrected paired Wilcoxon tests ($\alpha = 0.05$) over the $66$ (regime, side, baseline) comparisons show the predictor beats the baseline in all $65$ cases where mean regret is lower. The test ranks the per-row differences, so it tests the typical row, not the mean: with heavy-tailed differences (small gains on most rows, a few large losses), a cell can pass while its mean regret is higher (Appendix~\ref{appendix:mean_regret} gives an example). A mean-difference test finds the same $65$ wins and no losses. On the joint side the predictor beats all ten NC-free baselines in every regime ($p_{\text{Holm}} \leq 0.002$ under both tests).

\section{Limitations}\label{results:limitations}
\paragraph{Established and exploratory claims.} The top cliques, the NC--ranking association, and the VGG-13 $\to$ ResNet-18 transfer rest on the full design. In comparison, the mechanistic interpretation of NC and projection filtering (Appendices~\ref{appendix:neural_collapse},~\ref{appendix:projection_full}) and all transfers beyond the CNN pools (Appendices~\ref{appendix:vit_transfer}--\ref{appendix:imagenet}) are exploratory. The NC metric associations established may not be causal; one could induce controlled degrees of collapse during training and check whether the competitive family moves as predicted.

\paragraph{Scope.} Our study is restricted to image classification on four small/medium datasets (CIFAR-10/100, SuperCIFAR-100, TinyImageNet) with two main backbones (VGG-13, fine-tuned ViT) and three training paradigms (ConfidNet, DeVries, Deep Gamblers). Masked-image pretraining, supervised contrastive training, and full fine-tuning of self-supervised encoders are not covered. Results may not transfer to detection/segmentation or to temporal/causal/task-specific drifts. Two untested hypotheses: under class imbalance, minority collapse distorts the classifier weights~\cite{fang2021exploring}, so head-side CSFs should degrade first and the weight-side NC metrics should carry the signal. In multi-label training, the means of multi-label features are scaled averages of single-label means~\cite{li2023multilabel}, so the eight NC metrics would need multi-label descriptors.

\paragraph{Shift stratification.} The regimes depend on the CLIP encoder and the clustering choices. CLIP embeddings may be less faithful for fine-grained classes, such as aircraft variants or flower species~\cite{radford2021learning}, and for industrial defects, such as a scratch on a known part~\cite{zhou2024anomalyclip}. They also need not match the failure modes of the classifier (Appendix~\ref{appendix:clip_limits}).

\paragraph{Cross-architecture transfer.} The main transfer test is CNN $\to$ CNN (VGG-13 $\to$ ResNet-18). On 47 ImageNet models and fine-tuned ViTs, a random shortlist of the same size reaches about the same regret (Appendices~\ref{appendix:imagenet},~\ref{appendix:vit_transfer}). On self-supervised probes the predictor fails at far OOD (Appendix~\ref{appendix:ssl}). Heads trained on ViTs of other sources do not fix the ViT case (Appendix~\ref{appendix:vit_transfer}), so in our data the NC metrics select detectors well only between closely related CNN families. With one run per ResNet-18 cell, the rank pipeline cannot run there, so we score shortlists by per-row regret against an oracle CSF.

\paragraph{Statistical methodology.} The Mantel associations are correlational; the causal test, training with controlled degrees of collapse~\cite{harun2025controlling}, is future work. The rank-based pipeline (Friedman $\to$ Conover $\to$ Bron-Kerbosch cliques) assumes complete blocks and exchangeability, and its pairwise tests are not adjusted for multiplicity (Appendix~\ref{appendix:holm}). The size of the predictor's gain depends on how NC1 is normalized (Appendix~\ref{appendix:neural_collapse}). Hyperparameters were tuned for AUGRC, so the best CSFs may shift under another metric; under FPR@95TPR, the Deep Gamblers cliques change most (Appendix~\ref{appendix:fpr95}). Computational cost and latency are not evaluated.

\section{Conclusion}
Our main conclusion is that the family of competitive OOD detectors (CSFs) change systematically with the learned representation and the shift severity: probabilistic scores for in-distribution misclassification, margin-based scores in near-OOD on CNNs, geometry-aware scores in mid/far-OOD, and reconstruction/residual scores on fine-tuned ViTs. The three-group CLIP stratification matters here, because a binary near/far split hides the mid-OOD winners. The effects of PCA-based projection filtering also  follow the same geometry: global projection lifts feature-side CSFs on CNNs, and class-predicted projection gives the largest gains on fine-tuned ViTs. NC geometry and method rankings are significantly associated (Mantel $r=0.37$--$0.70$, $p \leq 0.002$), and a per-CSF $L_{2}$ logistic predictor that uses only the NC metrics transfers between CNN architectures: under VGG-13 $\rightarrow$ ResNet-18 transfer, one shortlist per model cuts mean set-regret over the strongest fixed CSF by 55--80\% on near/mid/far OOD, without OOD data for that model and without knowledge of the shift. However, transfer to other architecture families is only partial (Appendices~\ref{appendix:vit_transfer}--\ref{appendix:imagenet}). In summary, to know which OOD detector will work, first measure the geometry of the representation it will see.
\begin{ack}
This material is based upon work supported by the U.S. National Science Foundation under Cooperative Agreement No. 2433450. Research was sponsored by the Department of the Navy, Office of Naval Research under ONR award number N00014-24-1-2259. This research was partially supported by the University of Delaware's Data Science Institute through a seed grant from the Unidel Foundation.
\end{ack}

\bibliography{icml_bibliography}

\newpage
\appendix

\section{Hyperparameter selection}\label{appendix:hyperparameter_selection}
    This appendix reports the hyperparameters used to evaluate each (CSF, projection variant, source dataset, training paradigm) combination in our study. For every combination, we select the values that minimize the AUGRC on the validation set; these configurations underlie the top cliques reported in Section~\ref{results:top_cliques}. Temperature scaling is applied to all logit-based CSFs, with the temperature tuned on the validation set. Table~\ref{tab:hyperparam_cnn} lists the chosen training-time dropout setting for VGG-13 (CNN) models, Table~\ref{tab:hyperparam_vit} for the fine-tuned ViT, and Table~\ref{tab:hyperparam_dg} reports the chosen Deep Gamblers reservation reward $o$. Cells marked \emph{N/A} indicate that the (CSF, projection variant) combination is not applicable. The tables cover 20 CSFs; for Maha++ and NCI (base variant only) the configuration is selected by the same validation rule, and the selections are part of the released score files. The cliques used as prediction targets in Section~\ref{methods:prediction} are formed per (paradigm, source, dropout, reward, regime) cell, aggregating only over training runs (5 seeds per cell). This means that dropout and reward are not additionally optimized at the clique level for the prediction task.

\begin{table}[htbp]
    \scriptsize 
    \caption{CNN (VGG-13) hyperparameters: dropout setting used during training (1 = trained with dropout, 0 = without). \emph{N/A} indicates the (CSF, projection variant) combination is not applicable.}
    \label{tab:hyperparam_cnn}
    \hspace{-2.5cm}
    \begin{subtable}[t]{\textwidth}
        \centering
\begin{tabular}{clccccc@{\hspace{0.6em}}ccccc@{\hspace{0.6em}}ccccc}
\toprule
& & \multicolumn{5}{c}{ConfidNet} & \multicolumn{5}{c}{DeVries} & \multicolumn{5}{c}{Deep Gamblers} \\
\cmidrule(lr){3-7}\cmidrule(lr){8-12}\cmidrule(lr){13-17}
& CSF & base & class & class avg & class pred & global & base & class & class avg & class pred & global & base & class & class avg & class pred & global \\
\midrule
\multirow[c]{20}{*}{\rotatebox[origin=c]{90}{CIFAR-10}} & CTM & {\cellcolor[HTML]{E6FFE6}} 0 & {\cellcolor[HTML]{E6FFE6}} 0 & {\cellcolor[HTML]{E6FFE6}} 0 & {\cellcolor[HTML]{E6FFE6}} 0 & {\cellcolor[HTML]{E6FFE6}} 0 & {\cellcolor[HTML]{E6FFE6}} 0 & {\cellcolor[HTML]{E6FFE6}} 0 & {\cellcolor[HTML]{E6FFE6}} 0 & {\cellcolor[HTML]{E6FFE6}} 0 & {\cellcolor[HTML]{E6FFE6}} 0 & {\cellcolor[HTML]{E6FFE6}} 0 & {\cellcolor[HTML]{E6FFE6}} 0 & {\cellcolor[HTML]{E6FFE6}} 0 & {\cellcolor[HTML]{E6FFE6}} 0 & {\cellcolor[HTML]{E6FFE6}} 0 \\
& Confidence & {\cellcolor[HTML]{FFE6E6}} 1 & {\cellcolor[HTML]{FFFFFF}} N/A & {\cellcolor[HTML]{FFFFFF}} N/A & {\cellcolor[HTML]{FFFFFF}} N/A & {\cellcolor[HTML]{FFFFFF}} N/A & {\cellcolor[HTML]{E6FFE6}} 0 & {\cellcolor[HTML]{FFFFFF}} N/A & {\cellcolor[HTML]{FFFFFF}} N/A & {\cellcolor[HTML]{FFFFFF}} N/A & {\cellcolor[HTML]{FFFFFF}} N/A & {\cellcolor[HTML]{E6FFE6}} 0 & {\cellcolor[HTML]{FFFFFF}} N/A & {\cellcolor[HTML]{FFFFFF}} N/A & {\cellcolor[HTML]{FFFFFF}} N/A & {\cellcolor[HTML]{FFFFFF}} N/A \\
& Energy & {\cellcolor[HTML]{FFE6E6}} 1 & {\cellcolor[HTML]{FFE6E6}} 1 & {\cellcolor[HTML]{FFE6E6}} 1 & {\cellcolor[HTML]{FFE6E6}} 1 & {\cellcolor[HTML]{FFE6E6}} 1 & {\cellcolor[HTML]{FFE6E6}} 1 & {\cellcolor[HTML]{FFE6E6}} 1 & {\cellcolor[HTML]{E6FFE6}} 0 & {\cellcolor[HTML]{FFE6E6}} 1 & {\cellcolor[HTML]{FFE6E6}} 1 & {\cellcolor[HTML]{FFE6E6}} 1 & {\cellcolor[HTML]{FFE6E6}} 1 & {\cellcolor[HTML]{E6FFE6}} 0 & {\cellcolor[HTML]{FFE6E6}} 1 & {\cellcolor[HTML]{FFE6E6}} 1 \\
\bottomrule
\end{tabular}
\end{subtable}
\vfill
    \hspace{-2.5cm}
\begin{subtable}[t]{\textwidth}
    \centering
\begin{tabular}{clccccc@{\hspace{0.6em}}ccccc@{\hspace{0.6em}}ccccc}
\toprule
& & \multicolumn{5}{c}{ConfidNet} & \multicolumn{5}{c}{DeVries} & \multicolumn{5}{c}{Deep Gamblers} \\
\cmidrule(lr){3-7}\cmidrule(lr){8-12}\cmidrule(lr){13-17}
& CSF & base & class & class avg & class pred & global & base & class & class avg & class pred & global & base & class & class avg & class pred & global \\
\midrule
\multirow[c]{20}{*}{\rotatebox[origin=c]{90}{SuperCIFAR-100}} & CTM & {\cellcolor[HTML]{E6FFE6}} 0 & {\cellcolor[HTML]{E6FFE6}} 0 & {\cellcolor[HTML]{E6FFE6}} 0 & {\cellcolor[HTML]{E6FFE6}} 0 & {\cellcolor[HTML]{E6FFE6}} 0 & {\cellcolor[HTML]{E6FFE6}} 0 & {\cellcolor[HTML]{E6FFE6}} 0 & {\cellcolor[HTML]{E6FFE6}} 0 & {\cellcolor[HTML]{E6FFE6}} 0 & {\cellcolor[HTML]{E6FFE6}} 0 & {\cellcolor[HTML]{E6FFE6}} 0 & {\cellcolor[HTML]{E6FFE6}} 0 & {\cellcolor[HTML]{E6FFE6}} 0 & {\cellcolor[HTML]{E6FFE6}} 0 & {\cellcolor[HTML]{E6FFE6}} 0 \\
& Confidence & {\cellcolor[HTML]{E6FFE6}} 0 & {\cellcolor[HTML]{FFFFFF}} N/A & {\cellcolor[HTML]{FFFFFF}} N/A & {\cellcolor[HTML]{FFFFFF}} N/A & {\cellcolor[HTML]{FFFFFF}} N/A & {\cellcolor[HTML]{E6FFE6}} 0 & {\cellcolor[HTML]{FFFFFF}} N/A & {\cellcolor[HTML]{FFFFFF}} N/A & {\cellcolor[HTML]{FFFFFF}} N/A & {\cellcolor[HTML]{FFFFFF}} N/A & {\cellcolor[HTML]{E6FFE6}} 0 & {\cellcolor[HTML]{FFFFFF}} N/A & {\cellcolor[HTML]{FFFFFF}} N/A & {\cellcolor[HTML]{FFFFFF}} N/A & {\cellcolor[HTML]{FFFFFF}} N/A \\
& Energy & {\cellcolor[HTML]{FFE6E6}} 1 & {\cellcolor[HTML]{FFE6E6}} 1 & {\cellcolor[HTML]{E6FFE6}} 0 & {\cellcolor[HTML]{FFE6E6}} 1 & {\cellcolor[HTML]{FFE6E6}} 1 & {\cellcolor[HTML]{FFE6E6}} 1 & {\cellcolor[HTML]{FFE6E6}} 1 & {\cellcolor[HTML]{E6FFE6}} 0 & {\cellcolor[HTML]{FFE6E6}} 1 & {\cellcolor[HTML]{FFE6E6}} 1 & {\cellcolor[HTML]{FFE6E6}} 1 & {\cellcolor[HTML]{FFE6E6}} 1 & {\cellcolor[HTML]{FFE6E6}} 1 & {\cellcolor[HTML]{FFE6E6}} 1 & {\cellcolor[HTML]{FFE6E6}} 1 \\
\bottomrule
\end{tabular}
\end{subtable}
\end{table}
\pagebreak
\begin{table}[htbp]
    \scriptsize
    \centering
    \ContinuedFloat
\hspace{-4.5cm}
\begin{subtable}[t]{\textwidth}
    \begin{tabular}{clccccc@{\hspace{0.6em}}ccccc@{\hspace{0.6em}}ccccc}
\toprule
& & \multicolumn{5}{c}{ConfidNet} & \multicolumn{5}{c}{DeVries} & \multicolumn{5}{c}{Deep Gamblers} \\
\cmidrule(lr){3-7}\cmidrule(lr){8-12}\cmidrule(lr){13-17}
& CSF & base & class & class avg & class pred & global & base & class & class avg & class pred & global & base & class & class avg & class pred & global \\
\midrule
\multirow[c]{20}{*}{\rotatebox[origin=c]{90}{CIFAR-100}} & CTM & {\cellcolor[HTML]{E6FFE6}} 0 & {\cellcolor[HTML]{E6FFE6}} 0 & {\cellcolor[HTML]{E6FFE6}} 0 & {\cellcolor[HTML]{E6FFE6}} 0 & {\cellcolor[HTML]{E6FFE6}} 0 & {\cellcolor[HTML]{E6FFE6}} 0 & {\cellcolor[HTML]{E6FFE6}} 0 & {\cellcolor[HTML]{E6FFE6}} 0 & {\cellcolor[HTML]{E6FFE6}} 0 & {\cellcolor[HTML]{E6FFE6}} 0 & {\cellcolor[HTML]{E6FFE6}} 0 & {\cellcolor[HTML]{E6FFE6}} 0 & {\cellcolor[HTML]{E6FFE6}} 0 & {\cellcolor[HTML]{E6FFE6}} 0 & {\cellcolor[HTML]{E6FFE6}} 0 \\
& Confidence & {\cellcolor[HTML]{FFE6E6}} 1 & {\cellcolor[HTML]{FFFFFF}} N/A & {\cellcolor[HTML]{FFFFFF}} N/A & {\cellcolor[HTML]{FFFFFF}} N/A & {\cellcolor[HTML]{FFFFFF}} N/A & {\cellcolor[HTML]{E6FFE6}} 0 & {\cellcolor[HTML]{FFFFFF}} N/A & {\cellcolor[HTML]{FFFFFF}} N/A & {\cellcolor[HTML]{FFFFFF}} N/A & {\cellcolor[HTML]{FFFFFF}} N/A & {\cellcolor[HTML]{E6FFE6}} 0 & {\cellcolor[HTML]{FFFFFF}} N/A & {\cellcolor[HTML]{FFFFFF}} N/A & {\cellcolor[HTML]{FFFFFF}} N/A & {\cellcolor[HTML]{FFFFFF}} N/A \\
& Energy & {\cellcolor[HTML]{FFE6E6}} 1 & {\cellcolor[HTML]{FFE6E6}} 1 & {\cellcolor[HTML]{FFE6E6}} 1 & {\cellcolor[HTML]{FFE6E6}} 1 & {\cellcolor[HTML]{FFE6E6}} 1 & {\cellcolor[HTML]{E6FFE6}} 0 & {\cellcolor[HTML]{E6FFE6}} 0 & {\cellcolor[HTML]{E6FFE6}} 0 & {\cellcolor[HTML]{E6FFE6}} 0 & {\cellcolor[HTML]{E6FFE6}} 0 & {\cellcolor[HTML]{E6FFE6}} 0 & {\cellcolor[HTML]{E6FFE6}} 0 & {\cellcolor[HTML]{E6FFE6}} 0 & {\cellcolor[HTML]{E6FFE6}} 0 & {\cellcolor[HTML]{E6FFE6}} 0 \\
\bottomrule
\end{tabular}
\end{subtable}
\vfill
\hspace{-4.5cm}
    \begin{subtable}[t]{\textwidth}
    \centering
\begin{tabular}{clccccc@{\hspace{0.6em}}ccccc@{\hspace{0.6em}}ccccc}
\toprule
& & \multicolumn{5}{c}{ConfidNet} & \multicolumn{5}{c}{DeVries} & \multicolumn{5}{c}{Deep Gamblers} \\
\cmidrule(lr){3-7}\cmidrule(lr){8-12}\cmidrule(lr){13-17}
& CSF & base & class & class avg & class pred & global & base & class & class avg & class pred & global & base & class & class avg & class pred & global \\
\midrule
\multirow[c]{20}{*}{\rotatebox[origin=c]{90}{TinyImageNet}} & CTM & {\cellcolor[HTML]{E6FFE6}} 0 & {\cellcolor[HTML]{E6FFE6}} 0 & {\cellcolor[HTML]{E6FFE6}} 0 & {\cellcolor[HTML]{E6FFE6}} 0 & {\cellcolor[HTML]{E6FFE6}} 0 & {\cellcolor[HTML]{E6FFE6}} 0 & {\cellcolor[HTML]{E6FFE6}} 0 & {\cellcolor[HTML]{E6FFE6}} 0 & {\cellcolor[HTML]{E6FFE6}} 0 & {\cellcolor[HTML]{E6FFE6}} 0 & {\cellcolor[HTML]{E6FFE6}} 0 & {\cellcolor[HTML]{E6FFE6}} 0 & {\cellcolor[HTML]{E6FFE6}} 0 & {\cellcolor[HTML]{E6FFE6}} 0 & {\cellcolor[HTML]{E6FFE6}} 0 \\
& Confidence & {\cellcolor[HTML]{E6FFE6}} 0 & {\cellcolor[HTML]{FFFFFF}} N/A & {\cellcolor[HTML]{FFFFFF}} N/A & {\cellcolor[HTML]{FFFFFF}} N/A & {\cellcolor[HTML]{FFFFFF}} N/A & {\cellcolor[HTML]{FFE6E6}} 1 & {\cellcolor[HTML]{FFFFFF}} N/A & {\cellcolor[HTML]{FFFFFF}} N/A & {\cellcolor[HTML]{FFFFFF}} N/A & {\cellcolor[HTML]{FFFFFF}} N/A & {\cellcolor[HTML]{E6FFE6}} 0 & {\cellcolor[HTML]{FFFFFF}} N/A & {\cellcolor[HTML]{FFFFFF}} N/A & {\cellcolor[HTML]{FFFFFF}} N/A & {\cellcolor[HTML]{FFFFFF}} N/A \\
& Energy & {\cellcolor[HTML]{E6FFE6}} 0 & {\cellcolor[HTML]{E6FFE6}} 0 & {\cellcolor[HTML]{E6FFE6}} 0 & {\cellcolor[HTML]{E6FFE6}} 0 & {\cellcolor[HTML]{E6FFE6}} 0 & {\cellcolor[HTML]{E6FFE6}} 0 & {\cellcolor[HTML]{E6FFE6}} 0 & {\cellcolor[HTML]{E6FFE6}} 0 & {\cellcolor[HTML]{E6FFE6}} 0 & {\cellcolor[HTML]{E6FFE6}} 0 & {\cellcolor[HTML]{E6FFE6}} 0 & {\cellcolor[HTML]{E6FFE6}} 0 & {\cellcolor[HTML]{E6FFE6}} 0 & {\cellcolor[HTML]{E6FFE6}} 0 & {\cellcolor[HTML]{E6FFE6}} 0 \\
\bottomrule
\end{tabular}
\end{subtable}
\end{table}

\begin{table}[H]
    \centering
    \scriptsize
    \caption{ViT hyperparameters: dropout setting used during training (1 = trained with dropout, 0 = without). \emph{N/A} indicates the (CSF, projection variant) combination is not applicable.}
    \label{tab:hyperparam_vit}
    \begin{subtable}[t]{0.48\textwidth}
        \centering
        \caption*{CIFAR-10}
        \hspace*{-1.5cm}
        \begin{tabular}{lccccc}
        \toprule
        CSF & base & class & class avg & class pred & global \\
        \midrule
        CTM & {\cellcolor[HTML]{FFE6E6}} 1 & {\cellcolor[HTML]{FFE6E6}} 1 & {\cellcolor[HTML]{FFE6E6}} 1 & {\cellcolor[HTML]{FFE6E6}} 1 & {\cellcolor[HTML]{FFE6E6}} 1 \\
        Confidence & {\cellcolor[HTML]{FFE6E6}} 1 & {\cellcolor[HTML]{FFFFFF}} N/A & {\cellcolor[HTML]{FFFFFF}} N/A & {\cellcolor[HTML]{FFFFFF}} N/A & {\cellcolor[HTML]{FFFFFF}} N/A \\
        Energy & {\cellcolor[HTML]{FFE6E6}} 1 & {\cellcolor[HTML]{FFE6E6}} 1 & {\cellcolor[HTML]{FFE6E6}} 1 & {\cellcolor[HTML]{FFE6E6}} 1 & {\cellcolor[HTML]{FFE6E6}} 1 \\
        GE & {\cellcolor[HTML]{FFE6E6}} 1 & {\cellcolor[HTML]{FFE6E6}} 1 & {\cellcolor[HTML]{FFE6E6}} 1 & {\cellcolor[HTML]{FFE6E6}} 1 & {\cellcolor[HTML]{FFE6E6}} 1 \\
        GEN & {\cellcolor[HTML]{FFE6E6}} 1 & {\cellcolor[HTML]{FFE6E6}} 1 & {\cellcolor[HTML]{FFE6E6}} 1 & {\cellcolor[HTML]{FFE6E6}} 1 & {\cellcolor[HTML]{FFE6E6}} 1 \\
        GradNorm & {\cellcolor[HTML]{FFE6E6}} 1 & {\cellcolor[HTML]{FFFFFF}} N/A & {\cellcolor[HTML]{FFE6E6}} 1 & {\cellcolor[HTML]{FFE6E6}} 1 & {\cellcolor[HTML]{FFE6E6}} 1 \\
        KPCA RecError & {\cellcolor[HTML]{FFFFFF}} N/A & {\cellcolor[HTML]{FFE6E6}} 1 & {\cellcolor[HTML]{FFE6E6}} 1 & {\cellcolor[HTML]{FFE6E6}} 1 & {\cellcolor[HTML]{FFE6E6}} 1 \\
\bottomrule
        \end{tabular}
    \end{subtable}
    \begin{subtable}[t]{0.48\textwidth}
        \centering
        \caption*{SuperCIFAR-100}
        \begin{tabular}{lccccc}
        \toprule
        CSF & base & class & class avg & class pred & global \\
        \midrule
        CTM & {\cellcolor[HTML]{E6FFE6}} 0 & {\cellcolor[HTML]{E6FFE6}} 0 & {\cellcolor[HTML]{E6FFE6}} 0 & {\cellcolor[HTML]{E6FFE6}} 0 & {\cellcolor[HTML]{E6FFE6}} 0 \\
        Confidence & {\cellcolor[HTML]{E6FFE6}} 0 & {\cellcolor[HTML]{FFFFFF}} N/A & {\cellcolor[HTML]{FFFFFF}} N/A & {\cellcolor[HTML]{FFFFFF}} N/A & {\cellcolor[HTML]{FFFFFF}} N/A \\
        Energy & {\cellcolor[HTML]{E6FFE6}} 0 & {\cellcolor[HTML]{E6FFE6}} 0 & {\cellcolor[HTML]{E6FFE6}} 0 & {\cellcolor[HTML]{E6FFE6}} 0 & {\cellcolor[HTML]{E6FFE6}} 0 \\
        GE & {\cellcolor[HTML]{E6FFE6}} 0 & {\cellcolor[HTML]{E6FFE6}} 0 & {\cellcolor[HTML]{E6FFE6}} 0 & {\cellcolor[HTML]{E6FFE6}} 0 & {\cellcolor[HTML]{E6FFE6}} 0 \\
        GEN & {\cellcolor[HTML]{E6FFE6}} 0 & {\cellcolor[HTML]{E6FFE6}} 0 & {\cellcolor[HTML]{E6FFE6}} 0 & {\cellcolor[HTML]{E6FFE6}} 0 & {\cellcolor[HTML]{E6FFE6}} 0 \\
        GradNorm & {\cellcolor[HTML]{E6FFE6}} 0 & {\cellcolor[HTML]{FFFFFF}} N/A & {\cellcolor[HTML]{E6FFE6}} 0 & {\cellcolor[HTML]{E6FFE6}} 0 & {\cellcolor[HTML]{E6FFE6}} 0 \\
        KPCA RecError & {\cellcolor[HTML]{FFFFFF}} N/A & {\cellcolor[HTML]{E6FFE6}} 0 & {\cellcolor[HTML]{E6FFE6}} 0 & {\cellcolor[HTML]{E6FFE6}} 0 & {\cellcolor[HTML]{E6FFE6}} 0 \\
\bottomrule
        \end{tabular}
    \end{subtable}
    \begin{subtable}[t]{0.48\textwidth}
        \centering
        \caption*{CIFAR-100}
        \hspace*{-1.5cm}
        \begin{tabular}{lccccc}
        \toprule
        CSF & base & class & class avg & class pred & global \\
        \midrule
        CTM & {\cellcolor[HTML]{FFE6E6}} 1 & {\cellcolor[HTML]{FFE6E6}} 1 & {\cellcolor[HTML]{FFE6E6}} 1 & {\cellcolor[HTML]{FFE6E6}} 1 & {\cellcolor[HTML]{FFE6E6}} 1 \\
        Confidence & {\cellcolor[HTML]{FFE6E6}} 1 & {\cellcolor[HTML]{FFFFFF}} N/A & {\cellcolor[HTML]{FFFFFF}} N/A & {\cellcolor[HTML]{FFFFFF}} N/A & {\cellcolor[HTML]{FFFFFF}} N/A \\
        Energy & {\cellcolor[HTML]{FFE6E6}} 1 & {\cellcolor[HTML]{FFE6E6}} 1 & {\cellcolor[HTML]{FFE6E6}} 1 & {\cellcolor[HTML]{FFE6E6}} 1 & {\cellcolor[HTML]{FFE6E6}} 1 \\
        GE & {\cellcolor[HTML]{FFE6E6}} 1 & {\cellcolor[HTML]{FFE6E6}} 1 & {\cellcolor[HTML]{FFE6E6}} 1 & {\cellcolor[HTML]{FFE6E6}} 1 & {\cellcolor[HTML]{FFE6E6}} 1 \\
        GEN & {\cellcolor[HTML]{FFE6E6}} 1 & {\cellcolor[HTML]{FFE6E6}} 1 & {\cellcolor[HTML]{FFE6E6}} 1 & {\cellcolor[HTML]{FFE6E6}} 1 & {\cellcolor[HTML]{FFE6E6}} 1 \\
        GradNorm & {\cellcolor[HTML]{FFE6E6}} 1 & {\cellcolor[HTML]{FFFFFF}} N/A & {\cellcolor[HTML]{FFE6E6}} 1 & {\cellcolor[HTML]{FFE6E6}} 1 & {\cellcolor[HTML]{FFE6E6}} 1 \\
        KPCA RecError & {\cellcolor[HTML]{FFFFFF}} N/A & {\cellcolor[HTML]{FFE6E6}} 1 & {\cellcolor[HTML]{FFE6E6}} 1 & {\cellcolor[HTML]{FFE6E6}} 1 & {\cellcolor[HTML]{FFE6E6}} 1 \\
\bottomrule
        \end{tabular}
    \end{subtable}
    \begin{subtable}[t]{0.48\textwidth}
        \centering
        \caption*{TinyImageNet}
        \begin{tabular}{lccccc}
        \toprule
        CSF & base & class & class avg & class pred & global \\
        \midrule
        CTM & {\cellcolor[HTML]{E6FFE6}} 0 & {\cellcolor[HTML]{E6FFE6}} 0 & {\cellcolor[HTML]{E6FFE6}} 0 & {\cellcolor[HTML]{E6FFE6}} 0 & {\cellcolor[HTML]{E6FFE6}} 0 \\
        Confidence & {\cellcolor[HTML]{E6FFE6}} 0 & {\cellcolor[HTML]{FFFFFF}} N/A & {\cellcolor[HTML]{FFFFFF}} N/A & {\cellcolor[HTML]{FFFFFF}} N/A & {\cellcolor[HTML]{FFFFFF}} N/A \\
        Energy & {\cellcolor[HTML]{E6FFE6}} 0 & {\cellcolor[HTML]{E6FFE6}} 0 & {\cellcolor[HTML]{E6FFE6}} 0 & {\cellcolor[HTML]{E6FFE6}} 0 & {\cellcolor[HTML]{E6FFE6}} 0 \\
        GE & {\cellcolor[HTML]{E6FFE6}} 0 & {\cellcolor[HTML]{E6FFE6}} 0 & {\cellcolor[HTML]{E6FFE6}} 0 & {\cellcolor[HTML]{E6FFE6}} 0 & {\cellcolor[HTML]{E6FFE6}} 0 \\
        GEN & {\cellcolor[HTML]{E6FFE6}} 0 & {\cellcolor[HTML]{E6FFE6}} 0 & {\cellcolor[HTML]{E6FFE6}} 0 & {\cellcolor[HTML]{E6FFE6}} 0 & {\cellcolor[HTML]{E6FFE6}} 0 \\
        GradNorm & {\cellcolor[HTML]{E6FFE6}} 0 & {\cellcolor[HTML]{FFFFFF}} N/A & {\cellcolor[HTML]{E6FFE6}} 0 & {\cellcolor[HTML]{E6FFE6}} 0 & {\cellcolor[HTML]{E6FFE6}} 0 \\
        KPCA RecError & {\cellcolor[HTML]{FFFFFF}} N/A & {\cellcolor[HTML]{E6FFE6}} 0 & {\cellcolor[HTML]{E6FFE6}} 0 & {\cellcolor[HTML]{E6FFE6}} 0 & {\cellcolor[HTML]{E6FFE6}} 0 \\
\bottomrule
        \end{tabular}
    \end{subtable}
\end{table}

\begin{table}[htbp]
    \centering
    \scriptsize
    \caption{Deep Gamblers: reservation reward $o$ selected on the validation set, per (CSF, projection variant) and source dataset. \emph{N/A} indicates the combination is not applicable.}
    \label{tab:hyperparam_dg}
    \begin{subtable}[t]{0.48\textwidth}
        \centering
        \caption*{CIFAR-10}
        \hspace*{-1.5cm}
        \begin{tabular}{lccccc}
        \toprule
        CSF & base & class & class avg & class pred & global \\
        \midrule
        CTM & {\cellcolor[HTML]{00224E}} \color[HTML]{F1F1F1} 2.2 & {\cellcolor[HTML]{00224E}} \color[HTML]{F1F1F1} 2.2 & {\cellcolor[HTML]{6F7073}} \color[HTML]{F1F1F1} 10.0 & {\cellcolor[HTML]{00224E}} \color[HTML]{F1F1F1} 2.2 & {\cellcolor[HTML]{002A61}} \color[HTML]{F1F1F1} 3.0 \\
        Confidence & {\cellcolor[HTML]{00224E}} \color[HTML]{F1F1F1} 2.2 & {\cellcolor[HTML]{FFFFFF}} N/A & {\cellcolor[HTML]{FFFFFF}} N/A & {\cellcolor[HTML]{FFFFFF}} N/A & {\cellcolor[HTML]{FFFFFF}} N/A \\
        Energy & {\cellcolor[HTML]{002A61}} \color[HTML]{F1F1F1} 3.0 & {\cellcolor[HTML]{002A61}} \color[HTML]{F1F1F1} 3.0 & {\cellcolor[HTML]{6F7073}} \color[HTML]{F1F1F1} 10.0 & {\cellcolor[HTML]{002A61}} \color[HTML]{F1F1F1} 3.0 & {\cellcolor[HTML]{002A61}} \color[HTML]{F1F1F1} 3.0 \\
        GE & {\cellcolor[HTML]{6F7073}} \color[HTML]{F1F1F1} 10.0 & {\cellcolor[HTML]{6F7073}} \color[HTML]{F1F1F1} 10.0 & {\cellcolor[HTML]{6F7073}} \color[HTML]{F1F1F1} 10.0 & {\cellcolor[HTML]{002A61}} \color[HTML]{F1F1F1} 3.0 & {\cellcolor[HTML]{6F7073}} \color[HTML]{F1F1F1} 10.0 \\
        GEN & {\cellcolor[HTML]{6F7073}} \color[HTML]{F1F1F1} 10.0 & {\cellcolor[HTML]{6F7073}} \color[HTML]{F1F1F1} 10.0 & {\cellcolor[HTML]{6F7073}} \color[HTML]{F1F1F1} 10.0 & {\cellcolor[HTML]{6F7073}} \color[HTML]{F1F1F1} 10.0 & {\cellcolor[HTML]{6F7073}} \color[HTML]{F1F1F1} 10.0 \\
        GradNorm & {\cellcolor[HTML]{002A61}} \color[HTML]{F1F1F1} 3.0 & {\cellcolor[HTML]{FFFFFF}} N/A & {\cellcolor[HTML]{39486C}} \color[HTML]{F1F1F1} 6.0 & {\cellcolor[HTML]{002A61}} \color[HTML]{F1F1F1} 3.0 & {\cellcolor[HTML]{002A61}} \color[HTML]{F1F1F1} 3.0 \\
\bottomrule
        \end{tabular}
    \end{subtable}
    \hfill
    \begin{subtable}[t]{0.48\textwidth}
        \centering
        \caption*{SuperCIFAR-100}
        \begin{tabular}{lccccc}
        \toprule
        CSF & base & class & class avg & class pred & global \\
        \midrule
        CTM & {\cellcolor[HTML]{6F7073}} \color[HTML]{F1F1F1} 10.0 & {\cellcolor[HTML]{002A61}} \color[HTML]{F1F1F1} 3.0 & {\cellcolor[HTML]{6F7073}} \color[HTML]{F1F1F1} 10.0 & {\cellcolor[HTML]{002A61}} \color[HTML]{F1F1F1} 3.0 & {\cellcolor[HTML]{6F7073}} \color[HTML]{F1F1F1} 10.0 \\
        Confidence & {\cellcolor[HTML]{002A61}} \color[HTML]{F1F1F1} 3.0 & {\cellcolor[HTML]{FFFFFF}} N/A & {\cellcolor[HTML]{FFFFFF}} N/A & {\cellcolor[HTML]{FFFFFF}} N/A & {\cellcolor[HTML]{FFFFFF}} N/A \\
        Energy & {\cellcolor[HTML]{FEE838}} \color[HTML]{000000} 20.0 & {\cellcolor[HTML]{FEE838}} \color[HTML]{000000} 20.0 & {\cellcolor[HTML]{FEE838}} \color[HTML]{000000} 20.0 & {\cellcolor[HTML]{FEE838}} \color[HTML]{000000} 20.0 & {\cellcolor[HTML]{FEE838}} \color[HTML]{000000} 20.0 \\
        GE & {\cellcolor[HTML]{FEE838}} \color[HTML]{000000} 20.0 & {\cellcolor[HTML]{FEE838}} \color[HTML]{000000} 20.0 & {\cellcolor[HTML]{FEE838}} \color[HTML]{000000} 20.0 & {\cellcolor[HTML]{FEE838}} \color[HTML]{000000} 20.0 & {\cellcolor[HTML]{FEE838}} \color[HTML]{000000} 20.0 \\
        GEN & {\cellcolor[HTML]{6F7073}} \color[HTML]{F1F1F1} 10.0 & {\cellcolor[HTML]{6F7073}} \color[HTML]{F1F1F1} 10.0 & {\cellcolor[HTML]{6F7073}} \color[HTML]{F1F1F1} 10.0 & {\cellcolor[HTML]{FEE838}} \color[HTML]{000000} 20.0 & {\cellcolor[HTML]{6F7073}} \color[HTML]{F1F1F1} 10.0 \\
        GradNorm & {\cellcolor[HTML]{FEE838}} \color[HTML]{000000} 20.0 & {\cellcolor[HTML]{FFFFFF}} N/A & {\cellcolor[HTML]{FEE838}} \color[HTML]{000000} 20.0 & {\cellcolor[HTML]{FEE838}} \color[HTML]{000000} 20.0 & {\cellcolor[HTML]{FEE838}} \color[HTML]{000000} 20.0 \\
\bottomrule
        \end{tabular}
    \end{subtable}
    \\[1em]
    \begin{subtable}[t]{0.48\textwidth}
        \centering
        \caption*{CIFAR-100}
        \hspace*{-1.5cm}
        \begin{tabular}{lccccc}
        \toprule
        CSF & base & class & class avg & class pred & global \\
        \midrule
        CTM & {\cellcolor[HTML]{B4A76F}} \color[HTML]{F1F1F1} 15.0 & {\cellcolor[HTML]{6F7073}} \color[HTML]{F1F1F1} 10.0 & {\cellcolor[HTML]{B4A76F}} \color[HTML]{F1F1F1} 15.0 & {\cellcolor[HTML]{B4A76F}} \color[HTML]{F1F1F1} 15.0 & {\cellcolor[HTML]{B4A76F}} \color[HTML]{F1F1F1} 15.0 \\
        Confidence & {\cellcolor[HTML]{6F7073}} \color[HTML]{F1F1F1} 10.0 & {\cellcolor[HTML]{FFFFFF}} N/A & {\cellcolor[HTML]{FFFFFF}} N/A & {\cellcolor[HTML]{FFFFFF}} N/A & {\cellcolor[HTML]{FFFFFF}} N/A \\
        Energy & {\cellcolor[HTML]{B4A76F}} \color[HTML]{F1F1F1} 15.0 & {\cellcolor[HTML]{B4A76F}} \color[HTML]{F1F1F1} 15.0 & {\cellcolor[HTML]{6F7073}} \color[HTML]{F1F1F1} 10.0 & {\cellcolor[HTML]{B4A76F}} \color[HTML]{F1F1F1} 15.0 & {\cellcolor[HTML]{B4A76F}} \color[HTML]{F1F1F1} 15.0 \\
        GE & {\cellcolor[HTML]{B4A76F}} \color[HTML]{F1F1F1} 15.0 & {\cellcolor[HTML]{B4A76F}} \color[HTML]{F1F1F1} 15.0 & {\cellcolor[HTML]{6F7073}} \color[HTML]{F1F1F1} 10.0 & {\cellcolor[HTML]{6F7073}} \color[HTML]{F1F1F1} 10.0 & {\cellcolor[HTML]{B4A76F}} \color[HTML]{F1F1F1} 15.0 \\
        GEN & {\cellcolor[HTML]{B4A76F}} \color[HTML]{F1F1F1} 15.0 & {\cellcolor[HTML]{B4A76F}} \color[HTML]{F1F1F1} 15.0 & {\cellcolor[HTML]{6F7073}} \color[HTML]{F1F1F1} 10.0 & {\cellcolor[HTML]{6F7073}} \color[HTML]{F1F1F1} 10.0 & {\cellcolor[HTML]{B4A76F}} \color[HTML]{F1F1F1} 15.0 \\
        GradNorm & {\cellcolor[HTML]{FEE838}} \color[HTML]{000000} 20.0 & {\cellcolor[HTML]{FFFFFF}} N/A & {\cellcolor[HTML]{FEE838}} \color[HTML]{000000} 20.0 & {\cellcolor[HTML]{FEE838}} \color[HTML]{000000} 20.0 & {\cellcolor[HTML]{FEE838}} \color[HTML]{000000} 20.0 \\
\bottomrule
        \end{tabular}
    \end{subtable}
    \hfill
    \begin{subtable}[t]{0.48\textwidth}
        \centering
        \caption*{TinyImageNet}
        \begin{tabular}{lccccc}
        \toprule
        CSF & base & class & class avg & class pred & global \\
        \midrule
        CTM & {\cellcolor[HTML]{B4A76F}} \color[HTML]{F1F1F1} 15.0 & {\cellcolor[HTML]{B4A76F}} \color[HTML]{F1F1F1} 15.0 & {\cellcolor[HTML]{B4A76F}} \color[HTML]{F1F1F1} 15.0 & {\cellcolor[HTML]{B4A76F}} \color[HTML]{F1F1F1} 15.0 & {\cellcolor[HTML]{B4A76F}} \color[HTML]{F1F1F1} 15.0 \\
        Confidence & {\cellcolor[HTML]{B4A76F}} \color[HTML]{F1F1F1} 15.0 & {\cellcolor[HTML]{FFFFFF}} N/A & {\cellcolor[HTML]{FFFFFF}} N/A & {\cellcolor[HTML]{FFFFFF}} N/A & {\cellcolor[HTML]{FFFFFF}} N/A \\
        Energy & {\cellcolor[HTML]{B4A76F}} \color[HTML]{F1F1F1} 15.0 & {\cellcolor[HTML]{B4A76F}} \color[HTML]{F1F1F1} 15.0 & {\cellcolor[HTML]{6F7073}} \color[HTML]{F1F1F1} 10.0 & {\cellcolor[HTML]{B4A76F}} \color[HTML]{F1F1F1} 15.0 & {\cellcolor[HTML]{B4A76F}} \color[HTML]{F1F1F1} 15.0 \\
        GE & {\cellcolor[HTML]{B4A76F}} \color[HTML]{F1F1F1} 15.0 & {\cellcolor[HTML]{B4A76F}} \color[HTML]{F1F1F1} 15.0 & {\cellcolor[HTML]{6F7073}} \color[HTML]{F1F1F1} 10.0 & {\cellcolor[HTML]{B4A76F}} \color[HTML]{F1F1F1} 15.0 & {\cellcolor[HTML]{B4A76F}} \color[HTML]{F1F1F1} 15.0 \\
        GEN & {\cellcolor[HTML]{B4A76F}} \color[HTML]{F1F1F1} 15.0 & {\cellcolor[HTML]{B4A76F}} \color[HTML]{F1F1F1} 15.0 & {\cellcolor[HTML]{6F7073}} \color[HTML]{F1F1F1} 10.0 & {\cellcolor[HTML]{B4A76F}} \color[HTML]{F1F1F1} 15.0 & {\cellcolor[HTML]{B4A76F}} \color[HTML]{F1F1F1} 15.0 \\
        GradNorm & {\cellcolor[HTML]{B4A76F}} \color[HTML]{F1F1F1} 15.0 & {\cellcolor[HTML]{FFFFFF}} N/A & {\cellcolor[HTML]{6F7073}} \color[HTML]{F1F1F1} 10.0 & {\cellcolor[HTML]{6F7073}} \color[HTML]{F1F1F1} 10.0 & {\cellcolor[HTML]{B4A76F}} \color[HTML]{F1F1F1} 15.0 \\
\bottomrule
        \end{tabular}
    \end{subtable}
\end{table}

\newpage

\section{Neural-Collapse-based Analysis}\label{appendix:neural_collapse}
    We provide a geometric interpretation of our empirical findings through the lens of Neural Collapse (NC), focusing on the last-layer feature space and classifier geometry. Taking into consideration the Definitions and Notations in Section~\ref{methods:def_notations}, we quantify proximity to NC using the metrics described by \citet{papyan2020prevalence}:
\begin{itemize}
    \item \textbf{Equinormness of the class means} ($\lVert\boldsymbol{\mu}_i\rVert=\lVert\boldsymbol{\mu}_j\rVert$): $\frac{\text{Std}_c(\lVert\tilde{\boldsymbol{\mu}}_c\rVert_2)}{\text{Avg}_c(\lVert\tilde{\boldsymbol{\mu}}_c\rVert_2)}$
    \item \textbf{Equinormness of the classifier weights} ($\lVert\boldsymbol{w}_i\rVert=\lVert\boldsymbol{w}_j\rVert$): $\frac{\text{Std}_c(\lVert\boldsymbol{w}_{c}\rVert_2)}{\text{Avg}_c(\lVert\boldsymbol{w}_{c}\rVert_2)}$
    \item \textbf{Equiangularity of the class means} ($\cos_{\boldsymbol{\mu}}(i,j)=\beta$): $\text{Std}_{c,c'\neq c}(\cos_{\boldsymbol{\mu}}(c,c'))$
    \item \textbf{Equiangularity of the classifier weights} ($\cos_{\boldsymbol{w}}(i,j)=\rho$): $\text{Std}_{c,c'\neq c}(\cos_{\boldsymbol{w}}(c,c'))$
    \item \textbf{Maximal Equiangularity of the class means} ($\cos_{\boldsymbol{\mu}}(i,j)=-\frac{1}{C-1}$): $\text{Avg}_{c,c'}\lvert \cos_{\boldsymbol{\mu}}(c,c')+\frac{1}{C-1}\rvert$
    \item \textbf{Maximal Equiangularity of the classifier weights} ($\cos_{\boldsymbol{w}}(i,j)=-\frac{1}{C-1}$): $\text{Avg}_{c,c'}\lvert \cos_{\boldsymbol{w}}(c,c')+\frac{1}{C-1}\rvert$
    \item \textbf{Within-class variation collapse} ($\boldsymbol{h}\to \boldsymbol{\mu}_k$): $\frac{1}{C^{2}}\text{Tr}\left(\boldsymbol{\Sigma_{\boldsymbol{W}}}\boldsymbol{\Sigma_{\boldsymbol{B}}^{\dagger}}\right)$, i.e., the NC1 metric of \citet{papyan2020prevalence}, $\text{Tr}(\boldsymbol{\Sigma_{\boldsymbol{W}}}\boldsymbol{\Sigma_{\boldsymbol{B}}^{\dagger}})/C$, divided by $C$ (see the note below)
    \item \textbf{Self-duality} ($\boldsymbol{w}_k=\alpha\boldsymbol{\mu}_k$): $\left\lVert \left(\frac{\boldsymbol{W}}{\lVert \boldsymbol{W} \rVert_F}\right)^T - \frac{\boldsymbol{M}}{\lVert \boldsymbol{M} \rVert_F} \right\rVert_F^2=\left \lVert \frac{1}{\lVert \boldsymbol{W} \rVert_F} \boldsymbol{W}^T  - \frac{1}{\lVert \boldsymbol{M} \rVert_F}\boldsymbol{M} \right \rVert_F^2= \left \lVert \frac{1}{\lVert \boldsymbol{W} \rVert_F} \boldsymbol{W}^T \right \rVert_F^2 +\left \lVert  \frac{1}{\lVert \boldsymbol{M} \rVert_F}\boldsymbol{M} \right \rVert_F^2-2\frac{\mathrm{Tr}(\boldsymbol{W}\boldsymbol{M})}{\lVert \boldsymbol{W} \rVert_F\lVert \boldsymbol{M} \rVert_F}   =2-2\frac{\mathrm{Tr}(\boldsymbol{W}\boldsymbol{M})}{\lVert \boldsymbol{W} \rVert_F\lVert \boldsymbol{M} \rVert_F} $
\end{itemize}
where $\tilde{\boldsymbol{\mu}}_{c} = \boldsymbol{\mu}_{c}-\boldsymbol{\mu}$ is the class-centered mean, $\cos_{\boldsymbol{\mu}}(c,c')=\frac{\tilde{\boldsymbol{\mu}}_{c}^T \tilde{\boldsymbol{\mu}}_{c'}}{\lVert\tilde{\boldsymbol{\mu}}_{c}\rVert_2\lVert\tilde{\boldsymbol{\mu}}_{c'}\rVert_2}$ is cosine similarity between any pair of class-centered means $c$ and $c'$, $\cos_{\boldsymbol{w}}(c,c')=\frac{\boldsymbol{w}_{c}^T \boldsymbol{w}_{c'}}{\lVert\boldsymbol{w}_{c}\rVert_2\lVert\boldsymbol{w}_{c'}\rVert_2}$ is cosine similarity between any pair of classifier weights $c$ and $c'$, $\boldsymbol{\Sigma}_{\boldsymbol{W}}=\text{Avg}_{i,c}(\boldsymbol{h}_{i,c}-\boldsymbol{\mu}_{c})(\boldsymbol{h}_{i,c}-\boldsymbol{\mu}_{c})^T$ is the within-class covariance, $\boldsymbol{\Sigma}_{\boldsymbol{B}}=\text{Avg}_c(\tilde{\boldsymbol{\mu}}_{c}{\tilde{\boldsymbol{\mu}}^T_{c}})$ is the between-class covariance, $\dagger$ is the Moore-Penrose pseudoinverse, and $\boldsymbol{M}=[\tilde{\boldsymbol{\mu}}_{c}:c=1,\dots,C]$ is the centered class-mean matrix.

\paragraph{Note on the within-class variation metric.} The reported value is Papyan's NC1 divided by the number of classes $C$ in every pool of the paper. For VGG-13, ResNet-18, and ViT we compute it with the default pseudoinverse cutoff of PyTorch in float32; for the self-supervised probes and the ImageNet models, with a relative tolerance of $10^{-6}$.

We compute these metrics for each model and dataset under three activation regimes: unfiltered, global PCA projection filtering, and class-predicted PCA projection filtering, and interpret how shifts in these metrics align with shifts in the relative performance of CSFs. Tables~\ref{tab:nc_activations_vgg} and~\ref{tab:nc_activations_ViT} show the NC metrics for all the activation regimes, considering CNN and ViT models respectively, and Table~\ref{tab:nc_weights} contains the NC metrics for the classifier weights.

\begin{table}[htbp]
    \scriptsize
    \centering
    \caption{Neural Collapse metrics computed on penultimate-layer activations (subtables (a) and (b)) and on classifier weights (subtable (c)). (a) Per-dataset NC metrics for VGG-13 models trained from scratch. (b) Per-dataset NC metrics for fine-tuned ViT models. (c) NC metrics computed on the classifier weights $\boldsymbol{W}$, reported per backbone and source dataset. Lower values indicate stronger collapse on each metric. The corresponding values under PCA projection filtering (Global and Class pred) are reported in Table~\ref{tab:nc_proj_main} of Appendix~\ref{appendix:projection_full}.}
    \label{tab:nc_main}
    \begin{subtable}[t]{\textwidth}
        \centering
        \subcaption{NC metrics for VGG-13 models.}
        \label{tab:nc_activations_vgg}
        \begin{tabular}{lrrrrr}
\toprule
Dataset & EqNorm & EqAng & max-EqAng & Var Collapse & Self Duality \\
\midrule
CIFAR-10 & {\cellcolor[HTML]{452929}} \color[HTML]{F1F1F1} 0.0266 & {\cellcolor[HTML]{1E0000}} \color[HTML]{F1F1F1} 0.0651 & {\cellcolor[HTML]{FFFFFF}} \color[HTML]{000000} 0.2447 & {\cellcolor[HTML]{A66C6C}} \color[HTML]{F1F1F1} 0.0081 & {\cellcolor[HTML]{2D1313}} \color[HTML]{F1F1F1} 0.0320 \\
SuperCIFAR-100 & {\cellcolor[HTML]{1E0000}} \color[HTML]{F1F1F1} 0.0246 & {\cellcolor[HTML]{573636}} \color[HTML]{F1F1F1} 0.0678 & {\cellcolor[HTML]{CC9F8D}} \color[HTML]{F1F1F1} 0.1537 & {\cellcolor[HTML]{B77777}} \color[HTML]{F1F1F1} 0.0095 & {\cellcolor[HTML]{966161}} \color[HTML]{F1F1F1} 0.0519 \\
CIFAR-100 & {\cellcolor[HTML]{FFFFFE}} \color[HTML]{000000} 0.0743 & {\cellcolor[HTML]{FFFFFE}} \color[HTML]{000000} 0.1049 & {\cellcolor[HTML]{6D4545}} \color[HTML]{F1F1F1} 0.0950 & {\cellcolor[HTML]{A36969}} \color[HTML]{F1F1F1} 0.0079 & {\cellcolor[HTML]{FFFFFF}} \color[HTML]{000000} 0.1279 \\
TinyImageNet & {\cellcolor[HTML]{DAC5A1}} \color[HTML]{000000} 0.0544 & {\cellcolor[HTML]{DDCCA4}} \color[HTML]{000000} 0.0900 & {\cellcolor[HTML]{1E0000}} \color[HTML]{F1F1F1} 0.0761 & {\cellcolor[HTML]{351D1D}} \color[HTML]{F1F1F1} 0.0021 & {\cellcolor[HTML]{2D1313}} \color[HTML]{F1F1F1} 0.0322 \\
\bottomrule
\end{tabular}
    \end{subtable}
    \hfill
    \begin{subtable}[t]{\textwidth}
        \centering
        \subcaption{NC metrics for ViT models.}
        \label{tab:nc_activations_ViT}
        \begin{tabular}{lrrrrr}
\toprule
Dataset & EqNorm & EqAng & max-EqAng & Var Collapse & Self Duality \\
\midrule
CIFAR-10 & {\cellcolor[HTML]{1E0000}} \color[HTML]{F1F1F1} 0.0869 & {\cellcolor[HTML]{FFFFFF}} \color[HTML]{000000} 0.2505 & {\cellcolor[HTML]{FFFFFF}} \color[HTML]{000000} 0.3770 & {\cellcolor[HTML]{321A1A}} \color[HTML]{F1F1F1} 0.0120 & {\cellcolor[HTML]{FFFFFE}} \color[HTML]{000000} 1.2692 \\
SuperCIFAR-100 & {\cellcolor[HTML]{CDA18E}} \color[HTML]{F1F1F1} 0.1432 & {\cellcolor[HTML]{F8F8E8}} \color[HTML]{000000} 0.2403 & {\cellcolor[HTML]{DDCEA5}} \color[HTML]{000000} 0.2814 & {\cellcolor[HTML]{553434}} \color[HTML]{F1F1F1} 0.0152 & {\cellcolor[HTML]{D8BF9E}} \color[HTML]{000000} 1.0922 \\
CIFAR-100 & {\cellcolor[HTML]{7F5151}} \color[HTML]{F1F1F1} 0.1054 & {\cellcolor[HTML]{1E0000}} \color[HTML]{F1F1F1} 0.1322 & {\cellcolor[HTML]{1E0000}} \color[HTML]{F1F1F1} 0.1172 & {\cellcolor[HTML]{391F1F}} \color[HTML]{F1F1F1} 0.0124 & {\cellcolor[HTML]{1E0000}} \color[HTML]{F1F1F1} 0.8535 \\
TinyImageNet & {\cellcolor[HTML]{F7F7E5}} \color[HTML]{000000} 0.1950 & {\cellcolor[HTML]{D6BC9C}} \color[HTML]{000000} 0.1984 & {\cellcolor[HTML]{8A5858}} \color[HTML]{F1F1F1} 0.1647 & {\cellcolor[HTML]{C38480}} \color[HTML]{F1F1F1} 0.0372 & {\cellcolor[HTML]{E0D5A9}} \color[HTML]{000000} 1.1292 \\
\bottomrule
\end{tabular}
    \end{subtable}
    \hfill
    \begin{subtable}[t]{\textwidth}
        \centering
        \subcaption{NC metrics for classifier weights $\boldsymbol{w}$.}
        \label{tab:nc_weights}
        \begin{tabular}{llrrr}
\toprule
Model & Dataset & EqNorm & EqAng & max-EqAng  \\
\midrule
\multirow[c]{4}{*}{VGG-13} & CIFAR-10 & {\cellcolor[HTML]{230606}} \color[HTML]{F1F1F1} 0.0143 & {\cellcolor[HTML]{F6F6E4}} \color[HTML]{000000} 0.0707 & {\cellcolor[HTML]{F0F0CF}} \color[HTML]{000000} 0.2455 \\
 & SuperCIFAR-100 & {\cellcolor[HTML]{5B3939}} \color[HTML]{F1F1F1} 0.0184 & {\cellcolor[HTML]{C99789}} \color[HTML]{F1F1F1} 0.0529 & {\cellcolor[HTML]{C78F86}} \color[HTML]{F1F1F1} 0.1414 \\
 & CIFAR-100 & {\cellcolor[HTML]{FFFFFF}} \color[HTML]{000000} 0.0730 & {\cellcolor[HTML]{DFD3A8}} \color[HTML]{000000} 0.0611 & {\cellcolor[HTML]{6D4545}} \color[HTML]{F1F1F1} 0.0665 \\
 & TinyImageNet & {\cellcolor[HTML]{8B5959}} \color[HTML]{F1F1F1} 0.0250 & {\cellcolor[HTML]{FFFFFF}} \color[HTML]{000000} 0.0748 & {\cellcolor[HTML]{6D4545}} \color[HTML]{F1F1F1} 0.0666 \\
\midrule
\multirow[c]{4}{*}{ViT} & CIFAR-10 & {\cellcolor[HTML]{1E0000}} \color[HTML]{F1F1F1} 0.0141 & {\cellcolor[HTML]{402525}} \color[HTML]{F1F1F1} 0.0372 & {\cellcolor[HTML]{FFFFFF}} \color[HTML]{000000} 0.2884 \\
 & SuperCIFAR-100 & {\cellcolor[HTML]{764B4B}} \color[HTML]{F1F1F1} 0.0217 & {\cellcolor[HTML]{1E0000}} \color[HTML]{F1F1F1} 0.0360 & {\cellcolor[HTML]{CB9E8C}} \color[HTML]{F1F1F1} 0.1526 \\
 & CIFAR-100 & {\cellcolor[HTML]{996262}} \color[HTML]{F1F1F1} 0.0274 & {\cellcolor[HTML]{9B6464}} \color[HTML]{F1F1F1} 0.0450 & {\cellcolor[HTML]{553434}} \color[HTML]{F1F1F1} 0.0551 \\
 & TinyImageNet & {\cellcolor[HTML]{553434}} \color[HTML]{F1F1F1} 0.0178 & {\cellcolor[HTML]{280D0D}} \color[HTML]{F1F1F1} 0.0364 & {\cellcolor[HTML]{1E0000}} \color[HTML]{F1F1F1} 0.0390 \\
\bottomrule
\end{tabular}
    \end{subtable}
\end{table}

\subsection{Per-dataset case studies}
\subsubsection{Why is CTM the best-performing CSF when the model has been trained on TinyImageNet using CNNs?}
As Table~\ref{tab:nc_activations_vgg} shows, Maximal Equiangularity, Variability Collapse, and Self Duality are the strongest NC metrics when TinyImageNet is used to train the VGG-13 models, while Equinormness and Equiangularity are the weakest. In these conditions prototype-alignment becomes highly reliable, which can be exploited by the CTM score. When Self-Duality occurs the classifiers align with the class means up to a scalar : $\boldsymbol{W} \propto \boldsymbol{M}^\top \implies \boldsymbol{w}_c = \alpha \boldsymbol{\mu}_c$, then  $\text{CTM}(\boldsymbol{x})=\max_k \text{sim}(\boldsymbol{w}_k, \boldsymbol{h})=\max_k\frac{ \boldsymbol{w}_k^{\top} \boldsymbol{h} }{\|\boldsymbol{w}_k\|_{2} \|\boldsymbol{h}\|_{2}} = \max_k\frac{ \alpha\boldsymbol{\mu}_k^{\top} \boldsymbol{h} }{\|\alpha\boldsymbol{\mu}_k\|_{2} \|\boldsymbol{h}\|_{2}}$. If Variability Collapse is achieved $\boldsymbol{\Sigma}_{\boldsymbol{W}} \to 0$, meaning features $\boldsymbol{x}$ collapse to their class means $\boldsymbol{\mu}_k$: $\boldsymbol{h} \to \boldsymbol{\mu}_k$. Therefore, $\text{CTM}(\boldsymbol{x})=\max_k\frac{ \boldsymbol{\mu}_k^{\top} \boldsymbol{h} }{\|\boldsymbol{\mu}_k\|_{2} \|\boldsymbol{h}\|_{2}} \approx \frac{ \boldsymbol{\mu}_k^{\top} \boldsymbol{\mu}_k}{\|\boldsymbol{\mu}_k\|_{2} \|\boldsymbol{\mu}_k\|_{2}} = 1$, which creates a deterministic score for ID samples. Even if Equiangularity is not achieved completely (meaning the gap between Class A and Class B is different from Class A and Class C), the CTM score for an ID input is approximately 1. In the other hand, Maximal Equiangularity dictates that class means form a Simplex ETF, maximizing the separation angle $\theta_{ij}$ between any distinct classes $i, j$: $\cos(\boldsymbol{\mu}_i, \boldsymbol{\mu}_j) = -\frac{1}{K-1} \quad \forall i \neq j$. This implies that the collapsed ID feature space is maximally sparse in terms of angular distribution. For an OOD sample $\boldsymbol{x}^{\text{OOD}}$ lying in the subspace orthogonal to the ID manifold (or between vertices), the maximum cosine similarity is bounded by the geometry of the simplex. Unlike dense feature spaces where an OOD point might accidentally align with a cluster, the ETF geometry ensures wide angular gaps (negative correlations) between prototypes, statistically minimizing $\max_k \cos(\boldsymbol{w}_k, \boldsymbol{h})$. OOD samples, which lie in the angular gaps, will strictly have scores $< 1$ as long as Maximal Angularity holds (the vectors point in distinct directions). CTM measures alignment with prototype: $\boldsymbol{h}^{\text{OOD}} \in \text{Null Space} \implies \cos(\boldsymbol{\mu}_k, \boldsymbol{h}^{\text{OOD}}) \approx 0$. An OOD sample cannot be perfectly parallel to any prototype, so $\cos(\boldsymbol{\mu}_k, \boldsymbol{h}^{\text{OOD}}) \ll 1$. In summary, CTM strips away the noise of magnitude (Equinorm failure) and relies on the purity of direction (Maximal Angularity) and clustering (Variability Collapse).

\subsubsection{Why Energy, MLS and NNGuide perform the best as CSFs when the model has been trained on CIFAR-10 and SuperCIFAR-100 using CNNs?}
In the context of CIFAR-10, Equinormness, Equiangularity, Variability Collapse, and Self-Duality emerge as the most robust metrics, whereas Maximal Equiangularity demonstrates the least adherence to a collapsed regime, as Table~\ref{tab:nc_activations_vgg} exhibits. SuperCIFAR-100 shows a similar metrics profile to CIFAR-10 with an improved Maximal Equiangularity, but slight deterioration in Self-Duality. In both NC regimes, logit-derived scores such as Energy and MLS benefit from stable, class-unbiased logit scaling induced by Equinormness and the differentiated target vs. non-target angular margin implied by Equiangular geometry. More specifically, Self-Duality implies the classifiers weights $\boldsymbol{W}$ align with the class means $\boldsymbol{M}$ up to a scalar $\alpha$: $\boldsymbol{W} \propto \boldsymbol{M}^\top \implies \boldsymbol{w}_c = \alpha \boldsymbol{\mu}_c$ and Equinorm states that any pair of class means $c$ and $c'$ have equal $\ell_2$ norm: $\|\boldsymbol{\mu}_c\|_2 = \|\boldsymbol{\mu}_{c'}\|_2 = R$. Substituting these into the logit $g(\boldsymbol{h})_{k}$ for a correctly classified ID sample (target logit) $\boldsymbol{h} \to \boldsymbol{\mu}_c$: $g(\boldsymbol{h})_{k} = \boldsymbol{w}_k^\top \boldsymbol{h} + b_{k} \approx \boldsymbol{w}_k^\top \boldsymbol{\mu}_k = \alpha \|\boldsymbol{\mu}_k\|^2 = \alpha R^2$. Thus, the minimum possible energy for ID data is uniform across all classes: $\text{Energy}(\boldsymbol{x}) \approx - \alpha R^2$. This uniformity prevents class-conditional bias, where some ID classes might otherwise have naturally higher energy (and thus higher False Positive Rates) than others due to varying feature norms. Equiangularity dictates that any pair of class means are equally spaced $\cos_{\boldsymbol{\mu}}(i, j) = \beta,~\forall i \neq j$, meaning that for an off-target logit $g(\boldsymbol{h})_{j} = \boldsymbol{w}_j^\top \boldsymbol{h} + b_{k} \approx \boldsymbol{w}_j^\top \boldsymbol{\mu}_k = \alpha \|\boldsymbol{\mu}_k\|^2 \beta = \alpha R^2 \beta$. 

Let's analyze the behavior of the detection scores under these conditions. The Energy score sums over the exponentiated logits $\sum e^{g(\boldsymbol{h})_k} = e^{\alpha R^2} + (K-1)e^{\alpha R^2 \beta}$,  which in the optimal case (Maximal Equiangularity $\beta=-\frac{1}{C-1}$) would be dominated by $e^{\alpha R^2}$ since $e^{\alpha R^2}\gg (C-1)e^{\alpha R^2 \beta}$ or $\alpha R^2 >\frac{\ln (C-1)}{1-\beta}$. This last expression reveals that $\beta$ allows some tolerance to make the target logit distinguishable from the off-target logits as $\beta<1-\frac{\ln (C-1)}{\alpha R^2}$ when evaluating ID inputs. This explains why the Energy score is still very effective at detecting OOD inputs in this scenario where Maximal Equiangularity is not the strongest.

However, a fundamental failure mode for logit-based scores emerges in the far-OOD regime. Both Energy and MLS are sensitive to feature magnitude ($\text{Energy} \propto \|h\|, \text{MLS} \propto \|h\|$); if an OOD sample has an anomalously large norm, it can produce a high confidence score regardless of its angular alignment. In contrast, geometric scores like CTM and fDBD are norm-invariant. Our top-clique results show that while Energy/MLS excel in near-OOD where norms are comparable to ID, they are consistently replaced by CTM/fDBD in far-OOD. This transition confirms that atypical feature magnitude is the primary confounder for logit-based detectors under strong semantic shifts, whereas angular consistency remains a robust signal. Assuming that the class means are not clustered together, the failure mode of MLS lies on its dependence on feature magnitude. Given the similar performance of the Energy score and MLS when CIFAR-10 is the training set, we argue that OOD norms can be differentiated from ID norms in this case. Model trained SuperCIFAR-100, in the other hand, seems to be more vulnerable to atypical large norms, specially when far OOD datasets are tested.

\subsubsection{Why NNGuide performs well when the model has been trained on CIFAR-10 using CNNs?}
NNGuide acts as a confidence-weighted geometric filter that corrects the summation bias inherent in the Energy score when classes are clustered (non-maximal equiangularity). In this regime, the Energy score could become unreliable because the summation aggregates multiple medium-strength logits from clustered classes. If an OOD sample lies in the geometric center of a cluster with cosine similarity $\beta > 0$, it activates all $K$ classifiers moderately, resulting in a partition sum that can rival the single high-confidence logit of an ID sample, leading to false positives. NNGuide modulates this by multiplying the base score with a guidance term $G(\boldsymbol{h})$, computed from the $k$ nearest training neighbors:$\text{NNGuide}(\boldsymbol{x}) =  S_{\text{base}}(\boldsymbol{h})\cdot G(\boldsymbol{h}) = S_{\text{base}}(\boldsymbol{h}) \cdot \left( \frac{1}{k}\sum_{i=1}^k S_{\text{base}}(\boldsymbol{h}_{(i)}) \cdot \cos(\boldsymbol{h},\boldsymbol{h}_{(i)}) \right)$. This term leverages the Self-Duality of Neural Collapse, where the nearest neighbors $\boldsymbol{h}_{(i)}$ for valid samples are high-confidence prototypes ($S_{\text{base}}(\boldsymbol{h}_{(i)}) \approx S_{\max}$), effectively turning $G(\boldsymbol{h})$ into a clean geometric penalty.

NNGuide succeeds in the clustered regime ($\beta > 0$) because the guidance term enforces an angular margin that the raw Energy score ignores. For an ID sample aligned with its class direction, the cosine similarity is near 1, yielding a guidance factor of $G(\boldsymbol{h}) \approx S_{max} \cdot 1$. Conversely, for an OOD sample in the angular gap between clustered vectors, the cosine similarity to the nearest neighbors is bounded by the cluster geometry: $G(\boldsymbol{h}^{\text{OOD}}) \approx S_{\max} \cdot \beta$. Even if the base Energy scores are indistinguishable ($S_{\text{base}}(\boldsymbol{h}^{\text{OOD}}) \approx S_{\text{base}}(\boldsymbol{h})$) due to summation explosion, NNGuide suppresses the OOD score by the factor $\beta$: $\frac{\text{NNGuide}(\boldsymbol{x}^{\text{OOD}})}{\text{NNGuide}(\boldsymbol{x})} \approx \frac{S_{\text{base}}(\boldsymbol{h}^{\text{OOD}}) \cdot \beta}{S_{\text{base}}(\boldsymbol{h}) \cdot 1} = \beta$. As long as the classes are not perfectly overlapping ($\beta < 1$), this multiplicative correction restores the separability that the base score lost. However, NNGuide succumbs to failure modes where this geometric penalty is either insufficient or overridden. The first mode is Tight Clustering ($\beta \to 1$), where classes become so semantically similar that the angular margin vanishes; here, the penalty factor $\beta$ approaches 1, rendering the guidance term useless ($G(\boldsymbol{h}^{\text{OOD}}) \approx G(\boldsymbol{h})$). The second and more critical failure mode is its sensitivity to feature magnitude. While the guidance term $G(\boldsymbol{h})$ is scale-invariant (due to cosine normalization), the base term $S_{\text{base}}(\boldsymbol{h})$ scales linearly with the feature norm $\lVert \boldsymbol{h}\rVert$. If an OOD sample has an anomalously large magnitude where $\lVert \boldsymbol{h}^{\text{OOD}}\rVert= \gamma\lVert \boldsymbol{h}\rVert$ with $\gamma>1$, the linear growth of the base score overrides the constant geometric penalty: $\text{NNGuide}(\boldsymbol{x}^{\text{OOD}}) \approx (\gamma S_{\text{base}}) \cdot (S_{\max} \beta)$. 

\subsubsection{Why fDBD performs well when the model has been trained on CIFAR-100 using CNNs?}
The NC metrics for CIFAR-100 report the weakest adherence to a collapsed regime compared to the other datasets, with Maximal Equiangularity for class means and for classifier weights being the strongest metrics. The lack of a strong Variability Collapse, however, makes it difficult for CTM to be competitive in this case. fDBD succeeds because it exploits the rigid geometric skeleton of the classifier weights to create a robust angular detector that is tolerant of feature noise. The fDBD score is defined as the average distance to all non-predicted classes, regularized by the feature distance to the mean of training features: $\text{fDBD}(\boldsymbol{x}) = \frac{1}{|\mathcal{C}| - 1} \sum_{c \neq f(x)} \frac{D_m(\boldsymbol{h}, c)}{\|\boldsymbol{h} - \boldsymbol{\mu}_{\text{train}}\|_2}$, where $D_{m}(\boldsymbol{h}, j) = \frac{|(\boldsymbol{w}_{m(\boldsymbol{x})} - \boldsymbol{w}_j)^\top \boldsymbol{h} + (b_{m(\boldsymbol{x})} - b_j)|}{\|\boldsymbol{w}_{m(\boldsymbol{x})} - \boldsymbol{w}_j\|_2}$. In this regime fDBD outperforms prototype CSFs  because it measures the true safety margin relative to the decision boundary, whereas prototype CSFs measure centrality relative to a potentially unsafe centroid. The sumation of the fDBD score is proportional to the projection of the mean onto the weight difference vectors: $\text{fDBD}(\boldsymbol{x}) \propto \sum_{j \neq k} (\boldsymbol{w}_k - \boldsymbol{w}_j)^\top \boldsymbol{h}$. When Self-Duality fails, the geometric center ($\boldsymbol{\mu}_k$) and the decision center ($\boldsymbol{w}_k$) diverge. If $\boldsymbol{\mu}_k$ is close to the boundary (small projection onto $\boldsymbol{w}_k - \boldsymbol{w}_j$), fDBD correctly reports a low score, identifying the higher risk of misclassification that CTM would ignore in this scenario. Even with misalignment, Maximal Angularity of $\boldsymbol{W}$ ensures the decision region is a wide cone supported by the collective opposition of all other classes. Using the ETF property ($\sum_{j \neq k} \boldsymbol{w}_j = -\boldsymbol{w}_k$), the fDBD summation simplifies: $\sum_{j \neq k} (\boldsymbol{w}_k - \boldsymbol{w}_j)^\top \boldsymbol{h} = (C-1)\boldsymbol{w}_k^\top \boldsymbol{h} - (\sum_{j \neq k} \boldsymbol{w}_j)^\top \boldsymbol{h} = C (\boldsymbol{w}_k^\top \boldsymbol{h})$. As long as the classifier is accurate (meaning $\boldsymbol{w}_k^\top \boldsymbol{h} > 0$), the ETF geometry amplifies the margin by a factor of $C$. fDBD captures this collective margin, ensuring that even a misaligned ID sample is distinguished from OOD samples where $\boldsymbol{w}_k^\top \boldsymbol{h}^{\text{OOD}} \approx 0$. The regularization term improves the evaluation when atypically feature norms are atypically large.

\paragraph{Note on fine-tuned ViTs.}
Unlike CNNs trained from scratch, ViTs that were pretrained and finetuned might not have null space for OOD inputs in many cases. Since the ViT networks were pretrained using ImageNet1K, and finetuned using dataset with fewer number of classes, it is likely that the OOD inputs would be projected to existing subspaces that were created during pretraining. This can be corroborated by the NC metrics of ViT models compared to the NC metrics of VGG models. ViT's NC metrics are higher than their CNN's counterparts. Weaker collapse of Transformers than of CNNs on CIFAR-scale data has also been reported for models trained from scratch~\cite{zhou2025all}. When the pretraining data has more classes than the finetuning data, then the finetuning process would readjust the boundaries around the classes are the most similar, eg. if the finetuning data contains the class dog, and the pretraining data has classes dog, fox, wolf, and coyotes, then the boundaries after finetuning would enclose all these classes. However when it comes to OOD detection, the strategies cannot rely on the same approaches as models trained from scratch. The corresponding NC values under PCA projection filtering and the mechanistic analyses of how Global and Class-Predicted Projection reshape ViT geometry are reported in Appendix~\ref{appendix:projection_full}.

\section{Training Paradigms, CSF Baselines and Variations}\label{appendix:cfs_variations}

\subsection{Computing Infrastructure}\label{appendix:compute}
The main experiments were executed on an internal GPU cluster with two GPU types running in parallel: NVIDIA T4 GPUs for CNN-based work (VGG-13 and the held-out ResNet-18 pool) and NVIDIA A100 GPUs for ViT inference. We did not mix GPU types within an experiment, and every job for a given backbone used the same GPU class to avoid hardware-induced variance. The CSF inference, projection-filtering, neural-collapse, and statistical pipelines were dispatched as parallel jobs across 12 CPU workers on the same cluster.

\paragraph{Checkpoint inventory.}
Checkpoints span four source datasets, the three reported training paradigms (ConfidNet, DeVries, Deep~Gamblers), and the additional within-paradigm factors discussed in Section~\ref{methods:def_notations} (dropout setting on/off; Deep~Gamblers reservation reward $o$; per-cell seeds). For VGG-13 the per-source counts are CIFAR-10 / CIFAR-100 / SuperCIFAR-100 / TinyImageNet $=$ 60 / 70 / 90 / 60, totalling 280 checkpoints. For the fine-tuned ViT pool the per-source count is 10 (40 total). For the held-out ResNet-18 pool the same structure is reproduced with one seed per cell, giving 12 / 14 / 18 / 12 checkpoints per source, 56 in total.

\paragraph{Training compute used in this paper.}
The VGG-13 and fine-tuned-ViT checkpoints on CIFAR-10, CIFAR-100, and SuperCIFAR-100 were inherited from the FD-Shifts project~\cite{jaeger2022call,traub2024overcoming} and not retrained here. FD-Shifts does not ship TinyImageNet checkpoints, so those 60 VGG-13 and 10 ViT models were trained with the FD-Shifts code and are released with this paper; the total below does not include them. The ResNet-18 pool (56 trainings, see above) was trained for this paper at approximately 2.5 hours per training on a single T4, for a total of about 140 GPU-hours of training compute attributable to this paper.

\paragraph{CSF fitting and inference compute.}
CSF fitting and inference are run on every checkpoint regardless of who trained it, so the work is done on all $280 + 40 + 56 = 376$ checkpoints. For every (checkpoint, dataset) pair, NC metric and CSF feature extraction takes approximately 2 minutes (5 minutes on TinyImageNet), and per-CSF inference takes under 1 minute on most CSFs. The PCA and KPCA reconstruction-error CSFs require an additional fitting phase that takes about 20 minutes on CIFAR-100 and TinyImageNet, the two source datasets with the largest class counts. CLIP-based OOD aggregation (Appendix~\ref{appendix:clip}) runs once per dataset in roughly 10 minutes. Aggregated, the inference and analysis pipelines consume on the order of several hundred GPU-hours; with two GPUs running in parallel and 12 CPU workers, total wall-clock for the full sweep is one to two weeks.

\paragraph{Compute of the other experiments.}
Maha++, NCI, ReAct, and ASH-S use the same pipeline at the per-CSF cost above. The self-supervised probes (Appendix~\ref{appendix:ssl}) need no backbone training: feature extraction takes a few GPU-hours on one GPU, and the probes and analyses run on CPU. The ImageNet sweep (Appendix~\ref{appendix:imagenet}; 47 pretrained models, no training) takes about 13 GPU-hours on one GPU, and the CLIP regimes take one pass over its OOD suite on an NVIDIA V100. The predictor heads, the rank pipeline, and the sensitivity analyses run on CPU.

    \subsection{Training Paradigms}

\subsubsection{ConfidNet (regressing the true-class probability).}
Let $f_{\mathcal{W}}=g\circ h$ be a trained classifier with weights $\mathcal{W}$, logits $f(\boldsymbol{x})\in\mathbb{R}^C$ and softmax $p_k(\boldsymbol{x})=\exp(f_k(\boldsymbol{x}))/\sum_j \exp(f_j(\boldsymbol{x}))$. The standard confidence proxy is the maximum class probability,
$\mathrm{MSR}(\boldsymbol{x})\;=\;\max_{k\in\{1,\dots,C\}} p_k(\boldsymbol{x})$,
but this quantity can be spuriously large for both correct and erroneous predictions, hampering failure detection. ConfidNet replaces $\mathrm{MSR}$ with the \emph{true-class probability} (TCP), $\mathrm{TCP}(\boldsymbol{x},y)=p_y(\boldsymbol{x}),$ which is typically high for correct predictions and low for misclassifications \cite{corbiere2019addressing,corbiere2021confidence}. Because $y$ is unknown at test time, ConfidNet learns an auxiliary regressor $s_{\mathcal{W}^{\mathrm{conf}}}:\mathcal{X}\to[0,1]$ that \emph{predicts} $\mathrm{TCP}(\boldsymbol{x},y)$ from features of the trained classifier $f_{\mathcal{W}}$.

Denote by $E=h$ the encoder of $f_{\mathcal{W}}$ and by $g_{\mathcal{W}^{\mathrm{conf}}}:\mathbb{R}^D\to[0,1]$ a small MLP head. ConfidNet’s score is
$s_{\mathcal{W}^{\mathrm{conf}}}(\boldsymbol{x})=\big(g_{\mathcal{W}^{\mathrm{conf}}}\circ E\big)(\boldsymbol{x})\approx\mathrm{TCP}(\boldsymbol{x},y),$ trained on the labeled training set $\mathcal{D}_{\mathrm{train}}$ with a mean squared error loss
$\mathcal{L}_{\mathrm{conf}}(\mathcal{W}^{\mathrm{conf}};\mathcal{D}_{\mathrm{train}})=\frac{1}{|\mathcal{D}_{\mathrm{train}}|}\sum_{(\boldsymbol{x}_i,y_i)\in\mathcal{D}_{\mathrm{train}}}\Big(s_{\mathcal{W}^{\mathrm{conf}}}(\boldsymbol{x}_i)-\mathrm{TCP}(\boldsymbol{x}_i,y_i)\Big)^2.$ This \emph{post-hoc} module transfers knowledge from the classifier’s encoder and yields a calibrated, label-informed confidence score at inference. It should be noted that the weights $\mathcal{W}$ are kept frozen when training $g_{\mathcal{W}^{\mathrm{conf}}}$. In practice, ConfidNet can be viewed as a supervised alternative to $\mathrm{MSR}$ that aligns the confidence target with the Bayes-relevant quantity $\mathrm{TCP}$; see \citet{corbiere2019addressing,corbiere2021confidence} for more details on the implementation.

\subsubsection{DeVries \& Taylor: learned confidences via target–prediction interpolation.}
Given a classifier $f_{\mathcal W}=g\circ h$ with softmax output $p(x)\in\Delta^{C-1}$, \citet{devries2018learning} add a \emph{confidence branch} in parallel to the class head. The branch shares features with the penultimate layer and outputs $s_{\mathcal W^{\mathrm{devries}}}(\boldsymbol{x})=\mathrm{softmax}\big(\mathbf w^\top h(\boldsymbol{x})+b\big)\in[0,1],$ interpreted as the network’s confidence that it can correctly predict the label of $\boldsymbol{x}$. During training, the CSF provides the network with \textit{hints} by interpolating between its own prediction and the one-hot target $\boldsymbol{y}$ according to $s$: $p'(\boldsymbol{x},\boldsymbol{y})=s_{\mathcal W^{\mathrm{devries}}}(\boldsymbol{x})\,p(\boldsymbol{x})+\big(1-s_{\mathcal W^{\mathrm{devries}}}(\boldsymbol{x})\big)\,\boldsymbol{y}.$ The task loss is then computed on $p'$ rather than on $p$, e.g. the negative log-likelihood $\mathcal L_{\text{task}}(\mathcal W^{\mathrm{devries}},\mathcal W)=-\frac{1}{|\mathcal D_{\text{train}}|}\sum_{(\boldsymbol{x}_i,y_i)} \log \big([p'(\boldsymbol{x}_i,\boldsymbol{y}_i)]_{y_i}\big).$ Intuitively, when the model is \emph{uncertain} ($s\!\approx\!0$), it is allowed to rely more on the target distribution (a hint); when it is \emph{confident} ($s\!\approx\!1$), it must stand by its own prediction.

To avoid the degenerate solution $s=0$ (which would always copy the target), a confidence regularizer encourages high $s$ values: $\mathcal L_{\text{devries}}(\mathcal W^{\mathrm{devries}})=-\frac{1}{|\mathcal D_{\text{train}}|}\sum_{\boldsymbol{x}_i}\log s_{\mathcal W^{\mathrm{devries}}}(\boldsymbol{x}_i),~\mathcal L_{\text{total}}=\mathcal L_{\text{task}} + \lambda\,\mathcal L_{\text{devries}},$ with $\lambda>0$ balancing reliance on hints versus self-prediction. The resulting balancing where copying the target when the classifier struggles vs. being penalized for low confidence drives $s(\boldsymbol{x})$ to be large where the model is accurate and small where it is prone to error. At test time, the class prediction uses $p(\boldsymbol{x})$ from the base head, while $s(\boldsymbol{x})$ serves as a \emph{confidence score} for failure/OOD detection; no hints are used at inference. See \citet{devries2018learning} for more details on the implementation.

\subsubsection{Deep Gamblers: abstention via a \texorpdfstring{$(C{+}1)$}{(C+1)}-st class and a reward parameter.}
\citet{liu2019deep} cast classification as a Kelly-style gambling game~\cite{cover2006elements}. For a $C$-class task, the network predicts over $C{+}1$ outcomes. This means that an additional \emph{abstain} option is attached to the original $C$ labels. Let $f_{\mathcal W}=g\circ h$ with logits $f(\boldsymbol{x})\in\mathbb{R}^{C+1}$, $p(\boldsymbol{x})=\mathrm{softmax}(f(\boldsymbol{x}))\in\Delta^{C}$,
and write $p_{k}(\boldsymbol{x})$ for class $k\!\le\!C$ and $p_{C+1}(\boldsymbol{x})$ for \emph{abstain}. The training objective maximizes the expected ($\log$) wealth in a horse-race with \emph{reservation}: predicting the true class $y$ yields payoff $o>0$ (a tunable \emph{reward}), and abstaining yields payoff $1$. This leads to the loss $\mathcal{L}_{\text{DG}}(\mathcal W; \mathcal D_{\text{train}}, o)
~:=~
-\frac{1}{|\mathcal D_{\text{train}}|}
\sum_{(\boldsymbol{x}_i,y_i)\in\mathcal D_{\text{train}}}
\log\!\big(o\,p_{y_i}(\boldsymbol{x}_i) + p_{C+1}(\boldsymbol{x}_i)\big).$

When $p_{C+1}\!=\!0$ (no abstention), $\mathcal{L}_{\text{DG}}$ reduces to cross-entropy up to an additive constant (since $\log o$ adds to the true-class logit). The head is linear, $g(z)=Wz+b,\; W\in\mathbb{R}^{(C+1)\times D},\, b\in\mathbb{R}^{C+1},$
so the CSF is a drop-in replacement for a standard classifier.

At test time, the model accepts an output if its (reward-weighted) probability dominates abstention, and rejects otherwise. A sufficient decision rule is $\textsf{accept}~\Leftrightarrow~ o\cdot \max_{k\le C} p_k(x)\;>\;p_{C+1}(x), ~\textsf{reject}~\text{(abstain)}~\text{otherwise}.$ Hence, larger $o$ discourages abstention (the classifier must be more confident to reject), while smaller $o$ encourages it. $o$ is tuned on validation to meet a desired risk-coverage trade-off. For failure/OOD detection one can use the abstention mass as a score, $s_{\text{DG}}(x)=p_{C+1}(\boldsymbol{x})$ (higher $\Rightarrow$ more likely atypical), or a margin-like variant $s_{\text{DG}}(x)=p_{C+1}(x)-o\!\cdot\!\max_{k\le C} p_k(x)$
(negative $\Rightarrow$ accept). The former is used in this work. The $(C{+}1)$-st class thus implements a principled, calibrated abstention mechanism consistent with the Kelly criterion, while keeping the architecture and training pipeline simple. See \citet{liu2019deep} for details on the implementation and effect of $o$.

\subsection{ID/OOD Detection CSFs}
This section describes all the CSFs and their variations after applying the Projection Filtering described in Section~\ref{methods:projection_filtering}.

\subsubsection{Class Typical Matching (CTM)\texorpdfstring{~\cite{ngoc2023cosine}}{}}
Prototype matching in feature space consists of quantifying the similarity between a sample $\boldsymbol{x}$ and the last-layer trained weights $\{ \boldsymbol{w}_{1},\dots, \boldsymbol{w}_{C}\}$. Therefore the similarity to the closest trained weight is $\mathrm{CTM}(\boldsymbol{x}) = \max_{k\le C}\; \mathrm{sim}\!\big(\boldsymbol{w}_{k},\,\boldsymbol{h}\big).$ Following~\citet{ngoc2023cosine}, we use cosine similarity in this work, where $\mathrm{sim}(\boldsymbol{u},\boldsymbol{v})=\tfrac{\boldsymbol{u}^\top \boldsymbol{v}}{\|\boldsymbol{u}\|_2\|\boldsymbol{v}\|_2}$. This is the cosine-with-weights (CW) variant that \citet{ngoc2023cosine} analyze in their Section 3.3; the score they propose as CTM uses the class means of the ID training features instead of the weights. CTM scores the \emph{typicality} of $\boldsymbol{x}$ by comparing its feature $\boldsymbol{h}$ against a bank of class representatives. Higher similarity indicates greater in-distribution (ID) conformity.

\subsubsection{Energy\texorpdfstring{~\cite{liu2020energy}}{}}
The energy score is defined as
$\mathrm{Energy}(\boldsymbol{x}) = -T \log \sum_{k=1}^C \exp\!\big(g(\boldsymbol{h})_k/T\big),$ with temperature $T>0$. Higher Energy score typically indicates higher uncertainty.

\subsubsection{Maximum Softmax Response (MSR)\texorpdfstring{~\cite{hendrycks2016baseline}}{} and Maximum Logit Score (MLS)\texorpdfstring{~\cite{hendrycks2019scaling}}{}}
A baseline confidence score given by the maximum predicted probability $\mathrm{MSR}(\boldsymbol{x}) = \max_{k\le C} \boldsymbol{p}_k,$ widely used for OOD detection. Lower values indicate atypical inputs. Similarly, $\mathrm{MLS}$ is a confidence score measured in the logit space, $\mathrm{MLS}(\boldsymbol{x}) = \max_{k\le C} g(\boldsymbol{h})_k,$ often more stable than softmax under temperature changes.

\subsubsection{Predictive Entropy (PE), Generalized Entropy (GEN), Renyi Entropy (REN), Guessing Entropy (GE), and Predictive Collision Entropy (PCE) }

\paragraph{Predictive Entropy (PE)~\cite{ovadia2019can}.} Uncertainty via Shannon entropy~\cite{shannon1948mathematical} of the predictive distribution
$\mathrm{PE}(\boldsymbol{x}) = H\big(p(\boldsymbol{x})\big) = -\sum_{k=1}^C \boldsymbol{p}_k\,\log \boldsymbol{p}_k,$ with larger entropy signaling higher uncertainty.

\paragraph{Generalized Entropy (GEN)~\cite{liu2023gen}.}
GEN is a post-hoc OOD score that uses the softmax probabilities of a trained classifier. Let $\boldsymbol{p}_{(1)}\ge\cdots\ge \boldsymbol{p}_{(C)}$ denote the probabilities sorted in descending order for a given input $\boldsymbol{x}$. For sensitivity and numerical stability in many-class settings, GEN truncates to the top-$M$ classes and computes a generalized entropy with exponent $\gamma\in(0,1)$: $\mathrm{GEN}(\boldsymbol{x})\;=\;\sum_{k=1}^{M} \boldsymbol{p}_{(k)}^{\gamma}\big(1-\boldsymbol{p}_{(k)}\big)^{\gamma}.$ The confidence score is the \textit{negated} generalized entropy so that a larger $\mathrm{GEN}$ (lower entropy) indicates more ID-like predictions.%

\paragraph{Rényi Entropy (REN).}
The Rényi entropy~\cite{renyi1961measures} of order $\alpha$ is a smooth generalization of Shannon entropy that emphasizes different parts of the distribution. Similar to $\mathrm{GEN}$, $\mathrm{REN}$ is defined by a truncation paramater $M$ and exponent $\alpha\in(0,1)$: $\mathrm{REN}(\boldsymbol{x}) = \frac{1}{1-\alpha}\log \sum_{k=1}^M \boldsymbol{p}_{(k)}^{\alpha}$.

\paragraph{Guessing Entropy (GE).}
GE~\cite{massey1994guessing} quantifies the expected number of guesses to identify the true class when labels are guessed in decreasing probability $p_k(\boldsymbol{x})$: if $p_{(1)}\ge\dots\ge p_{(C)}$ are sorted, then $\mathrm{GE}(\boldsymbol{x}) = \sum_{k=1}^C k\, \boldsymbol{p}_{(k)},$ with larger values denoting higher uncertainty.

\paragraph{Predictive Collision Entropy (PCE)~\cite{granese2021doctor}.}
PCE measures prediction uncertainty via the \emph{collision} (order-2 Rényi) entropy of the softmax distribution: $\mathrm{PCE}(\boldsymbol{x}) = -\log \sum_{k=1}^C \boldsymbol{p}_k^2$. Since $\sum_k \boldsymbol{p}_k^2$ is the “collision probability,” PCE grows as the distribution spreads (uncertain/atypical) and shrinks as it peaks (confident/ID-like). This uncertainty score uses the entire predictive distribution rather than just its maximum.

\subsubsection{Mahalanobis Distance (Maha)\texorpdfstring{~\cite{lee2018simple}}{}}
Assuming Gaussian class-conditional features with a shared covariance, the score is the negative Mahalanobis distance to the nearest class centroid: $\mathrm{Maha}(\boldsymbol{x}) = \max_{k\le C}\; -\big(h(\boldsymbol{x})-\boldsymbol{\mu}_k\big)^\top \boldsymbol{\Sigma}^{-1} \big(h(\boldsymbol{x})-\boldsymbol{\mu}_k\big)=-\min_{k\le C}\;\overline{\boldsymbol{h}}_{k}^\top \boldsymbol{\Sigma}^{-1} \overline{\boldsymbol{h}}_{k},$ where $\overline{\boldsymbol{h}}_{k}=h(\boldsymbol{x})-\boldsymbol{\mu}_k$ and $\boldsymbol{\Sigma}$ is the empirical covariance of the class-centered training features, shared by all classes.

\subsubsection{Mahalanobis++ (Maha++)\texorpdfstring{~\cite{mueller2025mahalanobis}}{}}
Mahalanobis++ computes the Mahalanobis score above on L2-normalized features: the class means and the shared covariance are fitted on $\boldsymbol{h}/\lVert\boldsymbol{h}\rVert_2$, and test features are normalized the same way. The normalization removes the dependence of the score on the feature norm. The score has no hyperparameters.

\subsubsection{Neural-Collapse-Inspired detector (NCI)\texorpdfstring{~\cite{liu2023detecting}}{}}
With $\hat{y}$ the predicted class and $\boldsymbol{\mu}$ the mean ID training feature, $\mathrm{NCI}(\boldsymbol{x}) = \frac{\boldsymbol{w}_{\hat{y}}^\top(\boldsymbol{h}-\boldsymbol{\mu})}{\lVert \boldsymbol{h}-\boldsymbol{\mu}\rVert_2} + \alpha\,\lVert \boldsymbol{h}\rVert_1.$ The first term measures how well the centered feature aligns with the classifier weight of the predicted class, which is high for ID samples under collapse; the second term filters OOD samples with a small feature norm. We select $\alpha$ by Bayesian optimization over $[10^{-4}, 10^{-1}]$ (20 initial and 80 further evaluations), minimizing the failure-detection AUGRC on the ID validation split, as for the other tuned CSFs.

\subsubsection{Nearest Neighbor Guide (NNGuide)\texorpdfstring{~\cite{park2023nearest}}{}}
NNGuide is a post-hoc wrapper that modulates any classifier-based OOD score $S_{\text{base}}(\boldsymbol{h})$ using nearest neighbors from an ID bank of features. This bank is formed by sampling $\alpha\in(0,1)$ of ID training features $\boldsymbol{h}_i$ (L2-normalized) and their base scores $s_i=S_{\text{base}}(\boldsymbol{h}_i)$. More specifically, given an input $\boldsymbol{x}$, the corresponding feature $\boldsymbol{h}$ (L2-normalized) defines a confidence-scaled similarity list $\big\{s_i\,\mathrm{cos}(\boldsymbol{h},\boldsymbol{h}_i)\big\}_{i=1}^N$, which is sampled by taking the top-$k$ terms, where $k=\lfloor \alpha N\rfloor$. The top-$k$ terms set the guidance $G(\boldsymbol{h})\;=\;\frac{1}{k}\sum_{i\leq k}\, s_i\,\mathrm{cos}\!\big(\boldsymbol{h},\boldsymbol{h}_i\big)$, and the score $
\mathrm{NNGuide}(\boldsymbol{x})=S_{\text{base}}(\boldsymbol{h})\cdot G(\boldsymbol{h}).$ In practice, $S_{\text{base}}$ is the (negative) Energy score, but NNGuide can improve other classifier-based scores. Intuitively, $G(\boldsymbol{h})$ downscales overconfident far-OOD points where cosine similarities are small and preserves near-ID points using high-confidence neighbors.

\subsubsection{Fast Decision Boundary Detector (fDBD)\texorpdfstring{~\cite{liu2023fast}}{}}
fDBD scores a sample by how far its feature lies from the nearest class decision boundary, regularized by feature deviation from the in-distribution mean. For each non-predicted class $c\neq m(\boldsymbol{x})$, the (unknown) distance in the feature space to the $c$-boundary is lower bounded by $ D_m(\boldsymbol{h},k)=\frac{\big|(\boldsymbol{w}_{m}-\boldsymbol{w}_k)^\top \boldsymbol{h} + (b_{m}-b_k)\big|}{\|\,\boldsymbol{w}_{m}-\boldsymbol{w}_k\,\|_2},$ i.e., the Euclidean distance from $\boldsymbol{h}$ to the separating hyperplane between classes $m(\boldsymbol{x})$ and $c$.
Averaging these distances and \emph{regularizing} by the feature’s deviation from the ID mean $\boldsymbol{\mu}_{\text{train}}$ yields the score $\mathrm{fDBD}(\boldsymbol{x}) = \frac{1}{C-1}\sum^C_{\substack{ k=1\\ k\neq m(\boldsymbol{x}) }} \frac{ D_m(\boldsymbol{h},k)}{\|\,\boldsymbol{h}-\boldsymbol{\mu}_{\text{train}}\|_2}.$
The regularizer compares ID/OOD at equal deviation levels, empirically sharpening separation; the distance term captures that ID features tend to reside farther from decision boundaries than OOD features. 

\subsubsection{Predictive Normalized Maximum Likelihood (pNML)\texorpdfstring{~\cite{bibas2021single}}{}}
pNML treats a deep network as a fixed feature extractor and for each test samples computes a regret score by simulating in closed form the best last-layer update for every possible label. Given the matrix of normalized penultimate-layer training activations $\hat{\boldsymbol{H}}=[\boldsymbol{h}_1/\| \boldsymbol{h}_1\|_2,\dots,\boldsymbol{h}_N/\| \boldsymbol{h}_N\|_2]$, the online–update direction $\boldsymbol{g}$ via the kernel–range projection is $\boldsymbol{g}=\boldsymbol{h}_\perp/\|\boldsymbol{h}_\perp\|_2^2$ if $\boldsymbol{h}_\perp=(\boldsymbol{I}-\hat{\boldsymbol{H}}^+\hat{\boldsymbol{H}})
\boldsymbol{h}\neq 0$; else $\boldsymbol{g}=\frac{\hat{\boldsymbol{H}}^+\hat{\boldsymbol{H}}^{+^{\!\top}}\boldsymbol{h}}{1+\boldsymbol{h}^\top \hat{\boldsymbol{H}}^+\hat{\boldsymbol{H}}^{+^{\!\top}}\boldsymbol{h}}$, where $\hat{\boldsymbol{H}}^+$ is the Moore–Penrose inverse of the normalized training activations. The pNML regret is $\mathrm{pNML}(\boldsymbol{x})=\log\!\sum_{k=1}^{C}\frac{\boldsymbol{p}_k}{\,\boldsymbol{p}_k+\boldsymbol{p}_k^{\boldsymbol{h}^\top \boldsymbol{g}}\big(1-\boldsymbol{p}_k\big)}\,$ and serves as an OOD/failure score (larger $\mathrm{pNML}\Rightarrow$ less trustworthy prediction). Intuitively, $\mathrm{pNML}$ is small when $\boldsymbol{h}$ lies in the high-variance ID subspace or is far from decision boundaries (the genie’s label-specific update has little effect), and large otherwise.

\subsubsection{GradNorm\texorpdfstring{~\cite{huang2021importance}}{}}
Given a trained classifier with softmax $p(\boldsymbol{x})$, GradNorm defines the OOD score as the vector norm of the gradients obtained by backpropagating the Kullback–Leibler divergence from a uniform target, i.e. $\mathrm{GradNorm}(\boldsymbol{x})=\big\|\partial_{\boldsymbol{W}} \mathrm{KL}(\boldsymbol{u}\,\|\,p(\boldsymbol{x}))\big\|_p=\left\lVert\frac{1}{C}\sum_{k=1}^C\frac{\partial \mathcal{L}_{\text{CE}}\left(g\left(\boldsymbol{h}\right),k\right)}{\partial\boldsymbol{W}}\right\rVert_p$, typically using the $L_1$ norm on the \emph{last-layer} weights. This choice is label-agnostic and exploits that in-distribution inputs produce more \emph{peaked} predictions and thus larger gradients than OOD inputs. A simple analysis shows $\mathrm{GradNorm}(\boldsymbol{x})$ factorizes into a feature-space term and an output-space term, capturing joint information that improves separability over output-only scores.

\subsubsection{PCA Reconstruction Error (PCA RecError)\texorpdfstring{~\cite{guan2023revisit}}{}}
PCA Reconstruction Error models the in-distribution feature manifold by fitting a low-dimensional principal subspace on penultimate features and scores a test example by the energy of its component orthogonal to that subspace, so larger residuals indicate atypicality. This approach computes the ID mean $\boldsymbol{\mu}$ and covariance $\boldsymbol{\Sigma}$, then takes the top-$k$ eigenvectors $\boldsymbol{U}_k$ of $\boldsymbol{\Sigma}$, and forms the projector $\boldsymbol{M}=\boldsymbol{U}_k\boldsymbol{U}_k^\top$. The \emph{PCA reconstruction error} for a test point is $e(\boldsymbol{x})\;=\;\big\|(\boldsymbol{I}-\boldsymbol{M})\,\big(h(\boldsymbol{x})-\boldsymbol{\mu}\big)\big\|_2,$ i.e., the energy of the component orthogonal to the ID principal subspace. Although intuitively $e(\boldsymbol{x})$ should be smaller on ID than OOD, a detailed analysis shows that $e(\boldsymbol{x})$ (i) mixes the angle between $h(\boldsymbol{x})-\boldsymbol{\mu}$ and the principal subspace and (ii) the norm $\|h(\boldsymbol{x})-\boldsymbol{\mu}\|_2$, which is typically \emph{larger} for ID than OOD; this blurs separability for vanilla PCA-OOD. To mitigate the norm effect, a simple regularized score $r(\boldsymbol{x})=\frac{\|\boldsymbol{h}-\hat{\boldsymbol{h}}\|_2}{\|\boldsymbol{h}\|_2},$ where $\hat{\boldsymbol{h}}=\hat h(\boldsymbol{x})=\boldsymbol{M}\big(h(\boldsymbol{x})-\boldsymbol{\mu}\big)+\boldsymbol{\mu},$ improves discrimination, and can be fused multiplicatively with logit-based scores.%

\subsubsection{Kernel PCA Reconstruction Error (KPCA RecError)\texorpdfstring{~\cite{fang2025kernel}}{}}
KPCA Reconstruction Error models the in-distribution (ID) feature manifold in a \emph{non-linear} subspace and scores a test point by its reconstruction error in that subspace. To mitigate feature–norm imbalance and preserve useful Euclidean relations, KPCA first $\ell_2$–normalizes features $\hat{\boldsymbol{h}} =\boldsymbol{h}/\|\boldsymbol{h}\|_2$ and defines a Gaussian kernel on the unit sphere $k(\boldsymbol{x},\boldsymbol{x}') = \exp\!\Big(-\tfrac{1}{2\sigma^2}\,\big\|\hat{\boldsymbol{h}}-\hat{\boldsymbol{h}}'\big\|_2^2\Big)
= \exp\!\Big(-\tfrac{1}{\sigma^2}\,\big(1-\cos(\hat{\boldsymbol{h}},\hat{\boldsymbol{h}}')\big)\Big),$ which can be viewed as a Cosine–Gaussian composition. Given ID training points, the centered Gram matrix is defined as $\boldsymbol{K}_c = \boldsymbol{H}\boldsymbol{K}\boldsymbol{H}$ with $\boldsymbol{K}_{ij}=k(\boldsymbol{x}_i,\boldsymbol{x}_j)$ and $\boldsymbol{H}=\boldsymbol{I}-\tfrac{1}{N}\mathbf 1\mathbf 1^\top$. Using the centered Gram matrix, KPCA solves the eigenproblem $\boldsymbol{K}_c \boldsymbol{\alpha}_m \;=\; N\,\lambda_m \,\boldsymbol{\alpha}_m,
$ where $m=1,\dots,N,$ and defines principal coordinates for a test point $\boldsymbol{x}$ via the centered kernel $k_c(\boldsymbol{x},\boldsymbol{x}_i) = k(\boldsymbol{x},\boldsymbol{x}_i) - \tfrac{1}{N}\sum_j k(\boldsymbol{x},\boldsymbol{x}_j) - \tfrac{1}{N}\sum_j k(\boldsymbol{x}_j,\boldsymbol{x}_i) + \tfrac{1}{N^2}\sum_{j\ell} k(\boldsymbol{x}_j,\boldsymbol{x}_\ell)$: $\phi_m(\boldsymbol{x}) \;=\; \frac{1}{\sqrt{\lambda_m}}\,\sum_{i=1}^N \boldsymbol{\alpha}_{m i}\,k_c(\boldsymbol{x},\boldsymbol{x}_i).$ The squared reconstruction error in feature space after projecting onto the top $k$ components is $e(\boldsymbol{x}) = k_c(\boldsymbol{x},\boldsymbol{x}) - \sum_{m=1}^{k} \phi_m(\boldsymbol{x})^2.$ Similar to PCA Reconstruction Error, the larger $e(\boldsymbol{x})$ is, the more atypical $\boldsymbol{x}$ becomes.

Exact KPCA needs the full $N\times N$ kernel matrix and $O(N^2)$ memory. Following~\citet{fang2025kernel}, we approximate the kernel with an explicit Nystr\"om map $\psi:\mathbb{R}^D\!\to\!\mathbb{R}^M$, so that $k(\boldsymbol{x},\boldsymbol{x}')\approx \psi(\hat{\boldsymbol{h}})^\top\psi(\hat{\boldsymbol{h}}')$ with $M\!\ll\!N$. The map is $\psi(\hat{\boldsymbol{h}})=\boldsymbol{W}^{-1/2}\boldsymbol{c}(\hat{\boldsymbol{h}})$, where $c_\ell(\hat{\boldsymbol{h}})=k(\hat{\boldsymbol{h}},\hat{\boldsymbol{h}}^\star_\ell)$ and $\boldsymbol{W}$ is the kernel matrix of the $M$ landmarks $\hat{\boldsymbol{h}}^\star_\ell$. As in~\citet{fang2025kernel}, the landmarks are the $M$ ID training points with the smallest $\log\sum_{c=1}^{C} e^{g(\boldsymbol{h})_c}$, that is, the least confident ID points, which lie near the ID--OOD boundary. \citet{fang2025kernel} call them low-energy points; their energy has the opposite sign to the Energy score defined above. We run PCA on $\psi(\hat{\boldsymbol{h}})$ over the training set, with mean $\boldsymbol{\mu}$ and top-$k$ eigenvectors $\boldsymbol{U}_k$ of the covariance, and score a test point by the negative reconstruction error $\mathrm{KPCA}(\boldsymbol{x}) = -\big\|\big(\boldsymbol{I}-\boldsymbol{U}_k\boldsymbol{U}_k^\top\big)\big(\psi(\hat{\boldsymbol{h}})-\boldsymbol{\mu}\big)\big\|_2$. The kernel width, the number of landmarks ($2{,}000$ to $5{,}000$), and the number of components ($85\%$ to $99\%$ of the explained variance) are tuned as in Appendix~\ref{appendix:hyperparameter_selection}. \citet{fang2025kernel} report that the Cosine-Gaussian kernel and the low-energy landmarks improve detection over linear PCA, at a much lower cost than nearest-neighbor search.

\subsubsection{Residual Projection and Virtual-logit Matching (ViM)\texorpdfstring{~\cite{wang2022vim}}{}}
\paragraph{Residual Projection score.}
ViM first moves the origin of the feature space to $\boldsymbol{o} = -\boldsymbol{W}^{+}\boldsymbol{b}$, so that the logits are bias-free in the new coordinates. Let $\tilde{\boldsymbol{H}}$ be the matrix of shifted ID training features $\boldsymbol{h}-\boldsymbol{o}$. The ID principal subspace $P\subset\mathbb{R}^D$ is the span of the top-$d$ eigenvectors of $\tilde{\boldsymbol{H}}^\top \tilde{\boldsymbol{H}}$, and the columns of $\boldsymbol{R}\in\mathbb{R}^{D\times(D-d)}$ span $P^\perp$. The \emph{residual} of $\boldsymbol{x}$ is $r(\boldsymbol{x})=\boldsymbol{R}\boldsymbol{R}^\top (\boldsymbol{h}-\boldsymbol{o})$, and the residual projection score is $\mathrm{Residual}(\boldsymbol{x})=\|r(\boldsymbol{x})\|_2,$ which increases as the feature departs from the ID principal subspace. This score is class-agnostic and uses feature-space information that is lost in the projection to the logits.

\paragraph{ViM (Virtual-logit Matching).}
ViM combines the class-agnostic residual with the class-dependent logits. It turns the residual norm into a virtual $(C\!+\!1)$-st logit $\ell_0(\boldsymbol{x})=\alpha\,\|r(\boldsymbol{x})\|_2$, where $\alpha= \mathbb{E}_{\boldsymbol{x}\sim \text{ID}}\!\big[\max_{k\le C} g(\boldsymbol{h})_k\big]/ \mathbb{E}_{\boldsymbol{x}\sim \text{ID}}\!\big[\|r(\boldsymbol{x})\|_2\big]$ matches the scale of the virtual logit to the mean maximum logit on the training set. The ViM score is the softmax probability of this virtual class: $\mathrm{ViM}(\boldsymbol{x})\;=\;\frac{\exp\{\ell_0(\boldsymbol{x})\}}{\sum_{k=1}^{C}\exp\{g(\boldsymbol{h})_k\}+\exp\{\ell_0(\boldsymbol{x})\}}.$ The strictly increasing transformation $t(z)=-\ln (1/z-1)$ does not change the ranking and gives the equivalent form $t(\mathrm{ViM}(\boldsymbol{x}))
=\alpha\,\|r(\boldsymbol{x})\|_2-\log\!\sum_{k=1}^C e^{g(\boldsymbol{h})_k},$
that is, the virtual logit plus the Energy score defined above with $T=1$. ViM is large, which means more likely OOD, when the residual is large and the logits are small. Our implementation uses this equivalent form, with the sign reversed so that higher values mean higher confidence.

\subsubsection{Neural Collapse (NeCo)\texorpdfstring{~\cite{ammar2023neco}}{}}
This CSF is motivated by the Neural Collapse phenomena~\cite{papyan2020prevalence}, which unveils geometric properties that manifest at the end of the training process. NeCo's new observation establishes ID/OOD orthogonality, which implies that OOD features concentrate near the origin after projection onto the ID subspace. This CSF fits PCA on ID features to obtain the top-$d$ principal directions $\boldsymbol{P}\in\mathbb{R}^{D\times d}$ (orthonormal columns). Then it scores an input by the normalized projection of its feature onto the ID principal subspace, $\mathrm{NeCo}(\boldsymbol{x})\;=\;\frac{\|\boldsymbol{P}^\top \boldsymbol{h}\|_2}{\|\boldsymbol{h}\|_2},$ so that ID points (well aligned with the ID subspace) yield larger scores, while OOD points (near-orthogonal to that subspace) yield smaller scores. This score is optionally calibrated by multiplying with MLS to inject class-scale information. 

\subsection{CSF Variations}
As described in Section~\ref{methods:projection_filtering}, the global and class subspaces, $\boldsymbol{P}$ and $\boldsymbol{P}_c$ respectively, allow multiple projections that can be used to modify existing OOD detection CSFs. The list below enumerates the modifications applied to the feature, logit, and probability spaces; the resulting per-CSF baselines and variants are summarized in Table~\ref{tab:methods_variations}:
\begin{itemize}
    \item \textbf{Global projection}: $\hat{\boldsymbol{h}} = \boldsymbol{P}\boldsymbol{P}^\top(\boldsymbol{h}-\boldsymbol{\mu})  + \boldsymbol{\mu}$
    \item \textbf{Class projection}: $\hat{\boldsymbol{h}}_c = \boldsymbol{P}_c\boldsymbol{P}_c^\top(\boldsymbol{h}-\boldsymbol{\mu}_c)  + \boldsymbol{\mu}_c$
    \item \textbf{Class averaged projection}: $\hat{\boldsymbol{h}}^{\text{avg}} = \frac{1}{C} \sum_{c=1}^{C}\hat{\boldsymbol{h}}_c$
    \item \textbf{Class predicted projection}: $\hat{\boldsymbol{h}}^{\text{pred}} = \hat{\boldsymbol{h}}_{\hat{y}}$, where ${\hat y}=m(\boldsymbol{x})$
    \item \textbf{Class projected logit}: $\hat{\boldsymbol{g}}^{\text{class}} = \left[\hat{g}_{1},\dots,\hat{g}_{C}\right]$, where $\hat{g}_{c}=  \boldsymbol{w}_{c}^\top\hat{\boldsymbol{h}}_c + b_{c}$
    \item \textbf{Global projected probabilities}: $\hat{\boldsymbol{p}} =\mathrm{softmax}(g(\hat{\boldsymbol{h}}))$
    \item \textbf{Class projected probabilities}: $\hat{\boldsymbol{p}}^{\text{class}} =\mathrm{softmax}(\hat{\boldsymbol{g}}^{\text{class}} )$
    \item \textbf{Class-predicted probabilities}: $\hat{\boldsymbol{p}}^{\text{pred}} =\mathrm{softmax}(g(\hat{\boldsymbol{h}}^{\text{pred}}) )$
    \item \textbf{Class-averaged probabilities}: $\hat{\boldsymbol{p}}^{\text{avg}} =\mathrm{softmax}(g(\hat{\boldsymbol{h}}^{\text{avg}}) )$
    
\end{itemize}

\begin{sidewaystable}
    \centering
    \caption{OOD Detection CSF baselines and variations. This table synthesizes multiple variations using our proposed Projection Filtering approach for all the OOD detection CSFs considered in this work. The ConfidNet row refers to the auxiliary confidence score $s_{\mathcal{W}^{\mathrm{conf}}}$ described in Appendix~\ref{appendix:cfs_variations} (Training Paradigms); it does not admit projection variants.}
    \label{tab:methods_variations}
    \resizebox{0.9\columnwidth}{!}{%
    \begin{tabular}{cccccc}
        \toprule
        \textbf{Score} & \textbf{Unmodified} & \textbf{Global} & \textbf{Class} & \textbf{Class Pred}  & \textbf{Class Avg} \\
        \midrule
        CTM & $\max_k \mathrm{sim}(\boldsymbol{w}_k,\boldsymbol{h})$  & $\max_k \mathrm{sim}(\boldsymbol{w}_k,\hat{\boldsymbol{h}})$  & $\max_k \mathrm{sim}(\boldsymbol{w}_k,\hat{\boldsymbol{h}}_k)$  & $\max_k \mathrm{sim}(\boldsymbol{w}_k,\hat{\boldsymbol{h}}^{\text{pred}})$ & $\max_k \mathrm{sim}(\boldsymbol{w}_k,\hat{\boldsymbol{h}}^{\text{avg}})$\\
        Energy & $-T\log\sum_{k=1}^C e^{g(\boldsymbol{h})_k/T}$ & $-T\log\sum_{k=1}^C e^{g\left(\hat{\boldsymbol{h}}\right)_k/T}$ & $-T\log\sum_{k=1}^C e^{\hat{\boldsymbol{g}}_k^{\text{class}}/T}$ & $-T\log\sum_{k=1}^C e^{g\left(\hat{\boldsymbol{h}}^{\text{pred}}\right)_k/T}$ & $-T\log\sum_{k=1}^C e^{g\left(\hat{\boldsymbol{h}}^{\text{avg}}\right)_k/T}$\\
        MSR & $\max_k \boldsymbol{p}_k$ & $\max_k \hat{\boldsymbol{p}}_k$ & $\max_k\hat{\boldsymbol{p}}_k^{\text{class}}$ & $\max_k\hat{\boldsymbol{p}}_k^{\text{pred}}$  & $\max_k\hat{\boldsymbol{p}}_k^{\text{avg}}$ \\
        MLS & $\max_k g(\boldsymbol{h})_k$ & $\max_k g\left(\hat{\boldsymbol{h}}\right)_k$ & $\max_k\hat{\boldsymbol{g}}_k^{\text{class}}$ & $\max_k g\left(\hat{\boldsymbol{h}}^{\text{pred}}\right)_k$  & $\max_k g\left(\hat{\boldsymbol{h}}^{\text{avg}}\right)_k$\\
        PE & $-\sum_{k=1}^{C} \boldsymbol{p}_k\log\boldsymbol{p}_k$ & $-\sum_{k=1}^{C} \hat{\boldsymbol{p}}_k\log\hat{\boldsymbol{p}}_k$ & $-\sum_{k=1}^{C} \left(\hat{\boldsymbol{p}}_k^{\text{class}}\right)\log\left(\hat{\boldsymbol{p}}_k^{\text{class}}\right)$  & $-\sum_{k=1}^{C} \left(\hat{\boldsymbol{p}}_k^{\text{pred}}\right)\log\left(\hat{\boldsymbol{p}}_k^{\text{pred}}\right)$ & $-\sum_{k=1}^{C} \left(\hat{\boldsymbol{p}}_k^{\text{avg}}\right)\log\left(\hat{\boldsymbol{p}}_k^{\text{avg}}\right)$\\
        GEN & $\sum_{k=1}^{M} \boldsymbol{p}_k^\gamma\left(1-\boldsymbol{p}_k\right)^\gamma$ & $\sum_{k=1}^{M} \hat{\boldsymbol{p}}_k^\gamma\left(1-\hat{\boldsymbol{p}}_k\right)^\gamma$ & $\sum_{k=1}^{M} \left(\hat{\boldsymbol{p}}_k^{\text{class}}\right)^\gamma\left(1-\hat{\boldsymbol{p}}_k^{\text{class}}\right)^\gamma$  & $\sum_{k=1}^{M} \left(\hat{\boldsymbol{p}}_k^{\text{pred}}\right)^\gamma\left(1-\hat{\boldsymbol{p}}_k^{\text{pred}}\right)^\gamma$ & $\sum_{k=1}^{M} \left(\hat{\boldsymbol{p}}_k^{\text{avg}}\right)^\gamma\left(1-\hat{\boldsymbol{p}}_k^{\text{avg}}\right)^\gamma$ \\
        REN & $\frac{1}{1-\alpha}\log\sum_{k=1}^{M} \boldsymbol{p}_k^\alpha$ & $\frac{1}{1-\alpha}\log\sum_{k=1}^{M} \hat{\boldsymbol{p}}_k^\alpha$ & $\frac{1}{1-\alpha}\log\sum_{k=1}^{M} \left(\hat{\boldsymbol{p}}_k^{\text{class}}\right)^\alpha$  & $\frac{1}{1-\alpha}\log\sum_{k=1}^{M} \left(\hat{\boldsymbol{p}}_k^{\text{pred}}\right)^\alpha$ & $\frac{1}{1-\alpha}\log\sum_{k=1}^{M} \left(\hat{\boldsymbol{p}}_k^{\text{avg}}\right)^\alpha$\\
        GE & $\sum_{k=1}^{C} k\boldsymbol{p}_{(k)}$ & $\sum_{k=1}^{C} k\hat{\boldsymbol{p}}_{(k)}$ & $\sum_{k=1}^{C} k\hat{\boldsymbol{p}}_{(k)}^{\text{class}}$ &  $\sum_{k=1}^{C} k\hat{\boldsymbol{p}}_{(k)}^{\text{pred}}$ & $\sum_{k=1}^{C} k\hat{\boldsymbol{p}}_{(k)}^{\text{avg}}$ \\
        PCE & $-\log\sum_{k=1}^{C} \boldsymbol{p}_k^{2}$ & $-\log\sum_{k=1}^{C} \hat{\boldsymbol{p}}_k^{2}$ & $-\log\sum_{k=1}^{C} \left(\hat{\boldsymbol{p}}_k^{\text{class}}\right)^{2}$  & $-\log\sum_{k=1}^{C} \left(\hat{\boldsymbol{p}}_k^{\text{pred}}\right)^{2}$  & $-\log\sum_{k=1}^{C} \left(\hat{\boldsymbol{p}}_k^{\text{avg}}\right)^{2}$\\
        Maha & $\max_k \overline{\boldsymbol{h}}_{k} ^\top\boldsymbol{\Sigma}^{-1}\overline{\boldsymbol{h}}_{k} $ & $\max_k \overline{\hat{\boldsymbol{h}}}_{k}^\top\hat{\boldsymbol{\Sigma}}^{-1}\overline{\hat{\boldsymbol{h}}}_{k} $ &  & $\max_k \left(\overline{\hat{\boldsymbol{h}}^{\text{pred}}}_{k}\right)^\top(\hat{\boldsymbol{\Sigma}}^{\text{pred}})^{-1}\left(\overline{\hat{\boldsymbol{h}}^{\text{pred}}}_{k}\right) $ & $\max_k \left(\overline{\hat{\boldsymbol{h}}^{\text{avg}}}_{k}\right)^\top(\hat{\boldsymbol{\Sigma}}^{\text{avg}})^{-1}\left(\overline{\hat{\boldsymbol{h}}^{\text{avg}}}_{k}\right) $\\
        NNGuide & $E(\boldsymbol{h})G(\boldsymbol{h})$ & $E\left(\hat{\boldsymbol{h}}\right)G\left(\hat{\boldsymbol{h}}\right)$ &  & $E(\hat{\boldsymbol{h}}^{\text{pred}})G(\hat{\boldsymbol{h}}^{\text{pred}})$ & $E(\hat{\boldsymbol{h}}^{\text{avg}})G(\hat{\boldsymbol{h}}^{\text{avg}})$ \\
        fDBD & $\frac{1}{C-1}\sum^C_{\substack{ k=1\\ k\neq m(\boldsymbol{x}) }}\frac{D_{m}(\boldsymbol{h},k)}{\left\lVert\boldsymbol{h}-\boldsymbol{\mu}\right\rVert}$ & $\frac{1}{C-1}\sum_{\substack{ k=1\\ k\neq m(\boldsymbol{x}) }}\frac{D_{m}(\hat{\boldsymbol{h}},k)}{\left\lVert\hat{\boldsymbol{h}}-\hat{\boldsymbol{\mu}}\right\rVert}$ &  & $\frac{1}{C-1}\sum_{\substack{ k=1\\ k\neq m(\boldsymbol{x}) }}\frac{D_{m}(\hat{\boldsymbol{h}}^{\text{pred}},k)}{\left\lVert\hat{\boldsymbol{h}}^{\text{pred}}-\hat{\boldsymbol{\mu}}^{\text{pred}}\right\rVert}$ & $\frac{1}{C-1}\sum_{\substack{ k=1\\ k\neq m(\boldsymbol{x}) }}\frac{D_{m}(\hat{\boldsymbol{h}}^{\text{avg}},k)}{\left\lVert\hat{\boldsymbol{h}}^{\text{avg}}-\hat{\boldsymbol{\mu}}^{\text{avg}}\right\rVert}$\\
        pNML & $\log\sum_{k=1}^C\frac{\boldsymbol{p}_k}{\boldsymbol{p}_k+\boldsymbol{p}_k^{\boldsymbol{h}^\top\boldsymbol{g}}(1-\boldsymbol{p}_k)}$  & $\log\sum_{k=1}^C\frac{\hat{\boldsymbol{p}}_k}{\hat{\boldsymbol{p}}_k+\hat{\boldsymbol{p}}_k^{\boldsymbol{h}^\top\boldsymbol{g}}(1-\hat{\boldsymbol{p}}_k)}$ &  & $\log\sum_{k=1}^C\frac{\hat{\boldsymbol{p}}^{\text{pred}}_k}{\hat{\boldsymbol{p}}^{\text{pred}}_k+(\hat{\boldsymbol{p}}_k^{\text{pred}})^{\boldsymbol{h}^\top\boldsymbol{g}}(1-\hat{\boldsymbol{p}}_k^{\text{pred}})}$ & $\log\sum_{k=1}^C\frac{\hat{\boldsymbol{p}}^{\text{avg}}_k}{\hat{\boldsymbol{p}}^{\text{avg}}_k+(\hat{\boldsymbol{p}}_k^{\text{avg}})^{\boldsymbol{h}^\top\boldsymbol{g}}(1-\hat{\boldsymbol{p}}_k^{\text{avg}})}$\\
        GradNorm & $\left\lVert\frac{1}{C}\sum_{k=1}^C\frac{\partial \mathcal{L}_{\text{CE}}\left(g\left(\boldsymbol{h}\right),k\right)}{\partial\boldsymbol{W}}\right\rVert$ & $\left\lVert\frac{1}{C}\sum_{k=1}^C\frac{\partial \mathcal{L}_{\text{CE}}\left(g\left(\hat{\boldsymbol{h}}\right),k\right)}{\partial\boldsymbol{W}}\right\rVert$ &  & $\left\lVert\frac{1}{C}\sum_{k=1}^C\frac{\partial \mathcal{L}_{\text{CE}}\left(g\left(\hat{\boldsymbol{h}}^{\text{pred}}\right),k\right)}{\partial\boldsymbol{W}}\right\rVert$ & $\left\lVert\frac{1}{C}\sum_{k=1}^C\frac{\partial \mathcal{L}_{\text{CE}}\left(g\left(\hat{\boldsymbol{h}}^{\text{avg}}\right),k\right)}{\partial\boldsymbol{W}}\right\rVert$ \\
        PCA Error &  & $ -\frac{\left\lVert\boldsymbol{h}-\hat{\boldsymbol{h}}\right\rVert}{\left\lVert\boldsymbol{h}\right\rVert} $ & $ \max_k-\frac{\left\lVert\boldsymbol{h}-\hat{\boldsymbol{h}}_k\right\rVert}{\left\lVert\boldsymbol{h}\right\rVert} $ & $ -\frac{\left\lVert\boldsymbol{h}-\hat{\boldsymbol{h}}^{\text{pred}}\right\rVert}{\left\lVert\boldsymbol{h}\right\rVert} $ & \\
        KPCA Error &  & $ -\left\lVert{\Phi}(\boldsymbol{h})-\hat{\boldsymbol{h}}^{\Phi}\right\rVert $ & $ \max_k-\left\lVert{\Phi}(\boldsymbol{h})-\left(\hat{\boldsymbol{h}}^{\Phi}\right)_k\right\rVert $ & $ -\left\lVert{\Phi}(\boldsymbol{h})-\left(\hat{\boldsymbol{h}}^{\Phi}\right)^{\text{pred}}\right\rVert $ & \\
        ViM & $\alpha\left\lVert \tilde{\boldsymbol{R}}\boldsymbol{h} \right\rVert + E\left(\boldsymbol{h}\right)$ &  &  &  & \\
        NeCo & $\frac{\left\lVert\tilde{\boldsymbol{P}}\boldsymbol{h}\right\rVert}{\left\lVert\boldsymbol{h}\right\rVert}$ &  &  &  & \\
        Residual & $\left\lVert\tilde{\boldsymbol{R}}\boldsymbol{h}\right\rVert$  &  &  &  & \\
        ConfidNet & $s_{\mathcal{W}^{\mathrm{conf}}}$ &  &  &  & \\
        \bottomrule
    \end{tabular}
    }
\end{sidewaystable}

\section{CLIP-based OOD Aggregation}\label{appendix:clip}
    This appendix details the CLIP-based OOD aggregation procedure introduced in Section~\ref{methods:clip}. We first define the four proximity metrics (two label-agnostic, two class-aware), describe the $k$-means clustering into near/mid/far buckets, and report the per-source assignment table together with the underlying distance values. Robustness checks (Appendix~\ref{appendix:clustering_robustness}) and a comparison against the OpenOOD binary grouping (Appendix~\ref{appendix:openood_comparison}) follow.

\begin{table}[htb]
    \centering
    \scriptsize
    \caption{OOD dataset clustering. For each source dataset, the corresponding OOD datasets are categorized in near, mid and far datasets based on the CLIP-derived distances.}
    \label{tab:clip_clustering}
    \begin{tabular}{lp{3.3cm}p{3.3cm}p{2cm}}
    \toprule
    \textbf{Source}          & \textbf{Near} & \textbf{Mid} & \textbf{Far} \\
    \midrule
    CIFAR-10        & CIFAR-100, TinyImageNet  & iSUN, LSUN(r), LSUN(c), SVHN & Places365, Textures\\
    SuperCIFAR-100  & CIFAR-10, TinyImageNet  & iSUN, LSUN(r), LSUN(c), SVHN  & Places365, Textures\\
    CIFAR-100       & CIFAR-10, TinyImageNet  & iSUN, LSUN(r), LSUN(c), SVHN  & Places365, Textures\\
    TinyImageNet    & CIFAR-10, CIFAR-100, iSUN, LSUN(r), LSUN(c) & Places365, Textures  & SVHN \\
    \bottomrule
    \end{tabular}
\end{table}

Let $\phi_{\mathrm{img}}$ and $\phi_{\mathrm{text}}$ be fixed CLIP encoders \cite{radford2021learning}.
For any image $\boldsymbol{x}$, we compute the $\ell_2$-normalized image embedding $\boldsymbol{z}=\phi_{\mathrm{img}}(\boldsymbol{x})/\|\phi_{\mathrm{img}}(\boldsymbol{x})\|_2 \in \mathbb{R}^{d_{\mathrm{CLIP}}}$.
Given an in-distribution set $\mathcal{D}_{\mathrm{ID}}=\{(\boldsymbol{x}_i,y_i)\}_{i=1}^N$ and a candidate OOD set $\mathcal{D}_{\mathrm{OOD}}=\{\boldsymbol{x}'_j\}_{j=1}^M$, we extract
$Z_{\mathrm{ID}}=\{\boldsymbol{z}_i\}_{i=1}^N$ and $Z_{\mathrm{OOD}}=\{\boldsymbol{z}'_j\}_{j=1}^M$ under identical preprocessing.

\textbf{Global distances.}
We summarize each set by its empirical Gaussian in CLIP space with means and covariances
$(\boldsymbol{\mu}_{\mathrm{ID}},\boldsymbol{\Sigma}_{\mathrm{ID}})$ and $(\boldsymbol{\mu}_{\mathrm{OOD}},\boldsymbol{\Sigma}_{\mathrm{OOD}})$, and compute the Fr\'echet distance (FD) \cite{dowson1982frechet,frechet1957distance}: $\mathrm{FD}^2(D_{\mathrm{OOD}}\!\to\!D_{\mathrm{ID}})=\|\boldsymbol{\mu}_{\mathrm{ID}}-\boldsymbol{\mu}_{\mathrm{OOD}}\|_2^2+\mathrm{Tr}\!\Big(\boldsymbol{\Sigma}_{\mathrm{ID}}+\boldsymbol{\Sigma}_{\mathrm{OOD}}-2\big(\boldsymbol{\Sigma}_{\mathrm{ID}}^{1/2}\,\boldsymbol{\Sigma}_{\mathrm{OOD}}\,\boldsymbol{\Sigma}_{\mathrm{ID}}^{1/2}\big)^{1/2}\Big).$

As a second global measure, we compute the Kernel Inception Distance (KID)~\cite{binkowski2018demystifying}, an unbiased estimator of the squared maximum mean discrepancy~\cite{gretton2006kernel} with the polynomial kernel $k(\boldsymbol{u},\boldsymbol{v})=(\boldsymbol{u}^\top \boldsymbol{v}/d_{\mathrm{CLIP}} + c_0)^{q}$ ($c_0=1,\;q=3$ following~\citet{binkowski2018demystifying}): $\widehat{\mathrm{KID}}^2=\tfrac{1}{N(N-1)}\sum_{i\ne i'}k(\boldsymbol{z}_i,\boldsymbol{z}_{i'})+\tfrac{1}{M(M-1)}\sum_{j\ne j'}k(\boldsymbol{z}'_j,\boldsymbol{z}'_{j'})-\tfrac{2}{NM}\sum_{i,j}k(\boldsymbol{z}_i,\boldsymbol{z}'_j).$ We average this estimator over $50$ random subsets of up to $1{,}000$ embeddings from each set. Both quantities are evaluated on CLIP embeddings; \emph{smaller} values indicate that $D_{\mathrm{OOD}}$ is closer to the ID manifold.

\textbf{Class-aware distances.}
For ID class $c\in\{1,\dots,C\}$, define the (normalized) image-prototype $\boldsymbol{\mu}_c=\big(\tfrac{1}{|D_c|}\sum_{i:y_i=c} \boldsymbol{z}_i\big)\big/\big\|\tfrac{1}{|D_c|}\sum_{i:y_i=c} \boldsymbol{z}_i\big\|_2$ and the (normalized) text prototype  $\boldsymbol{t}_c=\big(\tfrac{1}{L}\sum_{\ell=1}^L \phi_{\mathrm{text}}({\tt prompt}_\ell(c))\big)\big/\big\|\tfrac{1}{L}\sum_{\ell=1}^L \phi_{\mathrm{text}}({\tt prompt}_\ell(c))\big\|_2$, where $\{{\tt prompt}_\ell(c)\}_{\ell=1}^L$ is the standard 80-template CLIP prompt ensemble of~\citet{radford2021learning} ($L=80$).
For a test embedding $\boldsymbol{z}'$, define the nearest-centroid cosine distance $d_{\mathrm{NC}}(\boldsymbol{z}')=1-\max_{c\le C} \; \boldsymbol{z}'{}^\top \boldsymbol{\mu}_c,$ and the image–text cosine distance $d_{\mathrm{IT}}(\boldsymbol{z}')=1-\max_{c\le C} \; \boldsymbol{z}'{}^\top \boldsymbol{t}_c.$ Aggregate per-dataset by averaging:  $\;\overline{d}_{\mathrm{NC}}=\tfrac{1}{M}\sum_{j=1}^M d_{\mathrm{NC}}(\boldsymbol{z}'_j)$ and  $\;\overline{d}_{\mathrm{IT}}=\tfrac{1}{M}\sum_{j=1}^M d_{\mathrm{IT}}(\boldsymbol{z}'_j)$. Lower values mean the OOD set is \emph{class-closer} to ID.

\textbf{Clustering into proximity buckets.}
We orient all four metrics so that lower $\Rightarrow$ closer and form a feature vector $\mathbf{v}(D_{\mathrm{OOD}})=\big[\;\mathrm{FD}^2,\;\widehat{\mathrm{KID}}^2,\;\overline{d}_{\mathrm{NC}},\;\overline{d}_{\mathrm{IT}}\;\big]^\top.$ We standardize $\mathbf{v}$ across candidate OOD sets (z-score per coordinate) and run $k$-means with $k=3$ (fixed seed) to obtain proximity buckets labeled \emph{near/mid/far}. This CLIP-based protocol is detector-agnostic and applies unchanged to any ID label space or downstream OOD scoring rule.

\begin{table}[htb]
    \centering
    \setlength{\tabcolsep}{1pt}
    \caption{Per-source CLIP-based proximity metrics for every candidate OOD dataset, with the resulting near/mid/far group from $k$-means clustering on the standardized four-metric vector. Heat-map shading is per column (blue $=$ closer to ID, red $=$ farther). The four sub-tables report (a) CIFAR-10, (b) SuperCIFAR-100, (c) CIFAR-100, and (d) TinyImageNet as the source.}
    \label{tab:clip_distances_full}
\begin{subtable}[t]{0.49\textwidth}
\centering
\caption{Source Dataset: CIFAR-10}
{\scriptsize
\begin{tabular}{@{}rccccc@{}}
\toprule
 & \multicolumn{2}{c}{Global} & \multicolumn{2}{c}{Class-aware} & Group \\
\cmidrule(lr){2-3}\cmidrule(lr){4-5}
 & KID & FD & \makecell{Label-Text\\Alignment} & \makecell{Image Centroid\\Distance} &  \\
\midrule
Test & {\cellcolor[HTML]{3B4CC0}} \color[HTML]{F1F1F1} -0.0000 & {\cellcolor[HTML]{3B4CC0}} \color[HTML]{F1F1F1} 0.0028 & {\cellcolor[HTML]{3B4CC0}} \color[HTML]{F1F1F1} 0.7183 & {\cellcolor[HTML]{3B4CC0}} \color[HTML]{F1F1F1} 0.6349 & ID \\
CIFAR-100 & {\cellcolor[HTML]{536EDD}} \color[HTML]{F1F1F1} 0.0002 & {\cellcolor[HTML]{84A7FC}} \color[HTML]{F1F1F1} 0.1592 & {\cellcolor[HTML]{F2C9B4}} \color[HTML]{000000} 0.7885 & {\cellcolor[HTML]{ABC8FD}} \color[HTML]{000000} 0.8085 & Near \\
TinyImageNet & {\cellcolor[HTML]{CCD9ED}} \color[HTML]{000000} 0.0009 & {\cellcolor[HTML]{D2DBE8}} \color[HTML]{000000} 0.3233 & {\cellcolor[HTML]{F39577}} \color[HTML]{000000} 0.8060 & {\cellcolor[HTML]{EDD2C3}} \color[HTML]{000000} 0.9256 & Near \\
iSUN & {\cellcolor[HTML]{F6A385}} \color[HTML]{000000} 0.0015 & {\cellcolor[HTML]{F7AD90}} \color[HTML]{000000} 0.4890 & {\cellcolor[HTML]{F7BA9F}} \color[HTML]{000000} 0.7943 & {\cellcolor[HTML]{C0D4F5}} \color[HTML]{000000} 0.8393 & Mid \\
LSUN resize & {\cellcolor[HTML]{F18D6F}} \color[HTML]{F1F1F1} 0.0016 & {\cellcolor[HTML]{F49A7B}} \color[HTML]{000000} 0.5248 & {\cellcolor[HTML]{F49A7B}} \color[HTML]{000000} 0.8045 & {\cellcolor[HTML]{CEDAEB}} \color[HTML]{000000} 0.8634 & Mid \\
LSUN cropped & {\cellcolor[HTML]{F6A586}} \color[HTML]{000000} 0.0015 & {\cellcolor[HTML]{F5A081}} \color[HTML]{000000} 0.5129 & {\cellcolor[HTML]{E5D8D1}} \color[HTML]{000000} 0.7797 & {\cellcolor[HTML]{B1CBFC}} \color[HTML]{000000} 0.8168 & Mid \\
SVHN & {\cellcolor[HTML]{BB1B2C}} \color[HTML]{F1F1F1} 0.0020 & {\cellcolor[HTML]{B40426}} \color[HTML]{F1F1F1} 0.7009 & {\cellcolor[HTML]{D9DCE1}} \color[HTML]{000000} 0.7744 & {\cellcolor[HTML]{CCD9ED}} \color[HTML]{000000} 0.8607 & Mid \\
Places 365 & {\cellcolor[HTML]{B40426}} \color[HTML]{F1F1F1} 0.0021 & {\cellcolor[HTML]{D44E41}} \color[HTML]{F1F1F1} 0.6379 & {\cellcolor[HTML]{B40426}} \color[HTML]{F1F1F1} 0.8337 & {\cellcolor[HTML]{B40426}} \color[HTML]{F1F1F1} 1.1471 & Far \\
Textures & {\cellcolor[HTML]{C0282F}} \color[HTML]{F1F1F1} 0.0020 & {\cellcolor[HTML]{C43032}} \color[HTML]{F1F1F1} 0.6698 & {\cellcolor[HTML]{D44E41}} \color[HTML]{F1F1F1} 0.8231 & {\cellcolor[HTML]{E67259}} \color[HTML]{F1F1F1} 1.0647 & Far \\
\bottomrule
\end{tabular}}
\end{subtable}
\hfill
\begin{subtable}[t]{0.49\textwidth}
\centering
\caption{Source Dataset: SuperCIFAR-100}
{\scriptsize
\begin{tabular}{@{}rccccc@{}}
\toprule
 & \multicolumn{2}{c}{Global} & \multicolumn{2}{c}{Class-aware} & Group \\
\cmidrule(lr){2-3}\cmidrule(lr){4-5}
 & KID & FD & \makecell{Label-Text\\Alignment} & \makecell{Image Centroid\\Distance} &  \\
\midrule
Test & {\cellcolor[HTML]{3B4CC0}} \color[HTML]{F1F1F1} 0.0000 & {\cellcolor[HTML]{3B4CC0}} \color[HTML]{F1F1F1} 0.0748 & {\cellcolor[HTML]{455CCE}} \color[HTML]{F1F1F1} 0.7581 & {\cellcolor[HTML]{3B4CC0}} \color[HTML]{F1F1F1} 0.7031 & ID \\
CIFAR-10 & {\cellcolor[HTML]{536EDD}} \color[HTML]{F1F1F1} 0.0002 & {\cellcolor[HTML]{7295F4}} \color[HTML]{F1F1F1} 0.1705 & {\cellcolor[HTML]{A7C5FE}} \color[HTML]{000000} 0.7701 & {\cellcolor[HTML]{6180E9}} \color[HTML]{F1F1F1} 0.7511 & Near \\
TinyImageNet & {\cellcolor[HTML]{BAD0F8}} \color[HTML]{000000} 0.0008 & {\cellcolor[HTML]{9ABBFF}} \color[HTML]{000000} 0.2307 & {\cellcolor[HTML]{F7B99E}} \color[HTML]{000000} 0.7840 & {\cellcolor[HTML]{CBD8EE}} \color[HTML]{000000} 0.8738 & Near \\
iSUN & {\cellcolor[HTML]{F2CBB7}} \color[HTML]{000000} 0.0012 & {\cellcolor[HTML]{EDD2C3}} \color[HTML]{000000} 0.3856 & {\cellcolor[HTML]{5977E3}} \color[HTML]{F1F1F1} 0.7607 & {\cellcolor[HTML]{5977E3}} \color[HTML]{F1F1F1} 0.7425 & Mid \\
LSUN resize & {\cellcolor[HTML]{F7B99E}} \color[HTML]{000000} 0.0013 & {\cellcolor[HTML]{F5C0A7}} \color[HTML]{000000} 0.4244 & {\cellcolor[HTML]{6788EE}} \color[HTML]{F1F1F1} 0.7625 & {\cellcolor[HTML]{7295F4}} \color[HTML]{F1F1F1} 0.7720 & Mid \\
LSUN cropped & {\cellcolor[HTML]{EDD1C2}} \color[HTML]{000000} 0.0011 & {\cellcolor[HTML]{F1CCB8}} \color[HTML]{000000} 0.3999 & {\cellcolor[HTML]{A3C2FE}} \color[HTML]{000000} 0.7696 & {\cellcolor[HTML]{536EDD}} \color[HTML]{F1F1F1} 0.7351 & Mid \\
SVHN & {\cellcolor[HTML]{DE614D}} \color[HTML]{F1F1F1} 0.0017 & {\cellcolor[HTML]{B40426}} \color[HTML]{F1F1F1} 0.6208 & {\cellcolor[HTML]{3B4CC0}} \color[HTML]{F1F1F1} 0.7566 & {\cellcolor[HTML]{8CAFFE}} \color[HTML]{000000} 0.8012 & Mid \\
Places 365 & {\cellcolor[HTML]{B40426}} \color[HTML]{F1F1F1} 0.0020 & {\cellcolor[HTML]{DC5D4A}} \color[HTML]{F1F1F1} 0.5562 & {\cellcolor[HTML]{D65244}} \color[HTML]{F1F1F1} 0.7939 & {\cellcolor[HTML]{B40426}} \color[HTML]{F1F1F1} 1.0964 & Far \\
Textures & {\cellcolor[HTML]{EA7B60}} \color[HTML]{F1F1F1} 0.0016 & {\cellcolor[HTML]{E97A5F}} \color[HTML]{F1F1F1} 0.5246 & {\cellcolor[HTML]{B40426}} \color[HTML]{F1F1F1} 0.7980 & {\cellcolor[HTML]{F39778}} \color[HTML]{000000} 1.0003 & Far \\
\bottomrule
\end{tabular}}
\end{subtable}

\vspace{0.4em}

\begin{subtable}[t]{0.49\textwidth}
\centering
\caption{Source Dataset: CIFAR-100}
{\scriptsize
\begin{tabular}{@{}rccccc@{}}
\toprule
 & \multicolumn{2}{c}{Global} & \multicolumn{2}{c}{Class-aware} & Group \\
\cmidrule(lr){2-3}\cmidrule(lr){4-5}
 & KID & FD & \makecell{Label-Text\\Alignment} & \makecell{Image Centroid\\Distance} &  \\
\midrule
Test & {\cellcolor[HTML]{3B4CC0}} \color[HTML]{F1F1F1} -0.0000 & {\cellcolor[HTML]{3B4CC0}} \color[HTML]{F1F1F1} 0.0033 & {\cellcolor[HTML]{3B4CC0}} \color[HTML]{F1F1F1} 0.7043 & {\cellcolor[HTML]{3B4CC0}} \color[HTML]{F1F1F1} 0.6026 & ID \\
CIFAR-10 & {\cellcolor[HTML]{5572DF}} \color[HTML]{F1F1F1} 0.0002 & {\cellcolor[HTML]{8DB0FE}} \color[HTML]{000000} 0.1590 & {\cellcolor[HTML]{F5C1A9}} \color[HTML]{000000} 0.7494 & {\cellcolor[HTML]{96B7FF}} \color[HTML]{000000} 0.7268 & Near \\
TinyImageNet & {\cellcolor[HTML]{BED2F6}} \color[HTML]{000000} 0.0008 & {\cellcolor[HTML]{B2CCFB}} \color[HTML]{000000} 0.2235 & {\cellcolor[HTML]{F7B99E}} \color[HTML]{000000} 0.7512 & {\cellcolor[HTML]{E4D9D2}} \color[HTML]{000000} 0.8436 & Near \\
iSUN & {\cellcolor[HTML]{F3C8B2}} \color[HTML]{000000} 0.0012 & {\cellcolor[HTML]{F3C7B1}} \color[HTML]{000000} 0.3829 & {\cellcolor[HTML]{F4C5AD}} \color[HTML]{000000} 0.7484 & {\cellcolor[HTML]{8BADFD}} \color[HTML]{000000} 0.7128 & Mid \\
LSUN resize & {\cellcolor[HTML]{F7B599}} \color[HTML]{000000} 0.0013 & {\cellcolor[HTML]{F7B599}} \color[HTML]{000000} 0.4204 & {\cellcolor[HTML]{F5A081}} \color[HTML]{000000} 0.7562 & {\cellcolor[HTML]{9EBEFF}} \color[HTML]{000000} 0.7388 & Mid \\
LSUN cropped & {\cellcolor[HTML]{F1CDBA}} \color[HTML]{000000} 0.0011 & {\cellcolor[HTML]{F6BFA6}} \color[HTML]{000000} 0.3999 & {\cellcolor[HTML]{CFDAEA}} \color[HTML]{000000} 0.7364 & {\cellcolor[HTML]{89ACFD}} \color[HTML]{000000} 0.7120 & Mid \\
SVHN & {\cellcolor[HTML]{DA5A49}} \color[HTML]{F1F1F1} 0.0017 & {\cellcolor[HTML]{B40426}} \color[HTML]{F1F1F1} 0.6222 & {\cellcolor[HTML]{F7B99E}} \color[HTML]{000000} 0.7511 & {\cellcolor[HTML]{BCD2F7}} \color[HTML]{000000} 0.7789 & Mid \\
Places 365 & {\cellcolor[HTML]{B40426}} \color[HTML]{F1F1F1} 0.0019 & {\cellcolor[HTML]{DD5F4B}} \color[HTML]{F1F1F1} 0.5456 & {\cellcolor[HTML]{B40426}} \color[HTML]{F1F1F1} 0.7752 & {\cellcolor[HTML]{B40426}} \color[HTML]{F1F1F1} 1.0568 & Far \\
Textures & {\cellcolor[HTML]{E7745B}} \color[HTML]{F1F1F1} 0.0016 & {\cellcolor[HTML]{E57058}} \color[HTML]{F1F1F1} 0.5232 & {\cellcolor[HTML]{ED8366}} \color[HTML]{F1F1F1} 0.7613 & {\cellcolor[HTML]{E9785D}} \color[HTML]{F1F1F1} 0.9780 & Far \\
\bottomrule
\end{tabular}}
\end{subtable}
\hfill
\begin{subtable}[t]{0.49\textwidth}
\centering
\caption{Source Dataset: TinyImageNet}
{\scriptsize
\begin{tabular}{@{}rccccc@{}}
\toprule
 & \multicolumn{2}{c}{Global} & \multicolumn{2}{c}{Class-aware} & Group \\
\cmidrule(lr){2-3}\cmidrule(lr){4-5}
 & KID & FD & \makecell{Label-Text\\Alignment} & \makecell{Image Centroid\\Distance} &  \\
\midrule
Test & {\cellcolor[HTML]{3B4CC0}} \color[HTML]{F1F1F1} -0.0000 & {\cellcolor[HTML]{3B4CC0}} \color[HTML]{F1F1F1} 0.0036 & {\cellcolor[HTML]{3B4CC0}} \color[HTML]{F1F1F1} 0.7141 & {\cellcolor[HTML]{3B4CC0}} \color[HTML]{F1F1F1} 0.6319 & ID \\
CIFAR-100 & {\cellcolor[HTML]{9EBEFF}} \color[HTML]{000000} 0.0008 & {\cellcolor[HTML]{96B7FF}} \color[HTML]{000000} 0.2224 & {\cellcolor[HTML]{96B7FF}} \color[HTML]{000000} 0.7279 & {\cellcolor[HTML]{D3DBE7}} \color[HTML]{000000} 0.7956 & Near \\
CIFAR-10 & {\cellcolor[HTML]{B3CDFB}} \color[HTML]{000000} 0.0009 & {\cellcolor[HTML]{C1D4F4}} \color[HTML]{000000} 0.3220 & {\cellcolor[HTML]{9BBCFF}} \color[HTML]{000000} 0.7288 & {\cellcolor[HTML]{D5DBE5}} \color[HTML]{000000} 0.7979 & Near \\
iSUN & {\cellcolor[HTML]{D6DCE4}} \color[HTML]{000000} 0.0012 & {\cellcolor[HTML]{D7DCE3}} \color[HTML]{000000} 0.3808 & {\cellcolor[HTML]{F6BDA2}} \color[HTML]{000000} 0.7468 & {\cellcolor[HTML]{80A3FA}} \color[HTML]{F1F1F1} 0.7063 & Near \\
LSUN resize & {\cellcolor[HTML]{DDDCDC}} \color[HTML]{000000} 0.0013 & {\cellcolor[HTML]{DEDCDB}} \color[HTML]{000000} 0.4039 & {\cellcolor[HTML]{F7A889}} \color[HTML]{000000} 0.7500 & {\cellcolor[HTML]{8BADFD}} \color[HTML]{000000} 0.7186 & Near \\
LSUN cropped & {\cellcolor[HTML]{F5C0A7}} \color[HTML]{000000} 0.0016 & {\cellcolor[HTML]{F5C4AC}} \color[HTML]{000000} 0.4989 & {\cellcolor[HTML]{E3D9D3}} \color[HTML]{000000} 0.7406 & {\cellcolor[HTML]{AAC7FD}} \color[HTML]{000000} 0.7503 & Near \\
Places 365 & {\cellcolor[HTML]{EBD3C6}} \color[HTML]{000000} 0.0014 & {\cellcolor[HTML]{D9DCE1}} \color[HTML]{000000} 0.3887 & {\cellcolor[HTML]{B40426}} \color[HTML]{F1F1F1} 0.7645 & {\cellcolor[HTML]{B40426}} \color[HTML]{F1F1F1} 0.9846 & Mid \\
Textures & {\cellcolor[HTML]{EDD2C3}} \color[HTML]{000000} 0.0014 & {\cellcolor[HTML]{F0CDBB}} \color[HTML]{000000} 0.4697 & {\cellcolor[HTML]{F29274}} \color[HTML]{F1F1F1} 0.7528 & {\cellcolor[HTML]{E36C55}} \color[HTML]{F1F1F1} 0.9317 & Mid \\
SVHN & {\cellcolor[HTML]{B40426}} \color[HTML]{F1F1F1} 0.0025 & {\cellcolor[HTML]{B40426}} \color[HTML]{F1F1F1} 0.7948 & {\cellcolor[HTML]{E5D8D1}} \color[HTML]{000000} 0.7409 & {\cellcolor[HTML]{F7B396}} \color[HTML]{000000} 0.8726 & Far \\
\bottomrule
\end{tabular}}
\end{subtable}
\end{table}
\paragraph{Walk-through of the assignments.} Table~\ref{tab:clip_walkthrough} reads the groups off the distances of Table~\ref{tab:clip_distances_full}. For the CIFAR sources, the far group is set by the class-aware distances: Places365 and Textures are far from every ID class mean, while SVHN has the largest Fr\'echet distance but only moderate class-aware distances, so it joins the scene sets in the mid group. For TinyImageNet, the order changes: SVHN is the clear outlier in both distribution distances, and Places365 and Textures differ from the near sets only in the class-aware distances. The content column is for orientation only; the groups come from the four distances alone.

\begin{table}[htb]
    \centering
    \scriptsize
    \caption{Walk-through of the regime assignments: the distances of Table~\ref{tab:clip_distances_full} that place each OOD set in its group, and the content of the set. SuperCIFAR-100 and CIFAR-100 have the same groups as CIFAR-10, with the same pattern (for example, SVHN has the largest FD, $0.62$, but an image-centroid distance of only $0.78$--$0.80$).}
    \label{tab:clip_walkthrough}
    \begin{tabular}{@{}llp{5.6cm}p{3.6cm}@{}}
    \toprule
    \textbf{OOD set} & \textbf{Group} & \textbf{Distances that place it} & \textbf{Content} \\
    \midrule
    \multicolumn{4}{@{}l}{\emph{Source: CIFAR-10}} \\
    CIFAR-100 & Near & Lowest KID ($0.0002$) and FD ($0.16$) & Same image source and format, related object classes \\
    TinyImageNet & Near & Low KID ($0.0009$) and FD ($0.32$) & Downsampled object photographs \\
    iSUN, LSUN (both) & Mid & KID $0.0015$--$0.0016$; image-centroid distance $0.82$--$0.86$, lower than for TinyImageNet ($0.93$) & Scene photographs \\
    SVHN & Mid & Largest FD ($0.70$), but text and centroid distances ($0.77$, $0.86$) like those of the scene sets & House-number digits \\
    Places365, Textures & Far & Largest class-aware distances (centroid $1.06$--$1.15$, text $0.82$--$0.83$) & Scenes; texture patches without objects \\
    \midrule
    \multicolumn{4}{@{}l}{\emph{Source: TinyImageNet}} \\
    CIFAR-10, CIFAR-100 & Near & KID $0.0008$--$0.0009$, centroid distance $0.80$ & Object photographs at lower resolution \\
    iSUN, LSUN (both) & Near & KID $\leq 0.0016$; centroid distance $0.71$--$0.75$, lower than for the CIFAR sets & Scene photographs \\
    Places365, Textures & Mid & KID $0.0014$, as low as for the near sets, but the largest centroid distances ($0.93$--$0.98$) & Scenes; texture patches without objects \\
    SVHN & Far & Largest KID ($0.0025$) and FD ($0.79$), well above every other set & House-number digits \\
    \bottomrule
    \end{tabular}
\end{table}

\subsection{Clustering Robustness Analysis}\label{appendix:clustering_robustness}
    We validate the stability of our three-group (near/mid/far) stratification across three sources of variability: the CLIP encoder, the distance metrics, and the number of clusters $k$. Our evidence shows that the ordinal ranking of OOD datasets---which determines group membership---is highly stable, even though exact cluster boundaries may shift.

\paragraph{CLIP encoder robustness.} We computed proximity rankings using three CLIP backbones of increasing capacity (ViT-B/32, ViT-B/16, ViT-L/14) and measured pairwise Spearman rank correlations across all four source datasets. Table~\ref{tab:clip_encoder_robustness} reports the results.

\begin{table}[h]
    \centering
    \scriptsize
    \caption{Spearman rank correlation ($\rho$) of OOD proximity rankings across three CLIP backbones, aggregated over all 12 source--backbone pairs.}
    \label{tab:clip_encoder_robustness}
    \begin{tabular}{lccc}
    \toprule
    \textbf{Metric} & \textbf{Mean $\rho$ (12 pairs)} & \textbf{Min $\rho$} & \textbf{All sig.\ ($p<0.05$)?} \\
    \midrule
    FD                      & 0.92 & 0.83 & Yes \\
    KID                     & 0.95 & 0.86 & Yes \\
    Image centroid dist.    & 0.91 & 0.83 & Yes \\
    Text alignment          & 0.58 & $-$0.33 & \textbf{No} \\
    \bottomrule
    \end{tabular}
\end{table}

For the three distance-based metrics, all 12 combinations yield $\rho \geq 0.83$ with statistical significance. Text alignment is the exception: ViT-L/14 disagrees substantially with ViT-B/32 and ViT-B/16, including a negative correlation for CIFAR-100 ($\rho = -0.33$). However, text alignment is one of four metrics, and its instability is absorbed by the three robust ones. We recommend that text alignment not be used in isolation for stratification.

\paragraph{Distance metrics (leave-one-metric-out).} We tested whether removing any single metric from the four-metric composite changes the ordinal ranking. Across 48 tests (3 backbones $\times$ 4 datasets $\times$ 4 dropped metrics), no single metric removal drops Spearman $\rho$ below 0.85 versus the full composite (mean $\rho \geq 0.95$ for all backbones). Removing text alignment entirely does not change group membership.

\paragraph{The $k=3$ choice (silhouette analysis).} Beyond the semantic motivation (near/mid/far), we validated $k=3$ with a data-driven silhouette analysis across all 12 dataset $\times$ backbone configurations. Table~\ref{tab:silhouette_analysis} shows that $k=3$ achieves the highest mean silhouette and is the largest $k$ for which all clusters remain well-separated: for $k \geq 4$, at least one cluster degenerates to a singleton (min-cluster silhouette $= 0$) in every configuration. While $k=2$ is competitive for some backbones (particularly ViT-B/16), $k=3$ provides finer granularity (near/mid/far vs.\ near/far) at no cost to cluster quality.

\begin{table}[h]
    \centering
    \scriptsize
    \caption{Silhouette analysis for different values of $k$ across 12 source--backbone configurations.}
    \label{tab:silhouette_analysis}
    \begin{tabular}{ccc}
    \toprule
    $k$ & \textbf{Mean silhouette} & \textbf{Configs with min-cluster sil.\ $> 0$} \\
    \midrule
    2 & 0.45 & 12/12 \\
    \textbf{3} & \textbf{0.46} & \textbf{9/12} \\
    4 & 0.43 & 0/12 \\
    5 & 0.36 & 0/12 \\
    6 & 0.25 & 0/12 \\
    \bottomrule
    \end{tabular}
\end{table}

\paragraph{Hierarchical clustering validation.} To verify that groupings are not a $k$-means artifact, we ran Ward and average-linkage hierarchical clustering on the same feature vectors and compared assignments via the Adjusted Rand Index (ARI). In Table~\ref{tab:silhouette_analysis}, we see that $k$-means and Ward produce identical cluster memberships in every configuration (perfect ARI $= 1.0$) for $k=3$. Average linkage agrees in 8/12 cases; the 4 disagreements involve minor boundary shifts (e.g., SVHN grouping in TinyImageNet).

\begin{table}[h]
    \centering
    \scriptsize
    \caption{Agreement (ARI) between $k$-means and hierarchical clustering across 12 configurations.}
    \label{tab:hierarchical_ari}
    \begin{tabular}{ccc}
    \toprule
    $k$ & \textbf{$k$-means vs Ward (ARI)} & \textbf{$k$-means vs Average (ARI)} \\
    \midrule
    2 & 1.0 in 11/12, mean = 0.96 & 1.0 in 8/12, mean = 0.86 \\
    \textbf{3} & \textbf{1.0 in 12/12, mean = 1.00} & \textbf{1.0 in 8/12, mean = 0.90} \\
    4 & 1.0 in 12/12, mean = 1.00 & 1.0 in 8/12, mean = 0.88 \\
    \bottomrule
    \end{tabular}
\end{table}

\paragraph{Rank--group ordinal consistency (Kendall $\tau$).} As a direct test that group assignments respect the ordinal distance, we calculated the average ranks across all four metrics per OOD dataset and measured Kendall $\tau$ against the group number ($1=\text{Near}$, $2=\text{Mid}$, $3=\text{Far}$). In Table~\ref{tab:kendall_tau}, the correlations show a strong concordance (mean $\tau = 0.73$, range $0.59$--$0.87$); 10 of 12 are significant at $p < 0.05$. The two exceptions (CIFAR-100 and SuperCIFAR-100 on ViT-B/16, $p = 0.058$) are borderline and directionally consistent.

\begin{table}[h]
    \centering
    \scriptsize
    \caption{Kendall $\tau$ between average metric rank and assigned group number.}
    \label{tab:kendall_tau}
    \begin{tabular}{llcc}
    \toprule
    \textbf{Source dataset} & \textbf{CLIP backbone} & $\tau$ & \textbf{$p$-value} \\
    \midrule
    CIFAR-10       & ViT-B/16 & 0.716 & 0.026 \\
    CIFAR-10       & ViT-B/32 & 0.689 & 0.029 \\
    CIFAR-10       & ViT-L/14 & 0.798 & 0.011 \\
    CIFAR-100      & ViT-B/16 & 0.592 & 0.058 \\
    CIFAR-100      & ViT-B/32 & 0.676 & 0.030 \\
    CIFAR-100      & ViT-L/14 & 0.840 & 0.008 \\
    SuperCIFAR-100 & ViT-B/16 & 0.592 & 0.058 \\
    SuperCIFAR-100 & ViT-B/32 & 0.676 & 0.030 \\
    SuperCIFAR-100 & ViT-L/14 & 0.866 & 0.005 \\
    TinyImageNet   & ViT-B/16 & 0.779 & 0.014 \\
    TinyImageNet   & ViT-B/32 & 0.747 & 0.020 \\
    TinyImageNet   & ViT-L/14 & 0.747 & 0.020 \\
    \bottomrule
    \end{tabular}
\end{table}

\paragraph{Summary.} The ordinal structure on which our conclusions rest is robust: (1) three CLIP backbones produce highly correlated rankings ($\rho \geq 0.83$ for distance metrics); (2) dropping any single metric preserves the ranking ($\rho \geq 0.85$); (3) $k=3$ maximizes silhouette and is the largest $k$ with well-separated clusters; (4) $k$-means and Ward recover identical groupings (ARI $= 1.0$ in 12/12); and (5) average metric rank correlates strongly with group assignment (mean Kendall $\tau = 0.73$, 10/12 significant).

\subsection{Comparison with OpenOOD Binary Grouping}\label{appendix:openood_comparison}
    To examine whether our conclusions depend on the three-group (near/mid/far) stratification introduced in Section~\ref{methods:clip}, we re-ran the full statistical pipeline using the binary (near/far) grouping adopted by OpenOOD v1.5~\cite{zhang2023openood}. To ensure a fair comparison, we restrict to the intersection of OOD datasets present in both OpenOOD and our benchmark and evaluate only base CSFs (no projection filtering variants). TinyImageNet is excluded as a source because OpenOOD does not define near/far groupings for it. Table~\ref{tab:openood_grouping} shows the mapping.

\begin{table}[h]
    \centering
    \scriptsize
    \caption{OpenOOD-style binary grouping used for the comparison in Figure~\ref{fig:openood_comparison}. Only the intersection of OOD datasets shared between OpenOOD and our benchmark is retained.}
    \label{tab:openood_grouping}
    \begin{tabular}{lll}
    \toprule
    \textbf{Source} & \textbf{Near-OOD} & \textbf{Far-OOD} \\
    \midrule
    CIFAR-10        & CIFAR-100, TinyImageNet & SVHN, Textures, Places365 \\
    CIFAR-100       & CIFAR-10, TinyImageNet  & SVHN, Textures, Places365 \\
    SuperCIFAR-100  & CIFAR-10, TinyImageNet  & SVHN, Textures, Places365 \\
    \bottomrule
    \end{tabular}
\end{table}

\begin{figure*}[htb]
    \centering
    \includegraphics[width=0.85\textwidth, trim={0 0 0 1.5cm}, clip]{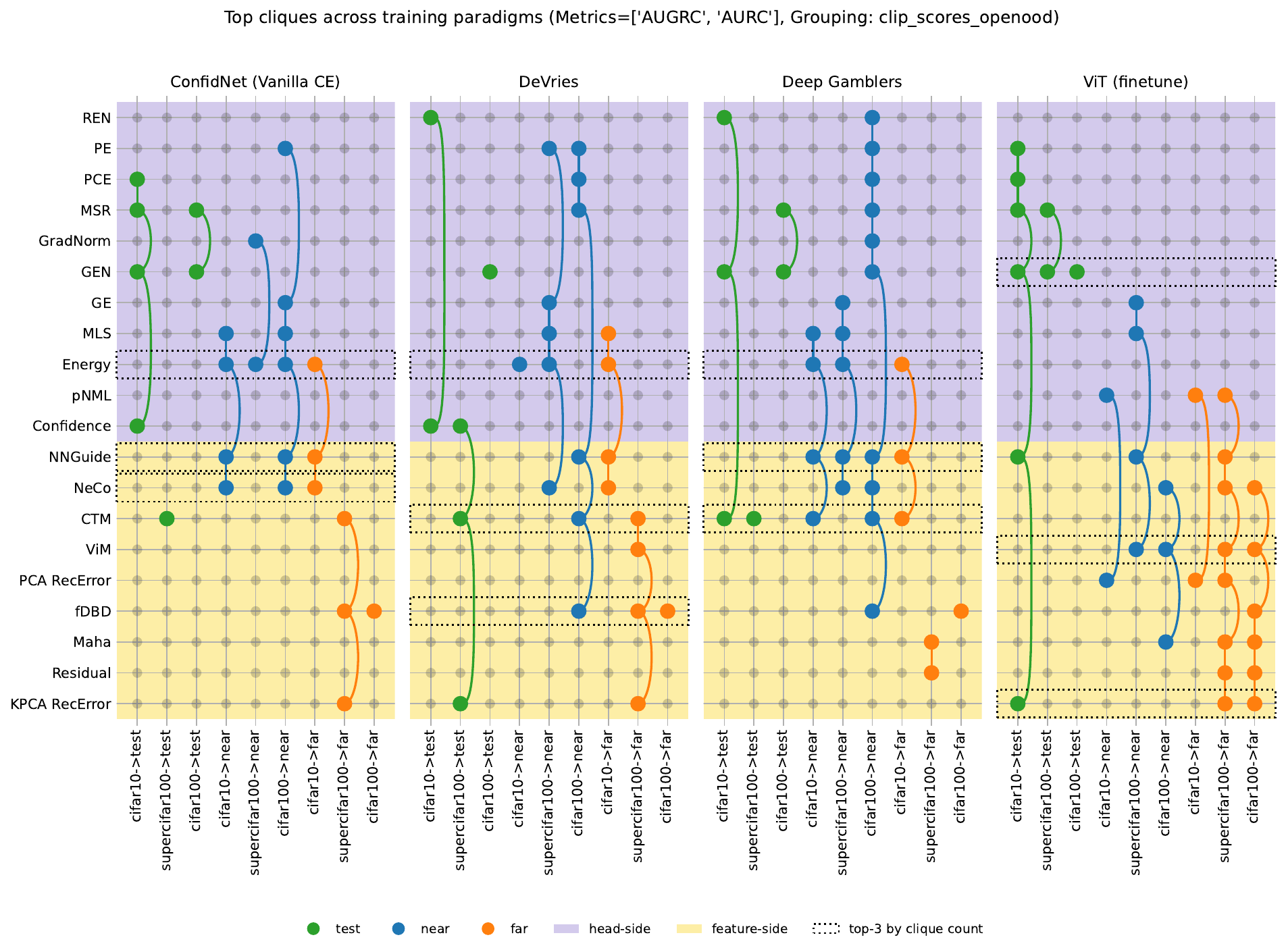}
    \caption{Top-clique maps under OpenOOD-style binary (near/far) grouping. Same panel layout as Figure~\ref{fig:top_clique_vgg13}: VGG-13 with vanilla cross-entropy / ConfidNet (left), VGG-13 with DeVries, VGG-13 with Deep Gamblers, and a fine-tuned ViT (right). Rows are head-side (top, blue band) and feature-side (bottom, peach band) CSFs; columns are source datasets crossed with regimes (\emph{test}, \emph{near}, \emph{far}) under the OpenOOD binary partition. TinyImageNet is excluded as a source because OpenOOD does not define a near/far split for it. Compared to the three-group maps in Figure~\ref{fig:top_clique_vgg13}: for CIFAR-10 on both backbones, the binary near and far cliques are identical, hiding the distinct mid-regime winners visible in the three-group analysis; for CIFAR-100 and SuperCIFAR-100 on ViT, the binary far clique contains eight and six tied CSFs, respectively, whereas the three-group split resolves this into a focused mid clique (KPCA RecError) and a distinct far clique (Residual, ViM).}
    \label{fig:openood_comparison}
\end{figure*}

Figure~\ref{fig:openood_comparison} shows the resulting top-clique maps under this binary grouping, for both VGG-13 and ViT backbones (AUGRC/AURC metrics). Three observations emerge from the comparison with the three-group maps in Figure~\ref{fig:top_clique_vgg13}:

\begin{enumerate}
    \item \textbf{Binary far cliques can collapse to match near, hiding regime transitions.} For CIFAR-10 on VGG-13, the binary near and far cliques are identical (Energy, MLS, NNGuide), whereas the three-group analysis reveals that the mid regime is dominated by Energy alone and the far regime adds NeCo. Similarly, for CIFAR-10 on ViT, binary near and far both yield PCA RecError and pNML, but the three-group analysis separates a distinct mid winner (KPCA RecError) from the true far clique (Residual, ViM, pNML). The binary partition merges these distinct regimes into a single group, making the transitions invisible.

    \item \textbf{Binary far cliques can become uninformatively broad.} For CIFAR-100 on ViT, the binary far clique contains eight CSFs (KPCA RecError, Maha, NNGuide, NeCo, Residual, ViM, fDBD, pNML), whereas the three-group analysis separates this into a focused mid clique (KPCA RecError) and a distinct far clique (Residual, ViM). The binary split pools datasets at different distributional distances from the source, inflating the number of statistically tied CSFs and reducing the discriminative power of the comparison.

    \item \textbf{The mid-OOD regime captures distinct CSF preferences lost under binary grouping.} For CIFAR-100 on VGG-13, the three-group analysis identifies geometry-aware CSFs (CTM, NNGuide, fDBD) as the mid-OOD winners, while only fDBD survives in the far clique. Under the binary split, these mid-regime winners are absorbed into the far group, where fDBD alone remains in the top clique, and the transition from margin-based to geometry-aware scoring is obscured.
\end{enumerate}

These results support our choice of a three-group stratification: the additional mid-OOD category captures a regime with distinct CSF preferences that is obscured by a binary near/far partition. Moreover, the comparison highlights the complementary value of the projection filtering analysis, which identifies architecture-aware scoring variants that would not be evaluated under the OpenOOD protocol.

\subsection{Limits of the CLIP-based Regimes}\label{appendix:clip_limits}
    The regimes measure how far an OOD set lies from the ID set in CLIP space. We use this distance as a proxy for how hard the OOD set is for the detectors of the classifier. The proxy can fail in two ways.

\paragraph{CLIP may not resolve the shift.} Zero-shot CLIP does poorly on fine-grained classification, such as telling apart aircraft variants, flower species, or car models~\cite{radford2021learning}. In industrial inspection, CLIP-type models encode the class of the object more than the defect on it~\cite{zhou2024anomalyclip}, so a set of scratched or dented parts can lie close to the clean parts in CLIP space. In such domains, the CLIP distances can compress OOD sets that differ in ways that matter, and the regimes can put these sets in the wrong group.

\paragraph{CLIP distance need not match the failure modes of the classifier.} A detector fails on an OOD input when the features of the classifier place that input among the ID classes, so that it looks like a confident ID sample. Whether this happens depends on the cues that the classifier uses, not on the meaning that CLIP encodes. A classifier that relies on texture, color, or background can be confident on a set that CLIP places far, if the set shares these cues with an ID class; the regime then calls the set far, but the detectors fail on it as on a near set. Conversely, a set that CLIP places near can be easy for the classifier, if a small change in those cues moves its inputs away from every ID class. Two OOD sets in the same regime can therefore differ in difficulty for the same classifier, and their competitive CSFs can differ.

\paragraph{What limits these effects.} Three of the four distances use only image embeddings and need no class names, so they do not depend on the zero-shot text matching in which the weaknesses above were measured. These three distances also order the datasets the same way across three CLIP encoders (Spearman $\rho \geq 0.83$, Section~\ref{methods:clip}), while the text-alignment distance does not, which is why we never use it alone. Neither check compares the regimes with the difficulty that the classifier sees. A direct test would group the OOD sets by detector performance on held-out models and compare these groups with the CLIP regimes.

\section{Statistical Tests for Clique Generation}\label{appendix:friedman}
    
Let $k$ be the number of CSFs and $N$ the number of complete blocks (e.g., each block is a dataset/condition on which all $k$ CSFs are evaluated on the same metric). Within block $i\in\{1,\dots,N\}$, rank the CSFs so that \(r_{ij}\in\{1,\dots,k\}\) is the rank of CSF \(j\) (use mid-ranks for ties). Define the average rank \(\bar R_j=\frac{1}{N}\sum_{i=1}^{N} r_{ij}\). The Friedman statistic tests the null \(H_0\): all CSFs are equivalent in distribution of ranks \cite{friedman1937use,demvsar2006statistical}:
\[
Q \;=\; \frac{12N}{k(k+1)}\sum_{j=1}^{k}\bar R_j^{\,2}\;-\;3N(k+1),
\]
(optionally applying a standard tie correction within blocks). For finite samples, the Iman--Davenport \(F\)-approximation is recommended \cite{iman1980approximations}:
\[
F_F \;=\; \frac{(N-1)\,Q}{\,N(k-1)\;-\;Q\,}\;\sim\;F_{k-1,\,(k-1)(N-1)}.
\]
If \(F_F\) exceeds the critical value at level \(\alpha\), we reject \(H_0\) and proceed with post-hoc pairwise comparisons.

\paragraph{Conover post-hoc \& Bron--Kerbosch cliques (top groups).}
For each pair \((i,j)\) we compare average ranks using Conover’s post-hoc test~\cite{conover1999practical}. With common standard error
\[
\mathrm{SE}\;=\;\sqrt{\frac{k(k+1)}{6N}},\qquad
T_{ij}\;=\;\frac{\lvert \bar R_i-\bar R_j\rvert}{\mathrm{SE}},
\]
two-sided \(p\)-values are obtained from the normal (or \(t\)) reference, and we use them without multiplicity adjustment (Appendix~\ref{appendix:holm} compares Holm-adjusted cliques). To summarize statistically indistinguishable winners, construct an \emph{indifference graph} \(G=(V,E)\) with nodes \(V=\{1,\dots,k\}\) (CSFs) and edges \((i,j)\in E\) iff \(p_{ij}\ge\alpha\) (i.e., the pair is not significantly different). Enumerate all \emph{maximal cliques} of \(G\) using the Bron--Kerbosch algorithm with pivoting (state \((R,P,X)\); recursively add \(v\in P\setminus N(u)\) for a high-degree pivot \(u\); output \(R\) when \(P=X=\varnothing\)) \cite{bron1973algorithm}. Each maximal clique is a set of CSFs that are mutually indistinguishable under the Conover test; reporting the leading clique(s), sorted by best/mean \(\bar R\), yields layered, statistically justified ``top groups,'' alongside \(\bar R_j\) and the full \(p\)-matrix for transparency.

\subsection{Simplified example}

To illustrate the evaluation methodology, we consider a simplified scenario involving $k=6$ Confidence Scoring Functions (CSFs): Confidence, GEN, MSR, CTM, fDBD, and Energy. We evaluate these models on the CIFAR-10 dataset using the AUGRC metric, focusing solely on the misclassification scenario. The evaluation aggregates results across three training paradigms, each executed five times, resulting in a total of $N=15$ experimental blocks.

As shown in Table~\ref{tab:block_ranking}, within each block (characterized by a unique combination of dataset, paradigm, metric, and run), the CSFs are ranked based on their performance scores ($r_{ij}\in\{1,\dots,6\}$, with $1$ best). These rankings are then averaged across all $N=15$ blocks, as summarized in Table~\ref{tab:average_ranking}.

Using these average ranks, we compute the Friedman statistic $Q\approx 46.16$ with $p<10^{-8}$. Since this $p$-value falls below the significance level $\alpha=0.05$, we reject the null hypothesis and conclude that significant differences exist between the CSFs. Consequently, we proceed with post-hoc pairwise comparisons.

\begin{table}[!htb]
\scriptsize
\caption{Worked example of the rank pipeline on six CSFs evaluated on CIFAR-10 (AUGRC, misclassification regime), with $N=15$ blocks formed by 3 paradigms $\times$ 5 runs. (a) Within-block ranking of the six CSFs by AUGRC for one example block (lower score $\Rightarrow$ better $\Rightarrow$ smaller rank). (b) Average rank $\bar R_j$ per CSF across all 15 blocks.}
\begin{subtable}[t]{0.70\textwidth}
    \centering
    \caption{Block example}
    \label{tab:block_ranking}
    \begin{tabular}{lllllll}
    \toprule
    \textbf{Dataset} & \textbf{Paradigm} & \textbf{Metric} & \textbf{Run} & \textbf{CSF} & \textbf{Score} & \textbf{Rank} \\
    \midrule
    test & ConfidNet & AUGRC & 1 & Confidence & 4.453 & 1 \\
    test & ConfidNet & AUGRC & 1 & GEN        & 4.619 & 2 \\
    test & ConfidNet & AUGRC & 1 & MSR        & 4.737 & 3 \\
    test & ConfidNet & AUGRC & 1 & CTM        & 5.479 & 4 \\
    test & ConfidNet & AUGRC & 1 & fDBD       & 5.677 & 5 \\
    test & ConfidNet & AUGRC & 1 & Energy     & 5.811 & 6 \\
    \bottomrule
    \end{tabular}
\end{subtable}
\hfill
\begin{subtable}[t]{0.25\textwidth}
    \centering
    \caption{Averaged rankings}
    \label{tab:average_ranking}
    \begin{tabular}{ll}
    \toprule
    \textbf{CSF} & \textbf{Avg Rank} \\
    \midrule
    GEN        & 1.667 \\
    Confidence & 2.200 \\
    MSR        & 3.067 \\
    CTM        & 3.933 \\
    fDBD       & 4.600 \\
    Energy     & 5.533 \\
    \bottomrule
    \end{tabular}
\end{subtable}
\end{table}

\begin{figure}[htb]
    \centering
    \includegraphics[width=0.5\linewidth]{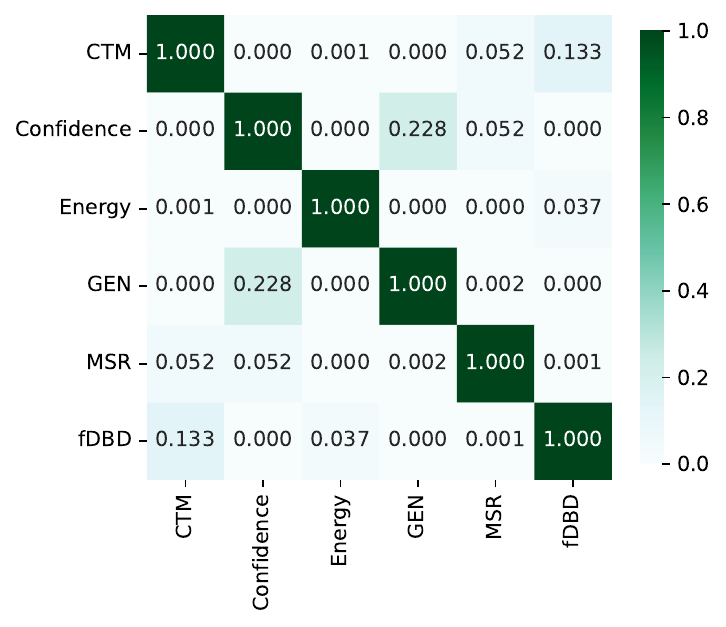}
    \caption{Unadjusted Conover $p$-values}
    \label{fig:conover-holm}
\end{figure}

We apply Conover’s post-hoc test to compare the average rankings, computing unadjusted two-sided $p$-values (see Figure~\ref{fig:conover-holm}). This analysis allows us to identify ``cliques'' or groups of CSFs that are statistically indistinguishable from one another ($p_{ij}\geq 0.05$). In this example, we identify five cliques: 
Clique 1: \texttt{['CTM', 'MSR']}, 
Clique 2: \texttt{['CTM', 'fDBD']}, 
Clique 3: \texttt{['Confidence', 'GEN']}, 
Clique 4: \texttt{['Confidence', 'MSR']}, and 
Clique 5: \texttt{['Energy']}. 

While visual identification of cliques is straightforward in this small-scale example, the combinatorial complexity increases significantly with a larger number of CSFs. To address this, we employ the Bron--Kerbosch algorithm to systematically identify maximal cliques in larger scenarios. Once identified, the cliques are ranked based on their constituent members. Finally, we select non-overlapping cliques using a greedy layering approach, organizing them from best to worst performance. The resulting hierarchical layers are presented in Table~\ref{tab:layered_cliques}. For Figure~\ref{fig:top_clique_vgg13}, we only report the first layer for all the possible scenarios.

\begin{table}[htb]
\centering
\scriptsize
\caption{Layered cliques ranked from best to worst.}
\label{tab:layered_cliques}
\begin{tabular}{lll}
\toprule
\textbf{Layer} & \textbf{Average Rank} & \textbf{Members} \\
\midrule
1 & 1.933 & Confidence, GEN \\
2 & 3.500 & CTM, MSR \\
3 & 5.533 & Energy \\
\bottomrule
\end{tabular}
\end{table}
\subsection{Unadjusted versus Holm-adjusted cliques}\label{appendix:holm}
The cliques in this paper use unadjusted Conover $p$-values. Adjusting them with Holm's procedure~\cite{holm1979simple} over all CSF pairs of a cell can only add edges to the indifference graph. On the 22-CSF inputs of Figure~\ref{fig:top_clique_vgg13}, the Holm top clique contains the reported top clique in all 48 OOD cells and equals it in 14; the mean clique size grows from $2.8$ to $5.0$ (VGG-13 $2.9$ to $5.2$, ViT $2.5$ to $4.3$). The near-OOD cliques broaden the most: on VGG-13, CTM joins 10 of the 12 near cliques instead of 6. The far cliques stay geometry-aware (fDBD in 9 of 12, Energy and MLS still in 3), and KPCA RecError still anchors three of the four ViT mid cliques. NCI enters 17 OOD cliques instead of 9, and Maha++ 8 instead of 2. The same holds for the per-configuration cliques that train the predictor (Section~\ref{methods:prediction}): the Holm clique contains the reported one in 168 of 168 near, mid, and far cells (mean size $2.7$ to $4.6$). Reported memberships, including the predictor's training labels, therefore hold under family-wise control; the exclusion of a CSF from a reported clique, and hence the sharpness of the near-OOD cliques, is not multiplicity-controlled.

\subsection{Clique stability under a threshold metric}\label{appendix:fpr95}
The top cliques of Figure~\ref{fig:top_clique_vgg13} use AURC and AUGRC blocks, which integrate over all thresholds. To test whether a threshold metric gives different winners, we rerun the same Friedman--Conover clique pipeline with FPR@95TPR blocks, and with paired failure-AUROC and FPR@95TPR blocks, on the 20 CSFs without Maha++ and NCI ($48$ cells: four panels, four sources, three OOD regimes). For each cell we compare the top clique with the AUGRC top clique.

\begin{table}[h]
    \centering
    \scriptsize
    \caption{Top cliques under threshold-based blocks against the AUGRC top cliques, per OOD regime ($16$ cells each). Jaccard = mean Jaccard index of the two cliques; Winner in = fraction of cells in which the AUGRC winner is in the other top clique; Size and Feature share = mean clique size and mean fraction of feature-side CSFs in the clique.}
    \label{tab:fpr95_cliques}
    \begin{tabular}{llcccccc}
    \toprule
    \textbf{Blocks} & \textbf{Regime} & \textbf{Jaccard} & \textbf{Winner in} & \textbf{Size (AUGRC)} & \textbf{Size (other)} & \textbf{Feat.\ share (AUGRC)} & \textbf{Feat.\ share (other)} \\
    \midrule
    FPR@95                & near & $0.45$ & $0.75$ & $3.5$ & $5.1$ & $0.49$ & $0.52$ \\
                          & mid  & $0.39$ & $0.88$ & $1.6$ & $3.9$ & $0.64$ & $0.42$ \\
                          & far  & $0.47$ & $0.88$ & $2.5$ & $3.4$ & $0.81$ & $0.61$ \\
    \midrule
    AUROC + FPR@95        & near & $0.48$ & $0.56$ & $3.5$ & $3.4$ & $0.49$ & $0.43$ \\
                          & mid  & $0.53$ & $0.88$ & $1.6$ & $2.9$ & $0.64$ & $0.46$ \\
                          & far  & $0.53$ & $0.81$ & $2.5$ & $2.9$ & $0.81$ & $0.54$ \\
    \bottomrule
    \end{tabular}
\end{table}

Over all $48$ cells, the AUGRC winner is in the FPR@95 top clique in $83\%$ of the cells, and in the AUROC + FPR@95 top clique in $75\%$. FPR@95 cliques are larger ($4.1$ vs $2.5$ members on average): one operating point resolves fewer ties, so the threshold metric mostly declares more winners rather than different ones. Per panel, the Jaccard index under AUROC + FPR@95 blocks is $0.63$--$0.73$ for ConfidNet, $0.60$--$0.68$ for DeVries, $0.41$--$0.75$ for ViT, and $0.14$--$0.26$ for Deep Gamblers across the three regimes. Deep Gamblers is the most metric-sensitive panel, and it is also the stratum with the weakest NC association (Appendix~\ref{appendix:mantel}). The share of feature-side CSFs in the top cliques is higher at far than at near under both metrics ($0.52$ to $0.61$ under FPR@95, $0.49$ to $0.81$ under AUGRC), but under FPR@95 the mid regime has the lowest share ($0.42$).

\section{Projection-Filtering: Full Results}\label{appendix:projection_full}
    Tables~\ref{tab:projection_full_conv} and~\ref{tab:projection_full_vit} report the paired AUGRC comparison for every base CSF and every projection variant. For each pair, we compute the AUGRC difference (base $-$ variant) across all source$\times$OOD dataset$\times$paradigm$\times$run combinations. A positive $\Delta$AUGRC means the projection variant is better (lower AUGRC). Significance is assessed with a two-sided Wilcoxon signed-rank test. Rows are sorted by $\Delta$AUGRC within each base CSF.

{\scriptsize
\setlength{\tabcolsep}{3pt}
\begin{longtable}{@{}l l r r r r r l@{}}
\caption{Full paired AUGRC comparison: base CSF vs.\ projection variants on \textbf{VGG-13 (Conv)}. $\Delta$AUGRC = mean(base $-$ variant); positive = variant better. Significant improvements ($p < 0.05$, variant better) are marked with $\checkmark$.}\label{tab:projection_full_conv}\\
\toprule
Base & Variant & $\Delta$AUGRC & Median $\Delta$ & Win & Loss & $n$ & $p$ \\
\midrule
\endfirsthead
\multicolumn{8}{l}{\tablename~\thetable{} \textit{(continued from previous page)}}\\
\toprule
Base & Variant & $\Delta$AUGRC & Median $\Delta$ & Win & Loss & $n$ & $p$ \\
\midrule
\endhead
\midrule
\multicolumn{8}{r}{\textit{continued on next page}}\\
\endfoot
\bottomrule
\endlastfoot
PCA RecError & class pred & $+38.6$ & $+44.9$ & 32 & 0 & 32 & $<$0.001 $\checkmark$ \\
PCA RecError & class & $+37.4$ & $+43.0$ & 32 & 0 & 32 & $<$0.001 $\checkmark$ \\
PCA RecError & class avg & $-116.1$ & $-119.6$ & 0 & 40 & 40 & $<$0.001 \\
\addlinespace
Maha & global & $+24.1$ & $+16.6$ & 50 & 14 & 64 & $<$0.001 $\checkmark$ \\
Maha & class pred & $+18.7$ & $+10.2$ & 40 & 16 & 56 & $<$0.001 $\checkmark$ \\
Maha & class avg & $+5.22$ & $+0.24$ & 38 & 18 & 56 & 0.221 \\
\addlinespace
pNML & class pred & $+6.73$ & $+5.12$ & 40 & 8 & 48 & $<$0.001 $\checkmark$ \\
pNML & global & $-18.3$ & $-18.1$ & 2 & 38 & 40 & $<$0.001 \\
pNML & class avg & $-67.9$ & $-74.3$ & 0 & 16 & 16 & $<$0.001 \\
\addlinespace
GradNorm & class pred & $+1.13$ & $+0.28$ & 67 & 21 & 88 & $<$0.001 $\checkmark$ \\
GradNorm & global & $+0.61$ & $+0.14$ & 60 & 36 & 96 & 0.001 $\checkmark$ \\
GradNorm & class avg & $-8.12$ & $-5.36$ & 8 & 72 & 80 & $<$0.001 \\
\addlinespace
CTM & class avg & $+0.36$ & $+0.14$ & 46 & 42 & 88 & 0.591 \\
CTM & global & $+0.34$ & $+0.21$ & 70 & 18 & 88 & $<$0.001 $\checkmark$ \\
CTM & class pred & $-0.67$ & $-0.49$ & 15 & 73 & 88 & $<$0.001 \\
CTM & class & $-0.71$ & $-0.63$ & 16 & 64 & 80 & $<$0.001 \\
\addlinespace
fDBD & global & $+0.20$ & $+0.19$ & 67 & 29 & 96 & 0.001 $\checkmark$ \\
fDBD & class avg & $-0.63$ & $-0.51$ & 39 & 49 & 88 & 0.126 \\
fDBD & class pred & $-0.69$ & $-0.33$ & 37 & 51 & 88 & 0.015 \\
\addlinespace
NNGuide & global & $+0.16$ & $+0.11$ & 63 & 33 & 96 & $<$0.001 $\checkmark$ \\
NNGuide & class pred & $-0.69$ & $-0.45$ & 18 & 78 & 96 & $<$0.001 \\
NNGuide & class avg & $-2.35$ & $-1.94$ & 13 & 59 & 72 & $<$0.001 \\
\addlinespace
REN & global & $+0.16$ & $+0.04$ & 52 & 36 & 88 & 0.028 $\checkmark$ \\
REN & class & $-0.20$ & $+0.23$ & 59 & 37 & 96 & 0.611 \\
REN & class pred & $-0.81$ & $-1.10$ & 17 & 79 & 96 & $<$0.001 \\
REN & class avg & $-1.85$ & $-0.58$ & 36 & 52 & 88 & 0.002 \\
\addlinespace
PCE & class & $+0.10$ & $+0.04$ & 51 & 37 & 88 & 0.044 $\checkmark$ \\
PCE & global & $+0.06$ & $+0.03$ & 58 & 38 & 96 & $<$0.001 $\checkmark$ \\
PCE & class pred & $-1.23$ & $-0.98$ & 9 & 71 & 80 & $<$0.001 \\
PCE & class avg & $-2.21$ & $-1.30$ & 23 & 41 & 64 & $<$0.001 \\
\addlinespace
MSR & global & $+0.07$ & $+0.04$ & 63 & 33 & 96 & $<$0.001 $\checkmark$ \\
MSR & class & $+0.07$ & $+0.03$ & 48 & 40 & 88 & 0.088 \\
MSR & class pred & $-1.33$ & $-1.01$ & 12 & 68 & 80 & $<$0.001 \\
MSR & class avg & $-2.30$ & $-1.42$ & 22 & 42 & 64 & $<$0.001 \\
\addlinespace
GEN & global & $+0.07$ & $+0.08$ & 68 & 28 & 96 & $<$0.001 $\checkmark$ \\
GEN & class & $+0.06$ & $+0.20$ & 60 & 28 & 88 & 0.024 $\checkmark$ \\
GEN & class pred & $-1.36$ & $-1.27$ & 3 & 69 & 72 & $<$0.001 \\
GEN & class avg & $-1.93$ & $-0.73$ & 27 & 53 & 80 & $<$0.001 \\
\addlinespace
PE & global & $+0.06$ & $+0.03$ & 59 & 29 & 88 & $<$0.001 $\checkmark$ \\
PE & class & $-0.35$ & $+0.04$ & 52 & 36 & 88 & 0.249 \\
PE & class pred & $-1.05$ & $-0.70$ & 7 & 73 & 80 & $<$0.001 \\
PE & class avg & $-2.62$ & $-1.77$ & 15 & 41 & 56 & $<$0.001 \\
\addlinespace
GE & global & $+0.05$ & $+0.03$ & 65 & 31 & 96 & $<$0.001 $\checkmark$ \\
GE & class & $-0.89$ & $+0.01$ & 50 & 46 & 96 & 0.390 \\
GE & class pred & $-0.26$ & $+0.07$ & 38 & 34 & 72 & 0.606 \\
GE & class avg & $-3.80$ & $-2.89$ & 11 & 45 & 56 & $<$0.001 \\
\addlinespace
Energy & global & $+0.01$ & $-0.00$ & 47 & 49 & 96 & 0.855 \\
Energy & class pred & $-0.21$ & $+0.05$ & 54 & 42 & 96 & 1.000 \\
Energy & class & $-1.40$ & $-0.68$ & 10 & 86 & 96 & $<$0.001 \\
Energy & class avg & $-5.51$ & $-4.50$ & 3 & 45 & 48 & $<$0.001 \\
\addlinespace
MLS & global & $+0.01$ & $+0.00$ & 49 & 47 & 96 & 0.698 \\
MLS & class & $-0.80$ & $-0.37$ & 18 & 78 & 96 & $<$0.001 \\
MLS & class pred & $-0.78$ & $-0.35$ & 16 & 80 & 96 & $<$0.001 \\
MLS & class avg & $-4.23$ & $-3.29$ & 7 & 49 & 56 & $<$0.001 \\
\addlinespace
KPCA RecError & class pred & $+0.74$ & $+0.33$ & 38 & 34 & 72 & 0.299 \\
KPCA RecError & class & $-0.51$ & $-0.46$ & 30 & 34 & 64 & 0.403 \\
KPCA RecError & class avg & $-161.9$ & $-158.6$ & 0 & 24 & 24 & $<$0.001 \\
\end{longtable}
}

{\scriptsize
\setlength{\tabcolsep}{3pt}
\begin{longtable}{@{}l l r r r r r l@{}}
\caption{Full paired AUGRC comparison: base CSF vs.\ projection variants on \textbf{ViT}. Same format as Table~\ref{tab:projection_full_conv}.}\label{tab:projection_full_vit}\\
\toprule
Base & Variant & $\Delta$AUGRC & Median $\Delta$ & Win & Loss & $n$ & $p$ \\
\midrule
\endfirsthead
\multicolumn{8}{l}{\tablename~\thetable{} \textit{(continued from previous page)}}\\
\toprule
Base & Variant & $\Delta$AUGRC & Median $\Delta$ & Win & Loss & $n$ & $p$ \\
\midrule
\endhead
\midrule
\multicolumn{8}{r}{\textit{continued on next page}}\\
\endfoot
\bottomrule
\endlastfoot
PCA RecError & class pred & $+11.6$ & $+3.07$ & 26 & 6 & 32 & $<$0.001 $\checkmark$ \\
PCA RecError & class & $+10.4$ & $+2.97$ & 25 & 7 & 32 & $<$0.001 $\checkmark$ \\
PCA RecError & class avg & $-41.5$ & $-28.6$ & 0 & 24 & 24 & $<$0.001 \\
\addlinespace
GradNorm & class pred & $+5.91$ & $+5.68$ & 25 & 7 & 32 & $<$0.001 $\checkmark$ \\
GradNorm & global & $+4.92$ & $+3.76$ & 22 & 10 & 32 & 0.001 $\checkmark$ \\
GradNorm & class avg & $-40.7$ & $-28.6$ & 10 & 22 & 32 & $<$0.001 \\
\addlinespace
REN & class pred & $+4.74$ & $+2.10$ & 24 & 8 & 32 & 0.001 $\checkmark$ \\
REN & global & $-3.59$ & $-0.31$ & 13 & 19 & 32 & 0.057 \\
REN & class & $-0.66$ & $-1.13$ & 10 & 22 & 32 & 0.144 \\
REN & class avg & $-3.11$ & $-4.95$ & 9 & 23 & 32 & 0.028 \\
\addlinespace
MLS & global & $+0.65$ & $+0.35$ & 27 & 5 & 32 & $<$0.001 $\checkmark$ \\
MLS & class & $-0.03$ & $+0.03$ & 20 & 12 & 32 & 0.651 \\
MLS & class pred & $-0.01$ & $+0.02$ & 20 & 12 & 32 & 0.599 \\
MLS & class avg & $-10.4$ & $-10.6$ & 2 & 30 & 32 & $<$0.001 \\
\addlinespace
GE & global & $+0.64$ & $+0.34$ & 23 & 9 & 32 & $<$0.001 $\checkmark$ \\
GE & class pred & $-0.14$ & $-0.03$ & 13 & 19 & 32 & 0.561 \\
GE & class & $-1.12$ & $-0.72$ & 0 & 32 & 32 & $<$0.001 \\
GE & class avg & $-7.35$ & $-6.28$ & 1 & 31 & 32 & $<$0.001 \\
\addlinespace
Energy & global & $+0.63$ & $+0.46$ & 28 & 4 & 32 & $<$0.001 $\checkmark$ \\
Energy & class & $+0.04$ & $+0.03$ & 20 & 12 & 32 & 0.586 \\
Energy & class pred & $+0.03$ & $+0.06$ & 21 & 11 & 32 & 0.379 \\
Energy & class avg & $-12.3$ & $-12.1$ & 3 & 29 & 32 & $<$0.001 \\
\addlinespace
PE & global & $+0.55$ & $+0.31$ & 23 & 9 & 32 & $<$0.001 $\checkmark$ \\
PE & class pred & $+0.09$ & $-0.03$ & 15 & 17 & 32 & 0.818 \\
PE & class & $-1.11$ & $-0.64$ & 1 & 31 & 32 & $<$0.001 \\
PE & class avg & $-5.72$ & $-4.95$ & 1 & 31 & 32 & $<$0.001 \\
\addlinespace
MSR & global & $+0.55$ & $+0.33$ & 22 & 10 & 32 & $<$0.001 $\checkmark$ \\
MSR & class pred & $+0.22$ & $-0.01$ & 16 & 16 & 32 & 0.512 \\
MSR & class & $-0.98$ & $-0.52$ & 2 & 30 & 32 & $<$0.001 \\
MSR & class avg & $-5.58$ & $-5.14$ & 2 & 30 & 32 & $<$0.001 \\
\addlinespace
PCE & global & $+0.54$ & $+0.31$ & 21 & 11 & 32 & 0.002 $\checkmark$ \\
PCE & class pred & $+0.16$ & $-0.02$ & 16 & 16 & 32 & 0.638 \\
PCE & class & $-1.00$ & $-0.55$ & 2 & 30 & 32 & $<$0.001 \\
PCE & class avg & $-5.48$ & $-5.04$ & 2 & 30 & 32 & $<$0.001 \\
\addlinespace
GEN & global & $+0.50$ & $+0.28$ & 27 & 5 & 32 & $<$0.001 $\checkmark$ \\
GEN & class pred & $+0.13$ & $+0.03$ & 19 & 13 & 32 & 0.599 \\
GEN & class & $-1.03$ & $-0.43$ & 4 & 28 & 32 & $<$0.001 \\
GEN & class avg & $-4.65$ & $-4.30$ & 3 & 29 & 32 & $<$0.001 \\
\addlinespace
fDBD & global & $+0.16$ & $-0.17$ & 13 & 19 & 32 & 0.692 \\
fDBD & class avg & $+0.67$ & $-0.40$ & 9 & 15 & 24 & 0.944 \\
fDBD & class pred & $-3.46$ & $-3.28$ & 0 & 32 & 32 & $<$0.001 \\
\addlinespace
CTM & global & $-0.28$ & $-0.33$ & 13 & 19 & 32 & 0.295 \\
CTM & class & $-3.16$ & $-2.13$ & 4 & 28 & 32 & $<$0.001 \\
CTM & class pred & $-3.04$ & $-2.14$ & 3 & 29 & 32 & $<$0.001 \\
CTM & class avg & $-4.69$ & $-3.03$ & 2 & 30 & 32 & $<$0.001 \\
\addlinespace
NNGuide & global & $-0.08$ & $-0.07$ & 14 & 18 & 32 & 0.678 \\
NNGuide & class pred & $-0.87$ & $-0.44$ & 8 & 24 & 32 & 0.012 \\
NNGuide & class avg & $-9.56$ & $-9.07$ & 1 & 31 & 32 & $<$0.001 \\
\addlinespace
Maha & global & $-3.56$ & $-2.01$ & 9 & 23 & 32 & 0.098 \\
Maha & class avg & $-1.58$ & $-0.44$ & 9 & 23 & 32 & 0.062 \\
Maha & class pred & $-20.0$ & $-9.92$ & 2 & 30 & 32 & $<$0.001 \\
\addlinespace
pNML & class pred & $+1.39$ & $-1.24$ & 9 & 23 & 32 & 0.390 \\
pNML & global & $-20.2$ & $-21.8$ & 2 & 30 & 32 & $<$0.001 \\
pNML & class avg & $-20.2$ & $-18.8$ & 1 & 31 & 32 & $<$0.001 \\
\addlinespace
KPCA RecError & class pred & $-5.35$ & $+0.54$ & 18 & 14 & 32 & 0.548 \\
KPCA RecError & class & $-9.20$ & $+0.08$ & 16 & 16 & 32 & 0.203 \\
KPCA RecError & class avg & $-108.1$ & $-122.7$ & 2 & 30 & 32 & $<$0.001 \\
\end{longtable}
}

Several patterns in these full tables are worth noting. First, \textbf{class avg is almost universally harmful}: across both architectures, the class-averaged projection variant significantly degrades performance for the vast majority of CSFs, often by large margins. This variant averages the projected representations across all class subspaces, which blurs class-specific structure rather than sharpening it. Second, on ViTs, \textbf{prototype-based CSFs (CTM, NNGuide) are hurt by all projection variants}, consistent with ViT representations being too far from collapse for prototype-based scoring to benefit from any subspace filtering. Third, on CNNs, \textbf{Energy and MLS show no significant gain from any projection variant} ($p > 0.69$ for global), suggesting that when the representation is already well collapsed, these logit-based scores are operating near their ceiling and projection filtering has little room to improve them.

\subsection{Mechanistic explanations for projection-filtering gains}\label{appendix:projection_mechanisms}

\subsubsection{CNNs: Why global projection lifts CTM and NNGuide on mid-OOD}
As Table~\ref{tab:nc_proj_vgg} shows, the NC metrics after Global Projection Filtering do not significantly change for the ID data. Therefore, the improvement in detection cannot be attributed to the ID classes becoming tighter or more separated from each other after filtering. Instead, the improvement comes from how the OOD data is transformed relative to the ID manifold after the projection. Since the improvement occurs for CTM and NNGuide, we can attribute it to the angular information as the main reason. Global Projection Filtering acts as a selective denoiser that benefits Mid-OOD detection by severing the shared low-level statistics, like texture or color, that confuse angular detectors. Mid-OOD samples often reside in the same positive feature cone as ID data. The projection operation $ \boldsymbol{P} \boldsymbol{P}^\top (\boldsymbol{h}^{\text{OOD}} - \boldsymbol{\mu})$ centers the data and removes the low-variance directions responsible for these spurious correlations. By stripping away these common-mode statistics, the projected Mid-OOD vector effectively collapses toward the global mean, becoming angularly ambiguous relative to the distinct class prototypes. This drastically lowers their cosine similarity scores in CSFs like CTM and NNGuide, reducing false positives. In contrast, Far-OOD samples do not benefit from this refinement because they are already geometrically distinct. These samples typically lie in the null space of the ID manifold or are orthogonal to the semantic class directions. Since their angular separation from ID prototypes is already maximized (they look nothing like the training classes), the projection step is redundant. Thus, the projection operation acts as an identity transformation regarding their detectability. However, this projection does not solve the problem of atypically large norms from Far-OOD samples. The principal components $\boldsymbol{P}$ often capture generic image statistics (e.g., contrast, brightness) shared by all natural images, not just semantic class features. If a Far-OOD sample has an anomalously large norm due to domain shift (e.g., high saturation), its projection onto these generic axes preserves this magnitude: $\lVert\hat{\boldsymbol{h}}^{\text{OOD}}\rVert \leq \lVert \boldsymbol{P} \boldsymbol{P}^\top \boldsymbol{h}^{\text{OOD}} \rVert + \lVert (\boldsymbol{I}-\boldsymbol{P} \boldsymbol{P}^\top) \boldsymbol{\mu} \rVert $. Consequently, magnitude-sensitive detectors like Energy or MLS can still fail on Far-OOD data if the norm is large enough to override the angular mismatch, necessitating the use of normalized scores like CTM (divides by $\lVert \boldsymbol{h}\rVert$) or fDBD (regularizes by $\lVert \boldsymbol{h}-\boldsymbol{\mu}\rVert$) which are invariant to this preserved scale.

\subsubsection{ViTs: Why global projection improves probabilistic and entropy-based scores}
In a finetuned ViT, the feature vector $\boldsymbol{h}$ has not fully collapsed. It contains a task-relevant component ($\boldsymbol{h}_{\text{task}}$) that drives the logits $\boldsymbol{g}$, and a large residual component ($\boldsymbol{h}_{\text{residual}}$) inherited from ImageNet pretraining that lies in the null space of the finetuning task: $\boldsymbol{h} = \boldsymbol{h}_{\text{task}} + \boldsymbol{h}_{\text{residual}}$. Since $\boldsymbol{h}_{\text{residual}}$ is orthogonal to the task weights $\boldsymbol{W}_{\text{task}}$, it does not affect the final prediction $\hat{y}$ for ID samples. However, it does affect the feature norm since $\|\boldsymbol{h}\|_2 = \sqrt{\|\boldsymbol{h}_{\text{task}}\|^2 + \|\boldsymbol{h}_{\text{residual}}\|^2}$ and the peakedness of the predictive distribution where $\boldsymbol{h}_{\text{residual}}$ often dominates for OOD samples. After global projection filtering, the influence of the subspaces that are more related to the task is preserved, while the subspaces that are not that relevant get discarded in the reconstruction $\boldsymbol{P}\boldsymbol{P}^\top\boldsymbol{h}_{\text{residual}}\approx 0$. While this process does not change the NC metrics relative to the unfiltered features, it sharpens the probability outputs over the finetuning classes, making them more discriminative for misclassification and OOD detection.

\subsubsection{ViTs: Why GradNorm global and CPP outperform vanilla GradNorm}
Vanilla GradNorm calculates the gradient of the KL divergence with respect to the last-layer weights $\boldsymbol{W}$: $\mathrm{GradNorm}(\boldsymbol{x})=\big\|\partial_{\boldsymbol{W}} \mathrm{KL}(\boldsymbol{u}\,\|\,p(\boldsymbol{x}))\big\|_p=\left\lVert\frac{1}{C}\sum_{k=1}^C\frac{\partial \mathcal{L}_{\text{CE}}\left(g\left(\boldsymbol{h}\right),k\right)}{\partial\boldsymbol{W}}\right\rVert_p$, typically using the $L_1$ norm. This expression can be factorized into a feature magnitude term $U = \|\boldsymbol{h}\|_1$ and an output-gradient term $V = \sum_{j=1}^C |1 - C \cdot \boldsymbol{p}_j|$, which is the sum of absolute differences between class probabilities and the uniform target.

As described previously, in a finetuned ViT, the raw feature $\boldsymbol{h}$ contains high-variance pretraining residuals inherited from ImageNet, which act as stochastic perturbations in the logit space. These residuals can move an ID sample closer to class boundaries or dilute the activation of the target class, spreading the softmax probabilities $\boldsymbol{p}_j$ and artificially lowering the output component $V$ at the same time the residual component artificially inflates the feature magnitude term $U$. By projecting features onto the principal subspace $\boldsymbol{P}$, global projection filtering aggressively prunes these irrelevant directions that do not support class-consistent structure. The filtered logit $\hat{\boldsymbol{g}}$ results in a higher dynamic range between the target class and off-target classes for ID samples, naturally amplifying the peakedness of the predictive distribution and increasing the discriminative score $V$. Conversely, for OOD samples, the projection $\boldsymbol{P} \boldsymbol{P}^\top$ discards the majority of their characteristic pretraining energy, forcing the feature toward the origin after centering. When projected, an OOD sample loses the spurious confidence provided by pretraining features, leading to a flatter, more uniform logit distribution $\hat{\boldsymbol{g}}$ where $\hat{\boldsymbol{p}}_j \approx 1/C$. This causes the output component $V$ to collapse toward zero, significantly expanding the safety gap between ID and OOD samples. Thus, projection filtering improves GradNorm by ensuring the output gradient is a true measure of semantic class sensitivity relative to relevant decision boundaries, rather than a noisy response to pretraining variance.

While Global projection filtering removes the residuals, it might still retain the variance of all classes in the task. If classes are crowded (e.g., 100 classes in CIFAR-100), the global subspace $\boldsymbol{P}$ is still quite large (high rank), allowing an OOD sample to retain significant magnitude by aligning with the principal components of incorrect classes. Class-Predicted Projection solves this by projecting the feature $\boldsymbol{h}$ onto the specific subspace $\boldsymbol{P}_{\hat{y}}$ of the predicted class $\hat{y}$: $\hat{\boldsymbol{h}}_{\hat{y}} = \boldsymbol{P}_{\hat{y}} \boldsymbol{P}_{\hat{y}}^\top (\boldsymbol{h} - \boldsymbol{\mu}_{\hat{y}}) + \boldsymbol{\mu}_{\hat{y}}$. This operation is much more aggressive. It discards not only the pretraining noise (null space) but also the variance directions associated with all $K-1$ other classes. For an ID sample correctly predicted as class $\hat{y}$, this preserves the signal perfectly because the sample lies in that specific subspace. However, for an OOD sample (or a confused sample) that aligned weakly with $\hat{y}$ only by chance, projecting it onto this narrow, class-specific manifold destroys its magnitude almost entirely.

\subsubsection{ViTs: Why KPCA RecError and CPP are competitive}

Kernel PCA (KPCA) excels at OOD detection for finetuned ViTs because it replaces the rigid linear assumptions of Neural Collapse with a flexible manifold matching approach. Standard detectors like CTM or fDBD assume that In-Distribution (ID) features collapse into simple, linearly separable clusters (Simplex ETF). However, finetuned ViTs retain a rich, complex geometry from pretraining that violates these assumptions. KPCA addresses this by mapping input features $\boldsymbol{h}$ into a high-dimensional Reproducing Kernel Hilbert Space via $\psi(\boldsymbol{h})$ and identifying the principal subspace $V$ that captures the intrinsic non-linear structure of the ID data. The detection metric is the reconstruction error, $e(\boldsymbol{x}) = \|\psi(\boldsymbol{h}) - \mathcal{P}_V \psi(\boldsymbol{h})\|^2$, which quantifies how well a test sample fits this learned manifold rather than measuring its distance to a potentially misaligned centroid. 

KPCA employs the same Cosine-Gaussian kernel introduced in Appendix~\ref{appendix:cfs_variations} to handle the specific geometric irregularities of finetuned representations. The cosine component neutralizes the magnitude variance common in finetuned models, ensuring the detector focuses purely on angular alignment. Simultaneously, the Gaussian component models local Euclidean distances on the hypersphere, allowing the subspace to wrap tightly around non-linear, sharpened decision boundaries. This enables KPCA to enclose complex class shapes that linear CSFs—which effectively fit a flat plane—would fail to capture, thereby reducing false positives from nearby OOD samples.

KPCA is distribution-agnostic regarding this global arrangement; it does not assume a specific prototype location ($\boldsymbol{\mu}_c$) but instead learns the aggregate manifold of all ID data. An OOD sample is detected not because it fails a specific angle test, but because its feature vector contains variance components from the pretraining distribution that are orthogonal to the finetuned task manifold. This ensures that OOD samples yield high reconstruction errors regardless of the symmetry or regularity of the ID class clusters. Similar to GradNorm class pred, KPCA class pred shows an improved performance because the class-predicted projection discards all variance components orthogonal to the predicted class $\hat{y}$, including generic image statistics and features from competing classes.

\begin{table}[htbp]
    \scriptsize
    \centering
    \caption{NC metrics under projection filtering (Global and Class pred). Compare against the unfiltered (None) values in Table~\ref{tab:nc_main}.}
    \label{tab:nc_proj_main}
\begin{subtable}[t]{\textwidth}
    \centering
    \subcaption{VGG-13 models.}
    \label{tab:nc_proj_vgg}
    \begin{tabular}{llrrrrr}
\toprule
Dataset & Projection & EqNorm & EqAng & max-EqAng & Var Collapse & Self Duality \\
\midrule
\multirow[c]{2}{*}{CIFAR-10} & Global & {\cellcolor[HTML]{452929}} \color[HTML]{F1F1F1} 0.0266 & {\cellcolor[HTML]{1E0000}} \color[HTML]{F1F1F1} 0.0651 & {\cellcolor[HTML]{FFFFFF}} \color[HTML]{000000} 0.2447 & {\cellcolor[HTML]{A76C6C}} \color[HTML]{F1F1F1} 0.0082 & {\cellcolor[HTML]{1E0000}} \color[HTML]{F1F1F1} 0.0307 \\
 & Class pred & {\cellcolor[HTML]{482B2B}} \color[HTML]{F1F1F1} 0.0268 & {\cellcolor[HTML]{1E0000}} \color[HTML]{F1F1F1} 0.0650 & {\cellcolor[HTML]{FFFFFF}} \color[HTML]{000000} 0.2447 & {\cellcolor[HTML]{A56B6B}} \color[HTML]{F1F1F1} 0.0080 & {\cellcolor[HTML]{2D1313}} \color[HTML]{F1F1F1} 0.0320 \\
 \midrule
\multirow[c]{2}{*}{SuperCIFAR-100} & Global & {\cellcolor[HTML]{1E0000}} \color[HTML]{F1F1F1} 0.0246 & {\cellcolor[HTML]{573636}} \color[HTML]{F1F1F1} 0.0679 & {\cellcolor[HTML]{CC9F8D}} \color[HTML]{F1F1F1} 0.1537 & {\cellcolor[HTML]{B77777}} \color[HTML]{F1F1F1} 0.0095 & {\cellcolor[HTML]{925E5E}} \color[HTML]{F1F1F1} 0.0507 \\
 & Class pred & {\cellcolor[HTML]{1E0000}} \color[HTML]{F1F1F1} 0.0248 & {\cellcolor[HTML]{573636}} \color[HTML]{F1F1F1} 0.0678 & {\cellcolor[HTML]{CC9F8D}} \color[HTML]{F1F1F1} 0.1537 & {\cellcolor[HTML]{B57575}} \color[HTML]{F1F1F1} 0.0093 & {\cellcolor[HTML]{966161}} \color[HTML]{F1F1F1} 0.0519 \\
 \midrule
\multirow[c]{2}{*}{CIFAR-100} & Global & {\cellcolor[HTML]{FFFFFF}} \color[HTML]{000000} 0.0745 & {\cellcolor[HTML]{FFFFFF}} \color[HTML]{000000} 0.1052 & {\cellcolor[HTML]{6D4545}} \color[HTML]{F1F1F1} 0.0952 & {\cellcolor[HTML]{FFFFFF}} \color[HTML]{000000} 0.0259 & {\cellcolor[HTML]{FBFBF3}} \color[HTML]{000000} 0.1232 \\
 & Class pred & {\cellcolor[HTML]{FFFFFF}} \color[HTML]{000000} 0.0743 & {\cellcolor[HTML]{FEFEFD}} \color[HTML]{000000} 0.1049 & {\cellcolor[HTML]{6D4545}} \color[HTML]{F1F1F1} 0.0950 & {\cellcolor[HTML]{966161}} \color[HTML]{F1F1F1} 0.0069 & {\cellcolor[HTML]{FFFFFF}} \color[HTML]{000000} 0.1278 \\
 \midrule
\multirow[c]{2}{*}{TinyImageNet} & Global & {\cellcolor[HTML]{DAC6A1}} \color[HTML]{000000} 0.0546 & {\cellcolor[HTML]{DDCEA5}} \color[HTML]{000000} 0.0903 & {\cellcolor[HTML]{1E0000}} \color[HTML]{F1F1F1} 0.0763 & {\cellcolor[HTML]{6F4646}} \color[HTML]{F1F1F1} 0.0044 & {\cellcolor[HTML]{3C2222}} \color[HTML]{F1F1F1} 0.0335 \\
 & Class pred & {\cellcolor[HTML]{DAC6A1}} \color[HTML]{000000} 0.0546 & {\cellcolor[HTML]{DDCCA4}} \color[HTML]{000000} 0.0901 & {\cellcolor[HTML]{1E0000}} \color[HTML]{F1F1F1} 0.0762 & {\cellcolor[HTML]{1E0000}} \color[HTML]{F1F1F1} 0.0016 & {\cellcolor[HTML]{2D1313}} \color[HTML]{F1F1F1} 0.0322 \\
\bottomrule
\end{tabular}
\end{subtable}
    \hfill
\begin{subtable}[t]{\textwidth}
    \centering
    \subcaption{ViT models.}
    \label{tab:nc_proj_vit}
    \begin{tabular}{llrrrrr}
\toprule
Dataset & Projection & EqNorm & EqAng & max-EqAng & Var Collapse & Self Duality \\
\midrule
\multirow[c]{2}{*}{CIFAR-10} & Global & {\cellcolor[HTML]{1E0000}} \color[HTML]{F1F1F1} 0.0870 & {\cellcolor[HTML]{FFFFFF}} \color[HTML]{000000} 0.2505 & {\cellcolor[HTML]{FFFFFF}} \color[HTML]{000000} 0.3771 & {\cellcolor[HTML]{321A1A}} \color[HTML]{F1F1F1} 0.0120 & {\cellcolor[HTML]{FFFFFF}} \color[HTML]{000000} 1.2716 \\
 & Class pred & {\cellcolor[HTML]{1E0000}} \color[HTML]{F1F1F1} 0.0870 & {\cellcolor[HTML]{FFFFFF}} \color[HTML]{000000} 0.2505 & {\cellcolor[HTML]{FFFFFF}} \color[HTML]{000000} 0.3770 & {\cellcolor[HTML]{321A1A}} \color[HTML]{F1F1F1} 0.0120 & {\cellcolor[HTML]{FFFFFE}} \color[HTML]{000000} 1.2693 \\
 \midrule
\multirow[c]{2}{*}{SuperCIFAR-100} & Global & {\cellcolor[HTML]{CDA28F}} \color[HTML]{000000} 0.1432 & {\cellcolor[HTML]{F8F8E8}} \color[HTML]{000000} 0.2403 & {\cellcolor[HTML]{DDCEA5}} \color[HTML]{000000} 0.2814 & {\cellcolor[HTML]{553434}} \color[HTML]{F1F1F1} 0.0153 & {\cellcolor[HTML]{D8C09E}} \color[HTML]{000000} 1.0937 \\
 & Class pred & {\cellcolor[HTML]{CDA28F}} \color[HTML]{000000} 0.1435 & {\cellcolor[HTML]{F8F8E9}} \color[HTML]{000000} 0.2406 & {\cellcolor[HTML]{DDCEA5}} \color[HTML]{000000} 0.2816 & {\cellcolor[HTML]{523232}} \color[HTML]{F1F1F1} 0.0150 & {\cellcolor[HTML]{D8BF9E}} \color[HTML]{000000} 1.0929 \\
 \midrule
\multirow[c]{2}{*}{CIFAR-100} & Global & {\cellcolor[HTML]{815252}} \color[HTML]{F1F1F1} 0.1058 & {\cellcolor[HTML]{230606}} \color[HTML]{F1F1F1} 0.1328 & {\cellcolor[HTML]{1E0000}} \color[HTML]{F1F1F1} 0.1177 & {\cellcolor[HTML]{5E3A3A}} \color[HTML]{F1F1F1} 0.0162 & {\cellcolor[HTML]{230606}} \color[HTML]{F1F1F1} 0.8566 \\
 & Class pred & {\cellcolor[HTML]{825353}} \color[HTML]{F1F1F1} 0.1065 & {\cellcolor[HTML]{230606}} \color[HTML]{F1F1F1} 0.1327 & {\cellcolor[HTML]{1E0000}} \color[HTML]{F1F1F1} 0.1177 & {\cellcolor[HTML]{1E0000}} \color[HTML]{F1F1F1} 0.0107 & {\cellcolor[HTML]{1E0000}} \color[HTML]{F1F1F1} 0.8548 \\
 \midrule
\multirow[c]{2}{*}{TinyImageNet} & Global & {\cellcolor[HTML]{F8F8EA}} \color[HTML]{000000} 0.1971 & {\cellcolor[HTML]{D8C19F}} \color[HTML]{000000} 0.2008 & {\cellcolor[HTML]{8D5A5A}} \color[HTML]{F1F1F1} 0.1665 & {\cellcolor[HTML]{FFFFFF}} \color[HTML]{000000} 0.0806 & {\cellcolor[HTML]{E4DDAD}} \color[HTML]{000000} 1.1429 \\
 & Class pred & {\cellcolor[HTML]{FFFFFF}} \color[HTML]{000000} 0.2069 & {\cellcolor[HTML]{DDCEA5}} \color[HTML]{000000} 0.2066 & {\cellcolor[HTML]{935F5F}} \color[HTML]{F1F1F1} 0.1716 & {\cellcolor[HTML]{A46A6A}} \color[HTML]{F1F1F1} 0.0289 & {\cellcolor[HTML]{E3DCAD}} \color[HTML]{000000} 1.1416 \\
\bottomrule
\end{tabular}
\end{subtable}
\end{table}

\subsection{Activation wrappers: ReAct and ASH-S}\label{appendix:react_ash}
ReAct~\cite{sun2021react} and ASH~\cite{djurisic2023extremely} also modify the representation before scoring, but with one global rule for all classes. We evaluate both on the CNN pools as wrappers of the nine head-side CSFs that read logits or softmax outputs. Both act on the feature map before the last max-pooling layer. ReAct clips the activations at the 90th percentile of the ID validation activations of each checkpoint; ASH-S keeps the largest $35\%$ of the activations of each sample, sets the rest to zero, and rescales the result. Our hook is a convolutional feature map, which the ViT does not have, so the wrappers are not evaluated on the ViT. For each (model, OOD set, CSF) we pair the wrapped AUGRC with the base AUGRC of the same CSF.

\begin{table}[h]
    \centering
    \scriptsize
    \caption{ReAct and ASH-S as wrappers of the nine head-side CSFs. A row is one model on one OOD dataset ($2{,}240$ rows per CSF for the VGG-13 models, $448$ for the ResNet-18 models). In each cell, the first number is the mean over the rows of the paired change in AUGRC ($\times 10^{3}$, wrapper~$-$~base; positive = worse). The number in parentheses is the share of rows on which the wrapped CSF has a lower AUGRC than the base CSF. The mean base AUGRC is $229$--$252$ on VGG-13 and $222$--$255$ on ResNet-18.}
    \label{tab:react_ash}
    \begin{tabular}{lcccc}
    \toprule
     & \multicolumn{2}{c}{VGG-13} & \multicolumn{2}{c}{ResNet-18} \\
    \cmidrule(lr){2-3}\cmidrule(lr){4-5}
    CSF & ReAct & ASH-S & ReAct & ASH-S \\
    \midrule
    Energy   & $+6.45$ ($33\%$) & $+0.04$ ($61\%$) & $+8.95$ ($21\%$) & $-0.03$ ($69\%$) \\
    GE       & $+6.29$ ($32\%$) & $+0.02$ ($60\%$) & $+12.50$ ($21\%$) & $-0.02$ ($68\%$) \\
    GEN      & $+3.73$ ($33\%$) & $+0.02$ ($55\%$) & $+6.05$ ($28\%$) & $+0.13$ ($56\%$) \\
    GradNorm & $+43.39$ ($21\%$) & $+0.08$ ($54\%$) & $+84.10$ ($1\%$) & $+1.10$ ($44\%$) \\
    MLS      & $+4.34$ ($30\%$) & $+0.03$ ($59\%$) & $+13.18$ ($21\%$) & $-0.01$ ($65\%$) \\
    MSR      & $+4.71$ ($24\%$) & $+0.02$ ($56\%$) & $+13.16$ ($20\%$) & $+0.02$ ($59\%$) \\
    PCE      & $+4.27$ ($25\%$) & $+0.02$ ($58\%$) & $+10.39$ ($21\%$) & $+0.01$ ($59\%$) \\
    PE       & $+4.52$ ($28\%$) & $+0.03$ ($60\%$) & $+9.91$ ($22\%$) & $-0.01$ ($64\%$) \\
    REN      & $+3.81$ ($30\%$) & $-0.10$ ($48\%$) & $+5.43$ ($36\%$) & $+0.09$ ($56\%$) \\
    \bottomrule
    \end{tabular}
\end{table}

Table~\ref{tab:react_ash} shows that ASH-S is close to neutral: its mean change is below $0.2$ for every CSF except GradNorm on ResNet-18, and it is better on more than half of the rows for 16 of the 18 (architecture, CSF) pairs. ReAct raises the mean AUGRC of every wrapped CSF on both architectures, most for GradNorm. Its effect depends on the source: pooled over the nine CSFs on VGG-13, the mean change is $+5.1$ on CIFAR-10 and CIFAR-100, $+1.4$ on SuperCIFAR-100 (median $-0.02$), and $+29.1$ on TinyImageNet.

To test whether the wrappers change the competitive set, we replace the nine base CSFs of the Figure~\ref{fig:top_clique_vgg13} inputs with their wrapped variants, one wrapper at a time, and recompute the 36 VGG-13 OOD top cliques with the same pipeline (each variant's configuration is selected on ID validation data like every CSF). We swap rather than add the variants because the Friedman--Conover result depends on the size of the roster, and an added near-duplicate variant ties with its base CSF. Without a swap, the pipeline reproduces the cliques of Figure~\ref{fig:top_clique_vgg13} in all 36 cells. With ASH-S, a wrapped variant enters the top clique in 4 cells where its base CSF was not, and a base member drops out in 4 cells. With ReAct, wrapped variants enter in 8 cells, all on CIFAR-100 and SuperCIFAR-100, where ReAct-wrapped Energy, GE, MLS, and REN join or replace the top clique; on Deep Gamblers CIFAR-100 near, they replace the feature-side members (CTM, NCI, NNGuide, NeCo, fDBD). Base members drop out in 8 cells, for example MLS on three CIFAR-10 cells. The wrapped variants are not candidates of the predictor.

\section{Mantel Test: NC Geometry vs.\ Method Rankings}\label{appendix:mantel}
    The Mantel test~\cite{mantel1967detection} assesses whether two distance matrices computed over the same set of objects are correlated. In our setting, the objects are model configurations (backbone $\times$ dataset $\times$ training paradigm $\times$ run), and the two matrices encode NC geometry and CSF rankings, respectively. A significant positive correlation means that configurations with similar NC profiles also produce similar detector rankings.

\paragraph{NC distance matrix.}
Each model configuration $i$ is described by a vector $\boldsymbol{\phi}^{\text{NC}}_i \in \mathbb{R}^8$ containing the 8 Papyan NC metrics defined in Appendix~\ref{appendix:neural_collapse}. Because the metrics have different natural scales (e.g., variability collapse $\sim 10^{-2}$, equinormness $\sim 10^{-1}$), we first standardize each metric to zero mean and unit variance across configurations. The NC distance between configurations $i$ and $j$ is then the Euclidean distance between the standardized vectors:
\begin{equation}
    D^{\text{NC}}_{ij} = \left\lVert \frac{\boldsymbol{\phi}^{\text{NC}}_i - \bar{\boldsymbol{\phi}}^{\text{NC}}}{\boldsymbol{s}} - \frac{\boldsymbol{\phi}^{\text{NC}}_j - \bar{\boldsymbol{\phi}}^{\text{NC}}}{\boldsymbol{s}} \right\rVert_2,
\end{equation}
where $\bar{\boldsymbol{\phi}}^{\text{NC}}$ and $\boldsymbol{s}$ are the column-wise mean and standard deviation. This produces a symmetric $n \times n$ matrix $\boldsymbol{D}^{\text{NC}}$.

\paragraph{CSF-ranking distance matrix.}
For each model configuration $i$, we rank the CSFs of Section~\ref{results:experimental_setup} (22 on VGG-13, 21 on the fine-tuned ViT) by their AUGRC score (lower is better), yielding a rank vector $\boldsymbol{r}_i \in \mathbb{R}^m$ where $m$ is the number of CSFs. The distance between two configurations is computed as the Spearman rank distance:
\begin{equation}
    D^{\text{rank}}_{ij} = 1 - \rho_{\text{S}}(\boldsymbol{r}_i, \boldsymbol{r}_j),
\end{equation}
where $\rho_{\text{S}}$ denotes the Spearman rank correlation. This distance ranges from 0 (identical rankings) to 2 (perfectly reversed rankings), and measures how similarly two model configurations order the set of detectors. The result is a symmetric $n \times n$ matrix $\boldsymbol{D}^{\text{rank}}$.

\paragraph{Permutation test.}
The Mantel test statistic is the Spearman correlation between the upper-triangular entries of the two distance matrices:
\begin{equation}
    r_{\text{obs}} = \rho_{\text{S}}\!\left(\text{vec}(\boldsymbol{D}^{\text{NC}}),\; \text{vec}(\boldsymbol{D}^{\text{rank}})\right),
\end{equation}
where $\text{vec}(\cdot)$ extracts the $\binom{n}{2}$ entries above the diagonal. Under the null hypothesis of no association, we generate a reference distribution by randomly permuting the row and column indices of $\boldsymbol{D}^{\text{rank}}$ simultaneously (preserving its internal structure while breaking any correspondence with $\boldsymbol{D}^{\text{NC}}$). At each permutation $k$, we draw a random index permutation $\pi_k$ and compute
\begin{equation}
    r_k = \rho_{\text{S}}\!\left(\text{vec}(\boldsymbol{D}^{\text{NC}}),\; \text{vec}(\boldsymbol{D}^{\text{rank}}_{\pi_k})\right),
\end{equation}
where $\boldsymbol{D}^{\text{rank}}_{\pi_k}$ denotes the matrix with rows and columns reordered by $\pi_k$. We use 9{,}999 permutations with a fixed random seed for reproducibility. The one-sided $p$-value is
\begin{equation}
    p = \frac{1 + \sum_{k=1}^{9{,}999} \mathbb{1}[r_k \geq r_{\text{obs}}]}{1 + 9{,}999},
\end{equation}
where the $+1$ terms provide a conservative estimate~\cite{phipson2010permutation}.

\paragraph{Interpretation.}
A significant positive $r_{\text{obs}}$ means that model configurations with similar NC geometry tend to produce similar CSF rankings. This is a global test: it does not specify \emph{which} NC metric or \emph{which} CSF drives the association. To identify the most informative NC metrics, we also run per-metric Mantel tests using a univariate distance matrix $D^{\text{NC}(j)}_{ij} = \big|\phi^{\text{NC},(j)}_i - \phi^{\text{NC},(j)}_j\big|$ for each metric $j$ separately. To identify which CSFs are most sensitive to NC geometry, we compute Spearman correlations between each CSF's rank vector and each NC metric's value vector across configurations.

\paragraph{Additional distance formulations.}
As a robustness check, we also compute the CSF distance matrix using Jaccard distance on top-$k$ sets. For a given $k$, we binarize each rank vector into an indicator $\boldsymbol{b}_i \in \{0,1\}^m$ where $b_{i,j} = 1$ if CSF $j$ has rank $\leq k$, and compute $D^{\text{Jaccard}}_{ij} = 1 - \frac{|\boldsymbol{b}_i \cap \boldsymbol{b}_j|}{|\boldsymbol{b}_i \cup \boldsymbol{b}_j|}$. We report results for $k \in \{1, 3, 5\}$. The Spearman rank distance and Jaccard formulations yield qualitatively consistent results (Table~\ref{tab:mantel_global}).

\paragraph{Results summary.}
Table~\ref{tab:mantel_global} reports the global Mantel test results across all four training paradigms. Table~\ref{tab:mantel_per_metric} reports the per-metric Mantel correlations, identifying which individual NC property is most predictive of CSF selection.

\begin{table}[h]
    \centering
    \scriptsize
    \caption{Global Mantel test: correlation between NC distance and CSF-ranking distance across training paradigms. All tests use 9{,}999 permutations.}
    \label{tab:mantel_global}
    \begin{tabular}{lcccc}
    \toprule
    \textbf{Paradigm} & $r$ \textbf{(rank dist.)} & $r$ \textbf{(Jaccard top-1)} & $r$ \textbf{(Jaccard top-3)} & $p$ \\
    \midrule
    Conv/ConfidNet     & 0.63 & 0.55 & 0.47 & $<$0.001 \\
    Conv/DeVries       & 0.62 & 0.27 & 0.51 & $<$0.001 \\
    Conv/Deep Gamblers & 0.37 & 0.44 & 0.39 & 0.001 \\
    ViT                & 0.70 & 0.40 & 0.56 & $<$0.001 \\
    \bottomrule
    \end{tabular}
\end{table}

\begin{table}[h]
    \centering
    \scriptsize
    \caption{Per-metric Mantel $r$ (single NC metric vs.\ Spearman rank distance). The strongest predictor in each paradigm is bolded. $^{\text{ns}}$ marks the entries that are not significant ($p > 0.05$); all other entries are significant at $p < 0.05$.}
    \label{tab:mantel_per_metric}
    \begin{tabular}{lcccc}
    \toprule
    \textbf{NC metric} & \textbf{ConfidNet} & \textbf{DeVries} & \textbf{Deep Gamblers} & \textbf{ViT} \\
    \midrule
    Var.\ collapse & 0.34 & 0.49 & 0.46 & 0.46 \\
    Equiangularity ($\boldsymbol{\mu}$) & 0.65 & 0.66 & 0.15 & 0.58 \\
    Equiangularity ($\boldsymbol{W}$) & 0.74 & 0.49 & \textbf{0.88} & 0.56 \\
    Equinormness ($\boldsymbol{\mu}$) & 0.51 & 0.45 & 0.12$^{\text{ns}}$ & 0.48 \\
    Equinormness ($\boldsymbol{W}$) & 0.33 & 0.21 & 0.07$^{\text{ns}}$ & 0.40 \\
    Max.\ equiang.\ ($\boldsymbol{\mu}$) & 0.71 & 0.51 & 0.13$^{\text{ns}}$ & 0.51 \\
    Max.\ equiang.\ ($\boldsymbol{W}$) & \textbf{0.77} & \textbf{0.68} & 0.74 & \textbf{0.77} \\
    Self-duality & 0.41 & 0.29 & 0.17 & 0.72 \\
    \bottomrule
    \end{tabular}
\end{table}

\subsection{Partial Mantel tests: convergence controls}\label{appendix:partial_mantel}
NC geometry and CSF rankings could be associated only because both follow how far training has converged. Convergence would then be a common cause of the two. NC is a property of the terminal phase of training: as training goes on, the within-class variation of the features shrinks, and the class means and the classifier weights move toward a simplex~\cite{papyan2020prevalence}. Convergence also changes the outputs that the CSFs read. A model that fits its training data more closely tends to be more accurate on ID data and to have larger logit margins, and ID accuracy is related to OOD detection performance~\cite{vaze2022open,krumpl2026one}. Two models that converged to different degrees can therefore differ both in NC geometry and in which CSFs work best on them. The Mantel test would then find an association even if the NC metrics carried no information beyond convergence, and a simpler summary of convergence, such as ID performance, would explain the rankings as well as the NC metrics do.

We test this with partial Mantel tests within each stratum. A partial Mantel test removes from both distance matrices the part that the covariates explain, and tests whether the association remains. We regress the rank-transformed NC and rank-distance matrices on the rank-transformed covariate distance matrices, and apply the permutation test above to the residuals (9{,}999 permutations, residual permutation). We do not measure convergence directly; the covariates are the convergence proxies in our stored results. ID performance is the ID-test AUGRC of MSR per model (and, as a variant, the mean ID-test AUGRC over all CSFs); a model that has converged further tends to have a lower ID-test AUGRC. Source identity ($0$ for two models with the same source, $1$ otherwise) reflects that the same schedule can take a model further on an easy source, such as CIFAR-10, than on a hard one, such as TinyImageNet with $200$ classes; it also removes every other difference between the sources. The learning rate varies only in the ViT stratum, where it changes how far fine-tuning moves from the pretrained weights. In the CNN strata the training recipe is fixed by design: SGD with base learning rate $0.1$ and cosine annealing, with $470$ epochs for ConfidNet (including its head-training stages at reduced learning rates), $250$ for DeVries, and $300$ for Deep Gamblers, the same for all four sources. Schedule and epoch count are therefore constants within a stratum and cannot cause the association there. Dropout also changes how closely a model fits its training data; Appendix~\ref{appendix:matched_pairs} tests its role separately. If the association vanishes under a control, the NC metrics add nothing beyond that proxy; if it remains, they carry information about the rankings that the proxy does not. Without covariates, the code reproduces Table~\ref{tab:mantel_global} exactly.

\begin{table}[h]
    \centering
    \scriptsize
    \caption{Partial Mantel tests (Spearman, 9{,}999 permutations): Mantel $r$ between NC distance and CSF-ranking distance after controlling for the listed covariates, with one-sided $p$-values. $p = 10^{-4}$ is the smallest value the permutation test can give. ID perf.\ = ID-test AUGRC of MSR; ID perf.\ (mean) = mean ID-test AUGRC over all CSFs.}
    \label{tab:partial_mantel}
    \begin{tabular}{lccccc}
    \toprule
    \textbf{Stratum} & \textbf{No control} & \textbf{ID perf.} & \textbf{ID perf.\ (mean)} & \textbf{Source} & \textbf{ID perf.\ + source} \\
    \midrule
    Conv/ConfidNet     & $0.63$ ($10^{-4}$) & $0.56$ ($10^{-4}$) & $0.46$ ($10^{-4}$) & $0.31$ ($4\times10^{-4}$) & $0.40$ ($10^{-4}$) \\
    Conv/DeVries       & $0.62$ ($10^{-4}$) & $0.56$ ($10^{-4}$) & $0.57$ ($10^{-4}$) & $0.37$ ($2\times10^{-4}$) & $0.51$ ($10^{-4}$) \\
    Conv/Deep Gamblers & $0.37$ ($0.001$) & $0.32$ ($8\times10^{-4}$) & $0.34$ ($0.002$) & $-0.08$ ($0.80$) & $0.36$ ($0.001$) \\
    ViT                & $0.70$ ($10^{-4}$) & $0.59$ ($10^{-4}$) & $0.59$ ($10^{-4}$) & $0.41$ ($2\times10^{-4}$) & $0.41$ ($2\times10^{-4}$) \\
    \bottomrule
    \end{tabular}
\end{table}

Table~\ref{tab:partial_mantel} shows that the control for ID performance keeps the association significant in all four strata. The source-identity control is stricter: it removes all variation between the four sources and keeps only the variation within a source. Under it the association survives in three strata and vanishes for Deep Gamblers, the weakest stratum in Table~\ref{tab:mantel_global}. With Papyan's NC1 in place of ours (Appendix~\ref{appendix:neural_collapse}), the same three strata survive this control ($r = 0.25$--$0.52$, $p \leq 0.004$ with raw and log10 values), and the ID-performance control stays significant in all four ($p \leq 0.013$). For ViT, adding the learning rate to the ID-performance control lowers $r$ from $0.59$ to $0.28$ ($p = 0.001$), and with ID performance, learning rate, and source together the association is no longer significant ($r = 0.09$, $p = 0.14$). These tests use the within-class variation metric of Appendix~\ref{appendix:neural_collapse}. A partial Mantel test removes only the covariates it is given, so it does not show that NC geometry causes the rankings; the causal test is to vary the degree of collapse during training at controlled strengths, for example with the methods of~\citet{harun2025controlling}.

\subsection{Matched pairs: training paradigm versus dropout}\label{appendix:matched_pairs}
The Mantel tests above pool two sources of variation: the training paradigm and the dropout setting. A matched-pair test separates them. We wrote its protocol and decision rules before the analysis; the changes made after it are listed below. The test uses the 20-CSF roster and the VGG-13 pool.

\paragraph{Protocol.} The unit is a \emph{slice}: one (source, paradigm entry, dropout) cell, where each Deep Gamblers reward counts as its own paradigm entry (6 entries for CIFAR-10, 7 for CIFAR-100, 9 for SuperCIFAR-100, and 6 for TinyImageNet, including ConfidNet and DeVries). For each slice we take the mean $z$-scored NC vector over its five runs and the top cliques of its four regimes (ID, near, mid, far), computed with the pipeline of Section~\ref{methods:statistical_tests} on the runs of that slice; the configuration is fixed by the slice instead of selected on validation data. For a pair of slices, $\Delta$NC is the Euclidean distance between their NC vectors and $\Delta$Clique is the mean Jaccard distance between their cliques over the regimes. A bootstrap over the five runs (200 resamples) flags unstable slice cliques (mean Jaccard below $0.60$), and their regimes are left out of $\Delta$Clique. The hypotheses are:
\begin{itemize}
    \item \textbf{H1 (paradigm axis).} Pairs that differ only in the paradigm entry, at fixed source and dropout. Supported if the Spearman $\rho$ between $\Delta$NC and $\Delta$Clique is at least $0.40$ and $p < 0.05$, where $p$ comes from a Mantel test that permutes the paradigm entries within each source (9{,}999 permutations), combined over the sources with Stouffer's method. The two dropout settings are pooled only if their $\rho$ differ by at most $0.10$.
    \item \textbf{H2 (dropout axis).} Pairs that differ only in the dropout setting, at fixed source and paradigm entry. Supported if $\rho \geq 0.30$ and $p < 0.05$, where $p$ comes from shuffling $\Delta$Clique within each source.
    \item \textbf{H3 (coverage).} Supported if at least $60\%$ of the H1 pairs in the top quartile of $\Delta$NC have different cliques ($\Delta$Clique $> 0$).
    \item \textbf{H4 (coordinates).} Which NC coordinates carry the H1 signal ($|\Delta x_j|$ against $\Delta$Clique, $\rho \geq 0.40$).
\end{itemize}

\paragraph{Changes after the protocol.} We excluded the SuperCIFAR-100 slices of Deep Gamblers at reward $2.2$: at this reward the model abstains too often, and its within-class variation is far larger than in any other slice ($1.46$ against at most $0.3$). This leaves 79 H1 pairs per dropout setting and 27 H2 pairs. The exclusion makes the result more conservative: with the slices included (87 pairs), H1 gives $\rho = 0.64$ and $0.56$, and H2 still fails ($\rho = 0.45$, $p = 0.79$). We report H4 per dropout setting, as the pooling rule requires, and only as a description, because its $p$-values come from shuffling $\Delta$Clique between pairs, which ignores that pairs share slices.

\begin{table}[h]
    \centering
    \scriptsize
    \caption{Matched-pair test on the VGG-13 pool (20-CSF roster). H1 per source: Mantel $p$ with paradigm-entry permutations; pooled H1: Stouffer combination over the sources.}
    \label{tab:matched_pairs}
    \begin{tabular}{llcccl}
    \toprule
    \textbf{Hypothesis} & \textbf{Scope} & $\rho$ & $p$ & $n$ & \textbf{Verdict} \\
    \midrule
    H1 (paradigm) & no dropout, pooled & $0.61$ & $6.4\times10^{-6}$ & 79 & Supported \\
     & \quad CIFAR-10 / CIFAR-100 & $0.38$ / $0.60$ & $0.055$ / $0.010$ & 15 / 21 & \\
     & \quad SuperCIFAR-100 / TinyImageNet & $0.88$ / $0.46$ & $<0.001$ / $0.11$ & 28 / 15 & \\
    H1 (paradigm) & dropout, pooled & $0.49$ & $1.8\times10^{-4}$ & 79 & Supported \\
     & \quad CIFAR-10 / CIFAR-100 & $0.16$ / $0.82$ & $0.21$ / $0.003$ & 15 / 21 & \\
     & \quad SuperCIFAR-100 / TinyImageNet & $0.38$ / $0.58$ & $0.072$ / $0.020$ & 28 / 15 & \\
    H1 pooling & $|\Delta\rho| = 0.12 > 0.10$ & & & & Not pooled \\
    H2 (dropout) & matched paradigm & $0.40$ & $0.94$ & 27 & Not supported \\
    H3 (coverage) & each dropout setting & \multicolumn{3}{c}{$100\%$ of 20 top-quartile pairs} & Supported \\
    \bottomrule
    \end{tabular}
\end{table}

\paragraph{Results.} Table~\ref{tab:matched_pairs} shows that changing the training paradigm moves NC geometry and the top cliques together in both dropout settings. CIFAR-10 is the weakest source in both dropout settings. The result holds for pairs of different paradigms only, without Deep Gamblers reward pairs ($\rho = 0.47$ and $0.59$, $n = 42$, both supported), for stability thresholds of $0.50$ and $0.70$, and for an $L_1$ NC distance. Per regime it is strongest at mid and far ($\rho = 0.41$--$0.62$) and falls below the $0.40$ floor at near ($0.36$ and $0.34$). Changing only the dropout setting does not move the two together: the H2 correlation is no larger than under shuffles within a source, so it comes from differences between sources. For H4, the angular coordinates (equiangularity and maximal equiangularity of class means and of classifier weights) reach $\rho = 0.51$--$0.58$ in both dropout settings, variability collapse reaches $0.55$ and $0.40$, self-duality reaches the floor only with dropout ($0.28$ and $0.42$), and equinormness never does ($\leq 0.30$). This agrees with the per-metric Mantel results of Table~\ref{tab:mantel_per_metric}. The association of Table~\ref{tab:mantel_global} therefore follows the training paradigm, not the dropout setting. The comparison is still observational: paradigms differ in more than their NC geometry, so the test does not show that the geometry causes the change in cliques.

\section{NC-Based Method Selection: Extended Results}\label{appendix:prediction}
    This appendix compares input configurations and CSF rosters of the predictor and presents the coefficient analysis that complements the regret evaluation of Section~\ref{results:neural_collapse}. The predictor of Section~\ref{methods:prediction} fits one $L_{2}$-regularized logistic regression per CSF on the VGG-13 pool and applies it, without re-training, to a held-out ResNet-18 pool. In the headline configuration, each CSF $m$ has a coefficient vector $\boldsymbol{\beta}_{m} \in \mathbb{R}^{8}$ over the per-architecture $z$-scored NC metrics; the sign and magnitude of each entry are directly interpretable. A positive $\beta_{m,j}$ means that increasing NC metric $j$ raises the predicted probability that CSF $m$ belongs to the top clique.

\paragraph{Input configurations and rosters.}
Table~\ref{tab:predictor_configs} compares the headline NC-only predictor with its ablations, on all 22 CSFs and on the 20 CSFs without Maha++ and NCI. NC + source adds a one-hot encoding of the four source datasets and keeps the predictor regime-free. The regime-input configurations, evaluated on the 20 CSFs only, also receive a one-hot OOD-regime indicator and are fit on one row per (model, regime), so they return a separate shortlist for each regime; this assumes that the regime of the deployment shift is known. In the $n_{\text{cls}}$ variant the source is encoded by its ordinal number of classes. Each roster has its own per-row oracle (the best of 22 CSFs can be better than the best of 20), so the regrets of the two rosters are measured against different references; Always-CTM, the strongest fixed CSF on both rosters, changes only slightly. Including Maha++ and NCI lowers the NC-only regret in all three regimes. On the 20-CSF roster, adding the source lowers regret in all three regimes, and adding the regime helps only at far (NC only $0.80 \rightarrow 0.40$; NC + source $0.64 \rightarrow 0.39$) while it raises the NC-only regret at near and mid ($1.23 \rightarrow 1.45$, $1.16 \rightarrow 1.44$). A regime-free alternative that trains on the top cliques of the pooled near+mid+far blocks (one label per model instead of three) is much worse on the 20-CSF roster (NC only $2.34$/$3.16$/$1.94$; NC + source $2.20$/$2.37$/$1.63$), and we do not use it.

\begin{table}[h]
    \centering
    \small
    \caption{Joint-side mean set-regret (raw AUGRC, empty-set imputed) under VGG-13 $\rightarrow$ ResNet-18 transfer for each input configuration and roster, and mean shortlist size per regime. No configuration produces an empty joint-side shortlist. Regime-free configurations return one shortlist per model.}
    \label{tab:predictor_configs}
    \begin{tabular}{@{}llcccc@{}}
    \toprule
    Configuration & Inputs & near & mid & far & Shortlist size \\
    \midrule
    \multicolumn{6}{@{}l}{\emph{22 CSFs}} \\
    NC only (headline) & 8 NC & $0.93$ & $0.88$ & $0.50$ & $8.1$--$8.4$ \\
    NC + source & 8 NC, source one-hot & $0.78$ & $0.82$ & $0.63$ & $6.4$--$7.0$ \\
    Always-CTM (best fixed CSF) & none & $2.08$ & $3.56$ & $2.57$ & $1$ \\
    \midrule
    \multicolumn{6}{@{}l}{\emph{20 CSFs (without Maha++ and NCI)}} \\
    NC only & 8 NC & $1.23$ & $1.16$ & $0.80$ & $7.3$--$7.5$ \\
    NC + source & 8 NC, source one-hot & $1.06$ & $0.96$ & $0.64$ & $5.6$--$5.8$ \\
    NC + regime & 8 NC, regime one-hot & $1.45$ & $1.44$ & $0.40$ & $4.9$--$7.8$ \\
    NC + $n_{\text{cls}}$ + regime & 8 NC, $n_{\text{cls}}$, regime one-hot & $1.24$ & $1.24$ & $0.36$ & $4.8$--$7.7$ \\
    NC + source + regime & 8 NC, source and regime one-hots & $1.02$ & $1.18$ & $0.39$ & $4.2$--$6.5$ \\
    Always-CTM (best fixed CSF) & none & $2.07$ & $3.52$ & $2.40$ & $1$ \\
    \bottomrule
    \end{tabular}
\end{table}

\paragraph{Labels and the CSF roster.}
The top cliques use all 22 CSFs in the rank pipeline. Without Maha++ and NCI in the pipeline, $98.1\%$ of the (cell, CSF) labels of the other 20 CSFs stay the same, and $144$ of the $168$ (paradigm, source, dropout, reward, regime) cells have identical cliques. NCI is a top-clique member in $40$ of the $168$ OOD cells and Maha++ in $7$. PCA RecError has no positive label, so it has no head and is never shortlisted. In Figure~\ref{fig:top_clique_vgg13}, NCI is in the top clique in 9 of the 48 OOD cells (among them CIFAR-100 far in all three CNN paradigms, next to fDBD) and in 3 misclassification columns, and Maha++ in 2 OOD cells. With them in the pipeline, no other member leaves the OOD cells they enter, and borderline ties shift in six OOD cells (two members leave, five enter); in one misclassification column (DeVries, SuperCIFAR-100) NCI takes the place of Confidence.

\begin{figure}[t]
    \centering
    \includegraphics[width=\textwidth]{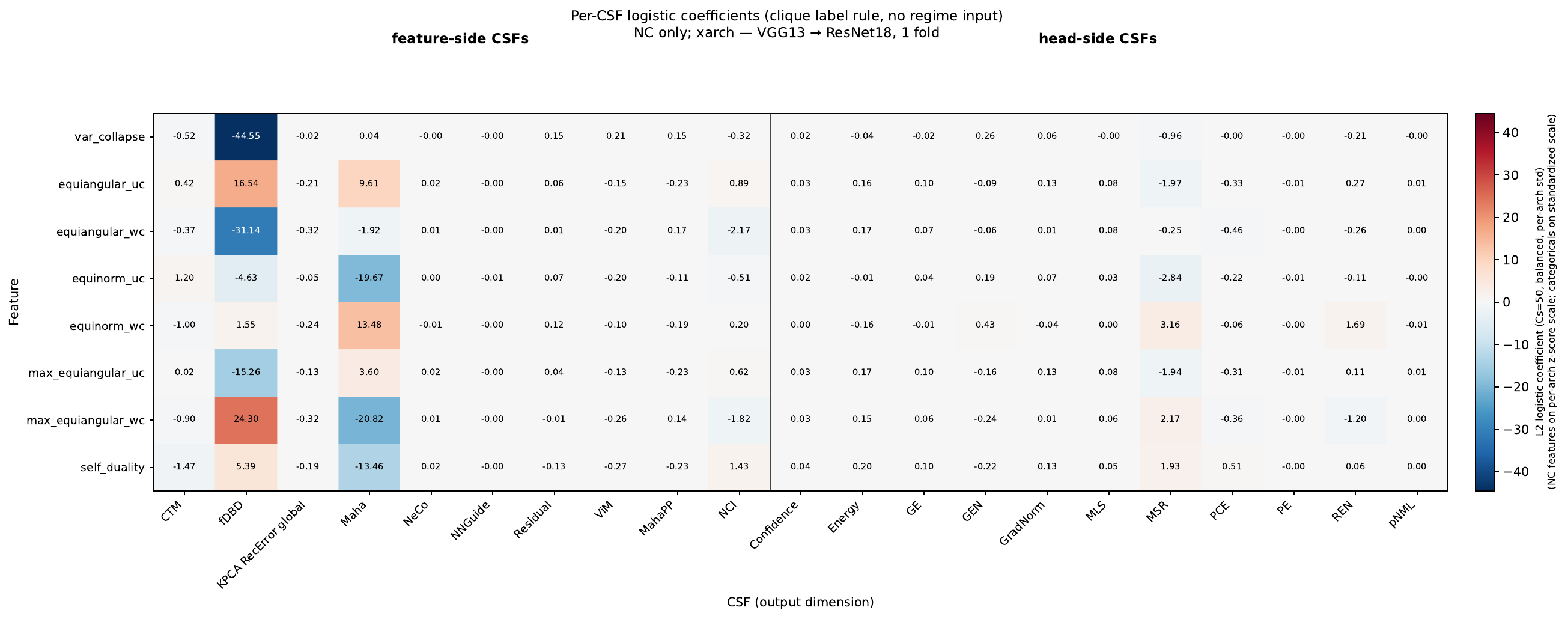}
    \caption{Coefficient heatmap of the headline NC-only predictor on the 22-CSF roster (8 NC metrics, no regime or source input; PCA RecError has no head). Red = positive weight (raises predicted top-clique probability); blue = negative. Columns grouped into feature-side (left) and head-side (right) CSFs. The same fitted models are applied without re-training to the ResNet-18 pool to obtain the regret numbers in Section~\ref{results:neural_collapse}.}
    \label{fig:coef_heatmap_xarch}
\end{figure}

\paragraph{Coefficients of the headline predictor.}
Figure~\ref{fig:coef_heatmap_xarch} shows the per-CSF coefficients of the NC-only predictor. Reading the heatmap with the convention that \emph{lower} raw NC values mean closer to collapse, four patterns dominate:
\begin{enumerate}
    \item \textbf{CTM and fDBD occupy opposite sides of the self-duality axis.} CTM has negative coefficients on within-class variability collapse ($-0.52$) and self-duality ($-1.47$), so CTM is favored under a collapsed and self-dual representation. fDBD has a positive coefficient on self-duality ($+5.39$): fDBD is favored when classifier weights \emph{do not} align with class means, so the decision-boundary distance $\boldsymbol{w}_m - \boldsymbol{w}_k$ carries information that prototype matching no longer does. NCI also has a positive self-duality coefficient ($+1.43$), together with negative coefficients on the equiangularity and maximal equiangularity of $\boldsymbol{W}$ ($-2.17$, $-1.82$): it is favored when the classifier weights are close to a simplex while the class means drift from them.
    \item \textbf{Two heads choose almost no regularization.} Cross-validation selects $C \approx 160$ for fDBD and $C \approx 110$ for Mahalanobis, against $C < 1$ for every other head, and their coefficients are correspondingly large (fDBD: $-44.6$ on within-class variability collapse, $-31.1$ on equiangularity-$\boldsymbol{w}$, $+24.3$ on max-equiangularity-$\boldsymbol{w}$; Mahalanobis: $-20.8$ on max-equiangularity-$\boldsymbol{w}$, $-19.7$ on equinormness-$\boldsymbol{\mu}$, $-13.5$ on self-duality, $+13.5$ on equinormness-$\boldsymbol{w}$). Their predicted probabilities span almost the full range from $0$ to $1$ across ResNet-18 models; we read the signs, not the magnitudes.
    \item \textbf{Maha++ and NCI heads carry real but moderate NC weight.} Their largest coefficients are $0.23$ (Maha++) and $2.17$ (NCI), and they enter $55\%$ and $46\%$ of the ResNet-18 shortlists. Pooled over regimes, their probabilities rank the ResNet-18 rows where each CSF is among the three best (AUROC $0.64$, $0.65$; permutation $p = 0.006$, $0.014$).
    \item \textbf{Several heads carry little NC information.} pNML, PE, and NNGuide select the strongest regularization in the grid ($C = 10^{-4}$); all their NC coefficients are below $0.01$ in magnitude, so their predicted probabilities stay within $0.031$ of $0.5$ for every ResNet-18 model. PE and NNGuide still enter $64\%$ of the shortlists (pNML $27\%$), by a margin of less than $0.007$ above the threshold, so their membership says little about the model's geometry. NeCo, Confidence, GE, MLS, GradNorm, Residual, and Energy have NC coefficients of at most $0.2$ in magnitude. This reflects the \emph{linear} combination of standardized NC features that the $L_{2}$ logistic regression learns; a nonlinear predictor could surface NC-driven structure that no single linear weighting expresses.
\end{enumerate}
Without the regime input, a head cannot move a CSF between regimes; it can only learn how often the CSF is competitive across near, mid, and far for a given geometry.

\begin{figure}[t]
    \centering
    \includegraphics[width=\textwidth]{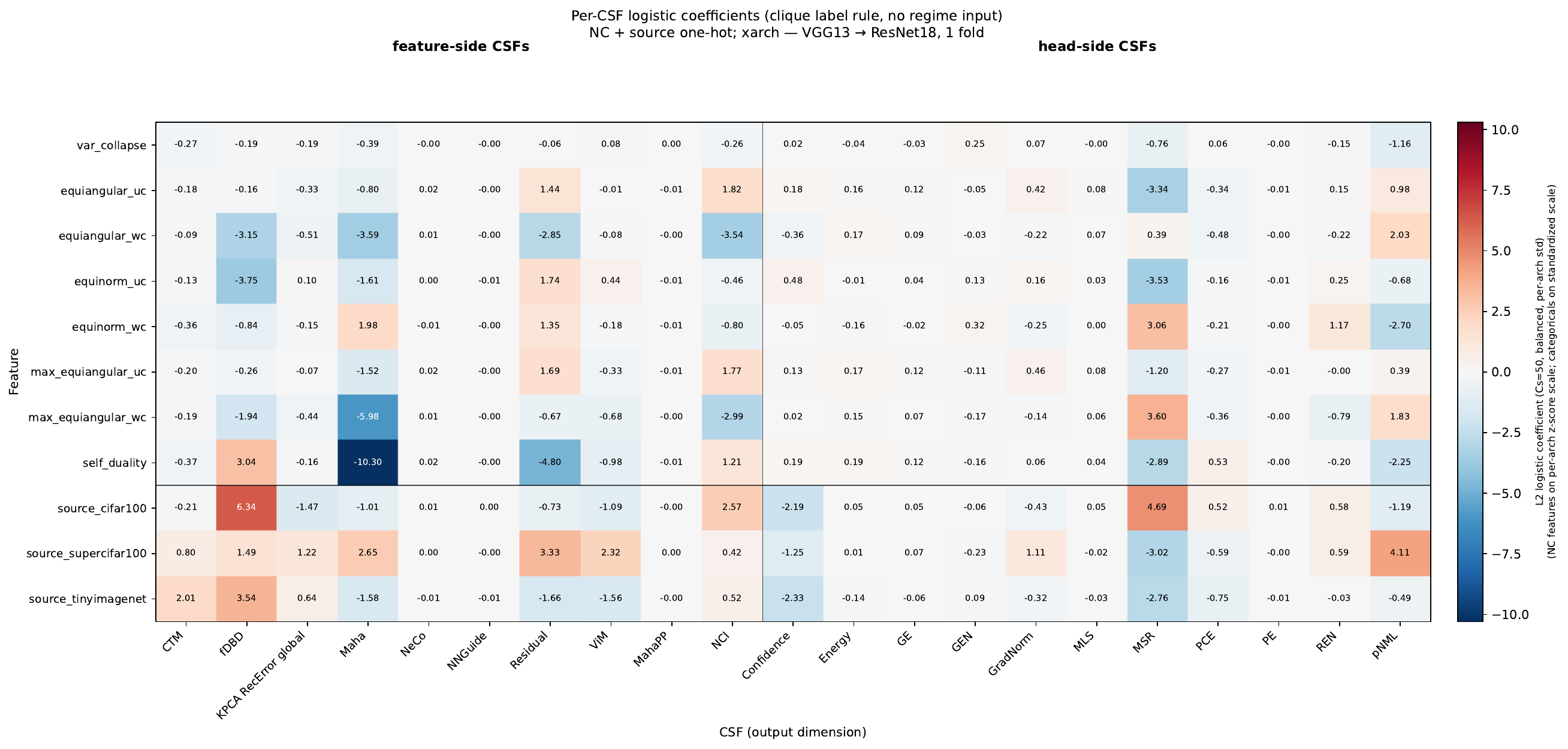}
    \caption{Coefficient heatmap of the NC + source ablation on the 22-CSF roster (8 NC metrics and the source-dataset one-hots, CIFAR-10 as reference; no regime input). Same axes as Figure~\ref{fig:coef_heatmap_xarch}.}
    \label{fig:coef_heatmap_xarch_source}
\end{figure}

\paragraph{Source-descriptor ablation.}
Figure~\ref{fig:coef_heatmap_xarch_source} shows the coefficients of the NC + source predictor. The source one-hots absorb part of the between-source structure: the largest NC coefficients shrink (fDBD from $44.6$ to $3.8$ in magnitude, Mahalanobis from $20.8$ to $10.3$), fDBD receives positive weights for all three non-reference sources (largest for CIFAR-100, $+6.34$), and the Maha++ head becomes nearly constant ($C = 10^{-4}$). The signs that support the reading above are preserved: CTM is negative on self-duality ($-0.37$), and fDBD and NCI are positive ($+3.04$, $+1.21$).

\subsection{Per-source results of the transfers}\label{appendix:per_source}
Table~\ref{tab:per_source} splits the three transfers of the headline predictor by source dataset (joint side). The best fixed CSF of each cell is chosen with hindsight on the same rows: among the Always-$m$ candidates of Section~\ref{methods:prediction} for ResNet-18, as in Section~\ref{results:neural_collapse}, and over the full 21-CSF roster for the fine-tuned ViTs and the self-supervised probes, as in Appendices~\ref{appendix:vit_transfer} and~\ref{appendix:ssl}. On ResNet-18 the predictor beats the best fixed CSF in 9 of the 12 per-source cells and in all pooled cells, and the random set of the same size in all 15 cells; the three losses are on TinyImageNet, where CTM is already close to the oracle (regret $0.32$--$0.56$). On the ViTs it beats the best fixed CSF in 4 of 12 per-source cells, and on the self-supervised probes in 3 of 12 (2 of 3 pooled cells on both); on both targets it beats the random set in only 5 of 12 per-source cells. Where one detector already serves a cell, a fixed choice is enough, and outside the CNN family the shortlists rarely target (Appendices~\ref{appendix:vit_transfer} and~\ref{appendix:ssl}).

\begin{table}[h]
    \centering
    \caption{Per-source transfer of the NC-only predictor (22-CSF heads trained on VGG-13): joint-side mean set-regret (AUGRC $\times 10^{3}$, empty-set imputed), the best fixed CSF of the cell with its regret, and a random shortlist of the same size. Bold: the predictor beats the best fixed CSF; $^{\dagger}$: it beats the random set. C10, C100, SC100, TI: the four source datasets; pooled: all rows of the regime, with the best fixed CSF recomputed on them.}
    \label{tab:per_source}
    \scriptsize
    \setlength{\tabcolsep}{2.5pt}
    \begin{tabular}{@{}llccccccccc@{}}
    \toprule
     & & \multicolumn{3}{c}{ResNet-18} & \multicolumn{3}{c}{Fine-tuned ViT} & \multicolumn{3}{c}{Self-supervised probes} \\
    \cmidrule(lr){3-5}\cmidrule(lr){6-8}\cmidrule(l){9-11}
    Regime & Source & NC only & Best fixed & Random & NC only & Best fixed & Random & NC only & Best fixed & Random \\
    \midrule
    near & C10 & \textbf{0.33}$^{\dagger}$ & 2.58 (NCI) & 0.80 & 0.57 & 0.34 (ViM) & 0.57 & \textbf{1.12} & 1.41 (ViM) & 0.95 \\
     & C100 & \textbf{0.66}$^{\dagger}$ & 1.54 (CTM) & 0.73 & \textbf{0.49}$^{\dagger}$ & 0.73 (ViM) & 0.53 & 2.47 & 1.75 (ViM) & 2.29 \\
     & SC100 & \textbf{1.42}$^{\dagger}$ & 4.19 (CTM) & 1.49 & \textbf{5.70} & 9.35 (ViM) & 2.67 & 16.18 & 5.20 (ViM) & 2.40 \\
     & TI & 1.00$^{\dagger}$ & 0.55 (CTM) & 2.04 & \textbf{0.48}$^{\dagger}$ & 1.51 (GradNorm) & 1.35 & \textbf{1.59}$^{\dagger}$ & 4.25 (NCI) & 1.79 \\
     & pooled & \textbf{0.93}$^{\dagger}$ & 2.08 (CTM) & 1.46 & \textbf{1.44} & 5.85 (GE) & 1.30 & \textbf{4.32} & 6.03 (ViM) & 1.84 \\
    \midrule
    mid & C10 & \textbf{0.24}$^{\dagger}$ & 1.59 (NNGuide) & 0.60 & 0.45 & 0.40 (KPCA-RE) & 0.34 & \textbf{0.21}$^{\dagger}$ & 0.37 (CTM) & 0.40 \\
     & C100 & \textbf{0.68}$^{\dagger}$ & 1.14 (NNGuide) & 1.15 & \textbf{0.63}$^{\dagger}$ & 2.75 (KPCA-RE) & 1.11 & 2.80$^{\dagger}$ & 0.44 (NCI) & 3.06 \\
     & SC100 & \textbf{1.57}$^{\dagger}$ & 6.64 (CTM) & 2.03 & 10.35 & 3.88 (KPCA-RE) & 9.03 & 1.97$^{\dagger}$ & 1.77 (NCI) & 5.15 \\
     & TI & 0.56$^{\dagger}$ & 0.56 (CTM) & 1.52 & 5.39 & 0.71 (ViM) & 1.60 & 1.41 & 0.57 (ViM) & 1.19 \\
     & pooled & \textbf{0.88}$^{\dagger}$ & 3.56 (CTM) & 1.38 & 4.04 & 3.67 (KPCA-RE) & 3.22 & \textbf{1.63}$^{\dagger}$ & 2.07 (NCI) & 2.63 \\
    \midrule
    far & C10 & \textbf{0.24}$^{\dagger}$ & 2.82 (CTM) & 0.98 & 0.12$^{\dagger}$ & 0.02 (Residual) & 0.15 & 1.00 & 0.07 (Maha) & 0.36 \\
     & C100 & \textbf{0.66}$^{\dagger}$ & 1.46 (CTM) & 1.05 & 1.41 & 0.66 (ViM) & 0.83 & 0.70$^{\dagger}$ & 0.18 (Maha++) & 1.85 \\
     & SC100 & \textbf{0.44}$^{\dagger}$ & 4.02 (CTM) & 1.91 & 10.42 & 0.18 (Residual) & 2.17 & 26.25 & 0.06 (Maha++) & 2.08 \\
     & TI & 0.88$^{\dagger}$ & 0.32 (CTM) & 1.77 & 0.12$^{\dagger}$ & 0.10 (GradNorm) & 0.92 & 0.64 & 0.50 (NCI) & 0.38 \\
     & pooled & \textbf{0.50}$^{\dagger}$ & 2.57 (CTM) & 1.43 & \textbf{3.43} & 5.90 (ViM) & 1.03 & 8.08 & 0.85 (Maha++) & 1.28 \\
    \bottomrule
    \end{tabular}
\end{table}

\paragraph{Limitations.}
The 8-Papyan-NC vector is a coarse summary of representation geometry; finer-grained features (e.g., per-class angular variance, layer-wise NC profiles) might reveal sub-structure that the current coefficients average over. The cross-architecture transfer is tested on a single CNN-to-CNN pair (VGG-13 $\rightarrow$ ResNet-18); transfer to fundamentally different architectures (CNN $\rightarrow$ ViT, or to self-supervised backbones) likely requires re-fitting on representative NC distributions of the target family, and Appendix~\ref{appendix:ssl} shows where it fails. Finally, the predictor produces a shortlist, not a single recommendation; users wanting a singleton must apply a tie-break (e.g., minimum train-AUGRC), which forfeits the multi-CSF safety margin documented in Section~\ref{results:neural_collapse}.

\section{Worked example: set-regret computation}\label{appendix:regret_example}
    To make the regret computation of Section~\ref{methods:prediction} concrete, we walk through a single test row from the cross-architecture transfer (VGG-13 $\rightarrow$ ResNet-18). The row is the held-out ResNet-18 model with ID \texttt{ResNet18|confidnet|cifar100|1|0|2.2} (ConfidNet paradigm, CIFAR-100 source, seed $1$, dropout off, Deep Gamblers reward placeholder $2.2$), evaluated on the iSUN OOD set under the mid-OOD regime. This is one of the 200 test rows that contribute to the ``mid / all'' cell of Figure~\ref{fig:regret_xarch}.

\paragraph{Step 1: feature vector $\boldsymbol{\phi}$.} The model's NC profile, $z$-scored within the ResNet-18 pool, gives $\boldsymbol{\phi} \in \mathbb{R}^{8}$. The predictor receives no regime or source input, so the same $\boldsymbol{\phi}$ and the same shortlist serve every near, mid, and far row of this model.

\paragraph{Step 2: predicted shortlist $\hat{\mathcal{C}}(\boldsymbol{\phi})$.} The 21 per-CSF $L_{2}$ logistic models (PCA RecError has no head on the 22-CSF roster, Appendix~\ref{appendix:prediction}) produce the probabilities reported in Table~\ref{tab:regret_example_probs}; CSFs at or above the $0.5$ threshold enter the shortlist.

\begin{center}
    \scriptsize
    \captionsetup{hypcap=false}\captionof{table}{Per-CSF predicted clique-membership probability $p_m(\boldsymbol{\phi})$ for the worked-example model. CSFs in the left column ($p_m \geq 0.5$) form the predicted shortlist $\hat{\mathcal{C}}(\boldsymbol{\phi})$; CSFs in the right column fall below the threshold.}
    \label{tab:regret_example_probs}
    \begin{tabular}{@{}lr@{\hspace{2em}}lr@{}}
    \toprule
    \multicolumn{2}{c}{In $\hat{\mathcal{C}}(\boldsymbol{\phi})$} & \multicolumn{2}{c}{Below threshold} \\
    CSF $m$ & $p_m(\boldsymbol{\phi})$ & CSF $m$ & $p_m(\boldsymbol{\phi})$ \\
    \midrule
    fDBD          & $> 0.999$ & pNML       & 0.497 \\
    MSR           & 0.718     & NeCo       & 0.488 \\
    CTM           & 0.712     & Confidence & 0.466 \\
    KPCA RecError & 0.577     & MLS        & 0.432 \\
    Maha++        & 0.545     & GEN        & 0.427 \\
    PCE           & 0.516     & GE         & 0.412 \\
    ViM           & 0.514     & Residual   & 0.409 \\
    PE            & 0.504     & REN        & 0.383 \\
    NNGuide       & 0.503     & Energy     & 0.380 \\
                  &           & GradNorm   & 0.370 \\
                  &           & NCI        & 0.267 \\
                  &           & Maha       & $< 10^{-3}$ \\
    \bottomrule
    \end{tabular}
\end{center}

The predicted shortlist $\hat{\mathcal{C}}(\boldsymbol{\phi})$ has nine members: fDBD, MSR, CTM, KPCA RecError, Maha++, PCE, ViM, PE, and NNGuide. NNGuide and PE sit just above the threshold; their heads are nearly constant across models (Appendix~\ref{appendix:prediction}).

\paragraph{Step 3: side restriction $\mathcal{M}_{\text{side}(i)}$.} For the joint head$+$feature side, $\mathcal{M}_{\text{side}(i)} = \mathcal{M}$ contains all $22$ CSFs.

\paragraph{Step 4: per-CSF AUGRC on the cell.} The AUGRC of each CSF on this (model, eval-set, regime) row is reported in Table~\ref{tab:regret_example_augrc}.

\begin{center}
    \scriptsize
    \captionsetup{hypcap=false}\captionof{table}{Per-CSF AUGRC on the worked-example test row, sorted ascending. The oracle CSF (CTM, bold) attains the minimum and is in the predicted shortlist.}
    \label{tab:regret_example_augrc}
    \begin{tabular}{@{}lr@{\hspace{2em}}lr@{}}
    \toprule
    CSF $m$ & AUGRC$_{i,m}$ & CSF $m$ & AUGRC$_{i,m}$ \\
    \midrule
    \textbf{CTM} (oracle) & \textbf{175.89} & Confidence           & 179.55 \\
    NNGuide               & 175.92          & PCE                  & 179.60 \\
    fDBD                  & 177.83          & MSR                  & 180.18 \\
    NeCo                  & 178.56          & GEN                  & 182.91 \\
    MLS                   & 178.71          & REN                  & 183.76 \\
    PE                    & 178.81          & pNML                 & 184.69 \\
    Energy                & 179.06          & Maha                 & 185.63 \\
    KPCA RecError         & 179.07          & ViM                  & 188.06 \\
    NCI                   & 179.21          & PCA RecError         & 190.63 \\
    Maha++                & 179.24          & Residual             & 213.94 \\
    GE                    & 179.33          & GradNorm             & 226.11 \\
    \bottomrule
    \end{tabular}
\end{center}

\paragraph{Step 5: oracle and shortlist minima.} We denote the predictor-shortlist minimum AUGRC by $B_{i}$ and compute it alongside the oracle minimum $A^{\star}_{i}$:
\begin{align*}
A^{\star}_{i} &= \min_{m \in \mathcal{M}_{\text{side}(i)}} \mathrm{AUGRC}_{i,m} \;=\; \mathrm{AUGRC}_{i,\text{CTM}} \;=\; 175.89, \\
B_{i}         &= \min_{m \in \hat{\mathcal{C}}(\boldsymbol{\phi}) \,\cap\, \mathcal{M}_{\text{side}(i)}} \mathrm{AUGRC}_{i,m} \\
              &= \min\{175.89,\, 175.92,\, 177.83,\, 178.81,\, 179.07,\, 179.24,\, 179.60,\, 180.18,\, 188.06\} \\
              &= \mathrm{AUGRC}_{i,\text{CTM}} \;=\; 175.89.
\end{align*}

\paragraph{Step 6: set-regret.}
\[
R_{i} \;=\; B_{i} - A^{\star}_{i} \;=\; 175.89 - 175.89 \;=\; 0.
\]
The shortlist contains the oracle CSF, so the regret is zero. For contrast, the configuration on 20 CSFs that also receives the source and the true regime as inputs (Appendix~\ref{appendix:prediction}) places CTM just below the threshold on this row ($p = 0.439$) and returns $\{\text{fDBD},\,\text{NNGuide},\,\text{PE}\}$; its best member, NNGuide, trails the oracle by $R_{i} = 175.92 - 175.89 = 0.028$ AUGRC ($\sim 0.016\%$ of $A^{\star}_{i}$).

\paragraph{Failure modes.} Two cases give nonzero regret. \emph{(a) Suboptimal but non-empty shortlist}, as for the regime-input configuration above. \emph{(b) Empty shortlist}: if no CSF reached the $0.5$ threshold, we set $B_{i}$ to the worst available AUGRC on the side (worst-case empty-set policy). On this row, the imputed value would be $\mathrm{AUGRC}_{i,\text{GradNorm}} = 226.11$, giving $R_{i} = 226.11 - 175.89 = 50.22$. The NC-only predictor returns no empty shortlist on the ResNet-18 pool.

\paragraph{Aggregation.} The mean regret per (regime, side) cell averages $R_{i}$ over all test rows in that cell, with $95\%$ bootstrap CIs from $n_{\text{boot}} = 2000$ resamples (fixed seed). For example, the mid-OOD all-side cell in Figure~\ref{fig:regret_xarch} aggregates $R_{i}$ over the $200$ rows of that cell, giving the headline $0.88$ AUGRC mean regret reported in Section~\ref{results:neural_collapse}. The Holm-corrected paired Wilcoxon tests of Section~\ref{methods:prediction} compare per-row $R_{i}$ for the predictor against each NC-free baseline within the same (regime, side) cell; Appendix~\ref{appendix:mean_regret} tests the mean difference.

\section{Mean-Regret Test and Random-Set Control}\label{appendix:mean_regret}
    Section~\ref{results:neural_collapse} tests mean set-regret with the paired Wilcoxon signed-rank test, which ranks the per-row differences: it tests whether the predictor wins on the typical row, not whether its mean regret is lower. The two can disagree because regret differences are heavy-tailed. For example, on the feature-only near cell the NC + source predictor has lower regret than Always-NCI on $75\%$ of the $148$ rows (median difference $-2.07$ AUGRC), but a few large losses (up to $63.8$) make its mean regret higher ($5.35$ vs $4.09$), while the Wilcoxon test calls it significantly better ($p_{\text{Holm}} < 10^{-4}$). This appendix adds a test of the mean difference and a control for shortlist size.

\paragraph{Mean-difference test.} For each comparison we sum the per-row differences $d_{i} = R^{\text{pred}}_{i} - R^{\text{baseline}}_{i}$ over the rows of each ResNet-18 model $k$, giving $S_{k}$, and test the total $T = \sum_{k} S_{k}$ with a sign-flip permutation test clustered by model: under the null hypothesis the signs of the $S_{k}$ are exchangeable, so $T$ is compared with $T^{*} = \sum_{k} \varepsilon_{k} S_{k}$ for random signs $\varepsilon_{k} \in \{-1, +1\}$ ($B = 100{,}000$ draws, fixed seed, one-sided $p$-values in both directions with the $+1$ correction, Holm within each cell). All rows of one model flip together because they are not independent.

For the headline predictor the two tests agree in every comparison: it is significantly better than the baseline in 65 of the 66 comparisons and never significantly worse, and the exception is feature-only near against Always-CTM ($1.76$ vs $1.58$). For NC + source the mean test finds 60 wins instead of 65, and no losses; four of the five disagreements are in the feature-only near cell of the example above.

\paragraph{Size-matched random-set control.}
The NC-only shortlists hold $8.1$--$8.4$ of the $22$ CSFs on the joint side, and a larger set is more likely to contain a near-oracle CSF by chance. The Random-CSF baseline draws a single CSF, so it does not control for this. We therefore compare each shortlist with a uniformly random subset of the same size $k$, drawn from the CSFs of the same side. Its expected regret is exact: with the side's AUGRCs on row $i$ sorted as $a_{(1)} \leq \dots \leq a_{(M)}$,
\[
\mathbb{E}\Bigl[\min_{m \in \text{random } k\text{-subset}} \mathrm{AUGRC}_{i,m}\Bigr] = \sum_{j=1}^{M} a_{(j)} \frac{\binom{M-j}{k-1}}{\binom{M}{k}},
\]
and the control's regret is this value minus $A^{\star}_{i} = a_{(1)}$. Rows with an empty shortlist keep the imputed regret, so the control differs from the predictor only in which CSFs it picks. We apply both one-sided tests (predictor better) and Holm-correct over the nine (regime, side) cells.

\begin{table}[h]
    \centering
    \small
    \caption{NC-only predictor (22 CSFs) against a random shortlist of the same size: mean set-regret, mean shortlist size $k$, and Holm-corrected $p$-values over the nine cells (predictor better).}
    \label{tab:random_set_control}
    \begin{tabular}{@{}llccccc@{}}
    \toprule
    Side & Regime & NC only & Random set & $k$ & Wilcoxon $p_{\text{Holm}}$ & Mean test $p_{\text{Holm}}$ \\
    \midrule
    joint & near & $0.93$ & $1.46$ & $8.1$ & $< 10^{-4}$ & $0.008$ \\
    joint & mid & $0.88$ & $1.38$ & $8.4$ & $< 10^{-4}$ & $< 10^{-3}$ \\
    joint & far & $0.50$ & $1.43$ & $8.4$ & $< 10^{-4}$ & $< 10^{-3}$ \\
    head & near & $2.41$ & $2.43$ & $4.0$ & $1.00$ & $0.96$ \\
    head & mid & $3.64$ & $3.65$ & $3.6$ & $1.00$ & $0.96$ \\
    head & far & $2.75$ & $2.95$ & $3.6$ & $0.37$ & $0.63$ \\
    feature & near & $1.76$ & $8.59$ & $4.1$ & $< 10^{-4}$ & $< 10^{-3}$ \\
    feature & mid & $1.10$ & $6.23$ & $4.8$ & $< 10^{-4}$ & $< 10^{-3}$ \\
    feature & far & $0.87$ & $5.32$ & $4.8$ & $< 10^{-4}$ & $< 10^{-3}$ \\
    \bottomrule
    \end{tabular}
\end{table}

Table~\ref{tab:random_set_control} shows that the NC-only shortlists are better targeted than random sets of the same size on the joint side and on the feature side, in every regime and under both tests. On the head side they are not: a random set of about four of the $11$ head-side CSFs already has lower mean regret ($2.43$--$3.65$) than the best fixed head-side CSF ($3.49$--$4.20$), so set size alone explains the head-side gain. The NC + source configuration passes this control on the head side as well ($p_{\text{Holm}} \leq 10^{-3}$ in all three regimes), so on the head side the useful targeting comes from the source descriptor, which the NC-only predictor does not have.

\section{Fine-Tuned ViT Transfer}\label{appendix:vit_transfer}
    This appendix applies the headline predictor of Section~\ref{results:neural_collapse}, trained on the VGG-13 pool, to the 40 fine-tuned ViT models of the main study without re-training. The ViT models have the same sources, OOD sets, and CLIP regimes as the CNNs. The NC metrics are $z$-scored within the ViT pool, and the roster has the 21 CSFs of the ViT panel of Figure~\ref{fig:top_clique_vgg13} (no Confidence readout). We score each shortlist by its mean set-regret against the per-row oracle and compare it with three references: the best fixed CSF of each cell, chosen with hindsight on the same rows; the best CNN fixed baseline of Section~\ref{methods:prediction} (Always-$m$ for CTM, Energy, MLS, MSR, NNGuide, fDBD, Maha++, and NCI); and a random shortlist of the same size (Appendix~\ref{appendix:mean_regret}). The tests are those of Appendix~\ref{appendix:imagenet}, with Holm correction over the four cells of Table~\ref{tab:vit_transfer}.

\begin{table}[h]
    \centering
    \caption{CNN $\to$ ViT transfer: mean set-regret (AUGRC $\times 10^{3}$, empty-set imputed) of the VGG-13-trained NC-only predictor over the 40 fine-tuned ViTs. The shortlists hold $8.0$--$8.5$ of the $21$ CSFs. ``Oracle in set'' is the share of rows whose shortlist contains the oracle CSF, with the expected share for a random set of the same size in parentheses. Holm-corrected one-sided $p$-values for the predictor against the best fixed CSF and against the random set (Wilcoxon / mean test).}
    \label{tab:vit_transfer}
    \scriptsize
    \setlength{\tabcolsep}{3pt}
    \begin{tabular}{@{}lccccccc@{}}
    \toprule
    Cell & NC only & Best fixed CSF & Best CNN fixed & Random set & Oracle in set & $p$, best fixed & $p$, random \\
    \midrule
    pooled & $3.01$ & $7.51$ (KPCA RecError) & $8.92$ (NCI) & $2.08$ & $33\%$ ($40\%$) & $<10^{-4}$ / $<10^{-4}$ & $0.79$ / $1.00$ \\
    near & $1.44$ & $5.85$ (GE) & $6.26$ (MLS) & $1.30$ & $45\%$ ($38\%$) & $<10^{-4}$ / $<10^{-4}$ & $0.007$ / $1.00$ \\
    mid & $4.04$ & $3.67$ (KPCA RecError) & $8.87$ (NCI) & $3.22$ & $28\%$ ($41\%$) & $0.50$ / $0.64$ & $0.98$ / $1.00$ \\
    far & $3.43$ & $5.90$ (ViM) & $10.62$ (fDBD) & $1.03$ & $26\%$ ($41\%$) & $0.074$ / $0.25$ & $0.60$ / $1.00$ \\
    \bottomrule
    \end{tabular}
\end{table}

\paragraph{Results.} Against the CNN fixed baselines, which do poorly on the ViTs, the predictor cuts the mean regret by $77\%$, $55\%$, and $68\%$ at near, mid, and far OOD. Against the stronger references, the transfer is weak. The predictor beats the best fixed ViT CSF only at near OOD; at mid it is slightly worse than KPCA RecError, and at far its gain is not significant. A random shortlist of the same size has a lower mean regret in every cell, and the shortlists contain the oracle less often than a random set of the same size would, except at near. The gain over the CNN baselines therefore comes from recommending a set of about eight CSFs, not from the geometry.

\paragraph{Why the geometry does not target.} On the ViTs, reconstruction-, residual-, and gradient-based CSFs and pNML are the oracle on most rows (KPCA RecError $22\%$, Residual $18\%$, GradNorm $13\%$, ViM $11\%$, pNML $8\%$, PCA RecError $5\%$). The VGG-13 cliques rarely reward these CSFs, so the heads rarely shortlist them (KPCA RecError and ViM in $28\%$ of the shortlists, Residual in $18\%$, PCA RecError never, because it has no head), and they favor the CNN winners (fDBD in $72\%$, NCI in $60\%$). This is not an extrapolation in NC space: the NC metrics are $z$-scored within each architecture, so the ViT inputs lie on the same standardized scale as the training inputs. The heads map the position of a model within its family to the CSFs that are competitive in the CNN family. Per source (exploratory), the predictor does well where the CNN winners stay competitive (CIFAR-100 mid: $0.63$ against $2.75$ for the best fixed CSF and $1.11$ for the random set; TinyImageNet near: $0.48$ against $1.51$ and $1.35$) and badly where they do not (SuperCIFAR-100 mid and far: $10.35$ and $10.42$ against $3.88$ and $0.19$ for the best fixed CSF).

\paragraph{Heads trained on ViT labels.} If the cause is missing ViT labels, heads trained on ViTs should fix it. We leave one ViT source out: with ViT top-clique labels from the per-configuration pipeline of Section~\ref{methods:prediction}, we train heads on the ViTs of three sources, or on these ViTs and the VGG-13 pool together, and test them on the 10 models of the fourth source. The decision rule, fixed before the run, requires the ViT heads to beat the VGG-13 heads and a random shortlist of the same size on the pooled mean. The planned base-rate control (the CSFs with a ViT label rate of at least $0.5$) was empty in about half of the folds; after the run we replaced it with the size-matched set of the CSFs most often competitive on the training ViTs.

\begin{table}[h]
    \centering
    \caption{Leave-one-ViT-source-out test: mean set-regret (AUGRC $\times 10^{3}$, empty-set imputed) on the held-out source, for heads trained on VGG-13 (the predictor of Table~\ref{tab:vit_transfer}), on the ViTs of the other three sources, and on both, against a base-rate shortlist (the CSFs most often competitive on the training ViTs; post hoc) and a random shortlist, both with the size of the ViT-head shortlist. Set size: mean size of the ViT-head shortlist.}
    \label{tab:vit_loso}
    \scriptsize
    \begin{tabular}{@{}lcccccc@{}}
    \toprule
    Held-out source & VGG-13 heads & ViT heads & VGG-13 + ViT heads & Base-rate set & Random set & Set size \\
    \midrule
    CIFAR-10 & $0.40$ & $0.13$ & $0.57$ & $0.10$ & $0.55$ & $6.0$ \\
    CIFAR-100 & $0.79$ & $0.74$ & $1.30$ & $0.52$ & $1.82$ & $6.1$ \\
    SuperCIFAR-100 & $9.20$ & $44.77$ & $19.19$ & $38.85$ & $40.26$ & $0.7$ \\
    TinyImageNet & $1.66$ & $2.47$ & $1.64$ & $3.35$ & $2.08$ & $4.5$ \\
    \midrule
    pooled & $3.01$ & $12.03$ & $5.67$ & $10.71$ & $11.18$ & $4.3$ \\
    \bottomrule
    \end{tabular}
\end{table}

Table~\ref{tab:vit_loso} shows that the test fails: neither the ViT heads nor the joint heads beat the VGG-13 heads or their own random sets on the pooled mean (one-sided mean test, $p \geq 0.97$). Most of the gap comes from the SuperCIFAR-100 fold, where the ViT heads shortlist $0.7$ CSFs on average and the regret reaches $44.8$. On CIFAR-10 and CIFAR-100 (exploratory), the ViT heads beat the VGG-13 heads and their random sets, but the base-rate set does as well or better: ViT labels help where the held-out source resembles the training sources, and the NC metrics add no targeting within the ViT family. With six (source, dropout) training configurations per fold, the test has little power for such a signal.

\section{Self-Supervised Probe Pool}\label{appendix:ssl}
    This appendix tests the predictor on representations from a training paradigm it has never seen: linear probes on frozen self-supervised encoders.

\paragraph{Pool and protocol.}
We freeze two pretrained image encoders, DINOv2 ViT-B/14~\cite{oquab2024dinov2} and CLIP ViT-B/16~\cite{radford2021learning}, and train linear probes on each of the four source datasets with five seeds, which gives $40$ probe models. DINOv2 is trained with self-supervision and CLIP with image--caption pairs; neither uses class labels, so we call both self-supervised for short. They are evaluated with the protocol of the main study on cached features: the same OOD datasets and CLIP regimes, $21$ CSFs (the 22-CSF roster without the training-based Confidence readout, which a linear probe does not have), and the CSF tuning of Appendix~\ref{appendix:cfs_variations}, except that the NCI parameter $\alpha$ is searched over $[0, 3\times 10^{-2}]$. The NC metrics are computed as in Appendix~\ref{appendix:neural_collapse} and $z$-scored within the probes of each encoder.

\paragraph{Geometry.}
The probes occupy a weak-collapse regime far from the CNN pools. The mean self-duality per (encoder, source) is $0.61$--$1.29$, against $0.03$--$0.13$ for VGG-13; only the fine-tuned ViTs come close ($0.85$--$1.27$). In addition, $32$--$81\%$ of the within-class variance is concentrated in a few dominant directions above the bulk of the within-class eigenvalue spectrum. CLIP probes are further from collapse than DINOv2 probes on both measures.

\paragraph{Cliques and NC association.}
The competitive families follow the geometry. For CLIP probes, the far cliques contain Residual (two sources), KPCA RecError, NCI, and fDBD; NCI (three sources) and ViM (two) lead at mid, and fDBD is in the near clique for all four sources. For DINOv2 probes, Maha++ is a far top-clique member for three of the four sources, and NeCo and ViM lead at near. Across the $40$ probes, the Mantel correlation between NC distance and CSF-ranking distance (Appendix~\ref{appendix:mantel}) is $r = 0.72$ ($p = 10^{-4}$), above the values of the CNN strata.

\paragraph{Transfer of the predictor.}
We apply the regime-free 22-CSF heads, trained on the VGG-13 pool only, to the $40$ probes without re-training. Table~\ref{tab:ssl_transfer} compares them with the best fixed CSF of the full 21-CSF roster in each cell, chosen with hindsight on the probe pool; the pooled best fixed CSF is one CSF for all sources and both encoders, so it can be worse than every per-source best. NC only beats the best fixed CSF at pooled near ($4.32$ vs $6.03$) and mid ($1.63$ vs $2.07$), but fails at far ($8.08$ vs $0.85$ for Maha++), and it wins only $3$ of the $12$ per-source cells. NC + source shows the same pattern ($5.93$/$1.24$/$9.87$).

\begin{table}[h]
    \centering
    \small
    \caption{Self-supervised probes: mean set-regret (raw AUGRC, empty-set imputed, joint side) of the VGG-13-trained 22-CSF heads against the best fixed CSF of each cell (name in parentheses). C10/C100/SC100/TI are the four sources; \emph{all} pools every (probe, OOD dataset) row of the regime. Bold: NC only beats the best fixed CSF.}
    \label{tab:ssl_transfer}
    \begin{tabular}{@{}llccc@{}}
    \toprule
    Regime & Source & Best fixed CSF & NC + source & NC only \\
    \midrule
    near & C10 & $1.41$ (ViM) & $10.55$ & $\mathbf{1.12}$ \\
    near & C100 & $1.75$ (ViM) & $4.72$ & $2.47$ \\
    near & SC100 & $5.20$ (ViM) & $7.58$ & $16.18$ \\
    near & TI & $4.25$ (NCI) & $3.90$ & $\mathbf{1.59}$ \\
    near & all & $6.03$ (ViM) & $5.93$ & $\mathbf{4.32}$ \\
    \midrule
    mid & C10 & $0.37$ (CTM) & $1.02$ & $\mathbf{0.21}$ \\
    mid & C100 & $0.44$ (NCI) & $0.37$ & $2.80$ \\
    mid & SC100 & $1.77$ (NCI) & $1.97$ & $1.97$ \\
    mid & TI & $0.57$ (ViM) & $1.95$ & $1.41$ \\
    mid & all & $2.07$ (NCI) & $1.24$ & $\mathbf{1.63}$ \\
    \midrule
    far & C10 & $0.07$ (Maha) & $20.81$ & $1.00$ \\
    far & C100 & $0.18$ (Maha++) & $3.24$ & $0.70$ \\
    far & SC100 & $0.06$ (Maha++) & $10.12$ & $26.25$ \\
    far & TI & $0.50$ (NCI) & $0.78$ & $0.64$ \\
    far & all & $0.85$ (Maha++) & $9.87$ & $8.08$ \\
    \bottomrule
    \end{tabular}
\end{table}

\paragraph{Why the predictor fails at far.}
The predictor can only recommend what the VGG-13 cliques reward. NCI joins many VGG-13 cliques (it is positive in $24\%$ of the training rows) and PCA RecError has no positive label (Appendix~\ref{appendix:prediction}). As a result, the probe shortlists rarely contain the CSFs that win at far on weak-collapse probes: Maha++ is shortlisted for $32\%$ of the probes, KPCA RecError for $28\%$, ViM for $22\%$, and Residual for $20\%$, against $55\%$ (KPCA RecError) and $45\%$ (ViM) with heads trained on cliques without Maha++ and NCI. The roster is not the only cause: with those heads the pooled far regret is $1.66$, which also loses to Maha++. Far shifts on weak-collapse representations are therefore a boundary of the transfer: when the CSFs that a new family favors are rarely competitive in the training pool, the shortlist can miss them.

\section{ImageNet-Scale Transfer}\label{appendix:imagenet}
    This appendix applies the headline predictor, trained on the CIFAR-scale VGG-13 pool, to ImageNet-1k models without re-training.

\paragraph{Pool and protocol.}
We use $47$ pretrained ImageNet-1k~\cite{deng2009imagenet} classifiers from the \texttt{timm} library~\cite{wightman2019pytorch} in five families: $16$ ResNet-50 training recipes (including semi- and weakly supervised pretraining), $8$ ViT-B/16 and DeiT models, $6$ Swin and $6$ ConvNeXt models (each with ImageNet-1k and ImageNet-22k pretraining), and $11$ other architectures (DenseNet, EfficientNet, MaxViT, MobileNetV3, RegNetY, ResNeXt, VGG, and Wide-ResNet). Top-1 accuracy ranges from $69.9\%$ to $85.8\%$. All models are evaluated at $224 \times 224$. CSFs are fitted on $100$ training images per class, their hyperparameters are tuned on $25$ further images per class as in Appendix~\ref{appendix:cfs_variations} (the NCI parameter $\alpha$ over $[0, 3\times 10^{-2}]$), and the ImageNet validation set ($50{,}000$ images) is the ID test set. The OOD suite has eight sets: the five of OpenOOD v1.5~\cite{zhang2023openood} (SSB-Hard~\cite{vaze2022open}, NINCO~\cite{bitterwolf2023ninco}, iNaturalist~\cite{vanhorn2018inaturalist}, Textures~\cite{cimpoi2014describing}, OpenImage-O~\cite{wang2022vim}) and ImageNet-O~\cite{hendrycks2021natural}, Places~\cite{zhou2017places}, and SUN~\cite{xiao2010sun}. The roster has $21$ CSFs (the 22 of the main study without the training-based Confidence readout). For the torchvision ResNet-50, accuracy ($76.15\%$ vs $76.18\%$) and the MSR and Energy AUROCs reproduce OpenOOD v1.5 within $1.3$ points (Energy within $0.2$). Rerunning three models with a different fit sample changes the AUGRC of 20 CSFs by a median of at most $0.07$; only KPCA RecError is unstable across fit samples, so we also report results without it.

\paragraph{Regimes.}
We apply the CLIP protocol of Section~\ref{methods:clip} with the ViT-B/32 encoder. The selection rule of Appendix~\ref{appendix:clip} (highest silhouette, with no single-dataset cluster) picks $k = 2$ on ImageNet, whereas it picks $k = 3$ on the CIFAR-scale suites: \emph{near} = \{NINCO, SSB-Hard\} and \emph{far} = the other six sets (silhouette $0.50$ at $k = 2$, $0.48$ at $k = 3$, where iNaturalist forms a cluster alone). ViT-B/16 gives the same grouping. With ViT-L/14, iNaturalist stays alone at $k = 2$ and $k = 3$, because its class-aware distances disagree with the smaller encoders (Spearman $0.14$ for the image-centroid distance), while the distribution distances agree (Spearman $\geq 0.95$). Our near set equals OpenOOD's near set; OpenOOD's far set (iNaturalist, Textures, OpenImage-O) is a subset of ours, and we report it separately.

\paragraph{Predictor and baselines.}
The regime-free NC-only heads of Section~\ref{results:neural_collapse} (trained on the VGG-13 pool with the 22-CSF cliques) predict a shortlist for each model; the NC metrics are $z$-scored within each family, the analogue of the per-architecture standardization. The source-descriptor ablation cannot be evaluated here, because ImageNet is not one of the training sources. We compare the mean set-regret over the $376$ (model, OOD set) rows with three references: the best fixed CSF of each cell, chosen with hindsight on the same rows (no prior ImageNet clique exists to choose it from); Always-CTM, the strongest fixed CSF on the CIFAR benchmark; and a random shortlist of the same size (Appendix~\ref{appendix:mean_regret}). Tests are one-sided (predictor better): the Wilcoxon test and the model-clustered sign-flip test on the mean difference, Holm-corrected over the cells of Table~\ref{tab:imagenet_main} (the shared near cell counted under both groupings).

\begin{table}[h]
    \centering
    \small
    \caption{ImageNet transfer: mean set-regret (AUGRC $\times 10^{3}$, empty-set imputed) of the VGG-13-trained NC-only predictor over $47$ models. The shortlists hold $7.8$ of the $21$ CSFs on average. Holm-corrected $p$-values for the predictor against the best fixed CSF and against the random set of the same size (Wilcoxon / mean test). The CLIP near cell equals OpenOOD's near cell.}
    \label{tab:imagenet_main}
    \footnotesize
    \setlength{\tabcolsep}{4pt}
    \begin{tabular}{@{}lcccccc@{}}
    \toprule
    Cell & NC only & Best fixed CSF & Always-CTM & Random set & $p$, best fixed & $p$, random \\
    \midrule
    pooled (8 sets) & $1.05$ & $4.64$ (NNGuide) & $7.82$ & $1.25$ & $<10^{-4}$ / $<10^{-4}$ & $10^{-4}$ / $0.58$ \\
    near (2 sets) & $1.00$ & $2.80$ (PE) & $14.93$ & $1.26$ & $<10^{-4}$ / $<10^{-4}$ & $0.002$ / $0.58$ \\
    CLIP far (6 sets) & $1.07$ & $3.34$ (fDBD) & $5.45$ & $1.25$ & $<10^{-4}$ / $<10^{-4}$ & $0.002$ / $0.58$ \\
    OpenOOD far (3 sets) & $0.92$ & $2.99$ (fDBD) & $5.18$ & $1.31$ & $<10^{-4}$ / $<10^{-4}$ & $0.002$ / $0.21$ \\
    \bottomrule
    \end{tabular}
\end{table}

\paragraph{Results.}
Table~\ref{tab:imagenet_main} shows that the transferred shortlists beat every fixed CSF by a wide margin in every cell, under both tests: pooled, their regret is $77\%$ below that of the best fixed CSF chosen with hindsight and $87\%$ below Always-CTM, which does poorly on ImageNet. But a random shortlist of the same size is almost as good ($1.25$ vs $1.05$ pooled). The predictor is better than the random set under the Wilcoxon test, not under the mean test. It does target: its shortlists contain the oracle CSF on $57\%$ of the rows, against $37\%$ expected for a random set of the same size. But a typical ImageNet row has two CSFs within $1$ AUGRC of the oracle, so a random set of about eight CSFs rarely misses by much. At ImageNet scale, most of the predictor's advantage over a fixed CSF therefore comes from recommending a set rather than from the geometry. Without KPCA RecError (the oracle on only $0.3\%$ of the rows), every regret in Table~\ref{tab:imagenet_main} changes by at most $0.02$, and no test changes its outcome.

\paragraph{Per family.}
Table~\ref{tab:imagenet_family} splits the result by family (exploratory; raw $p$-values). The geometry targets best for the ResNet-50 recipes, the family that varies only the training recipe on a fixed architecture, as the VGG-13 training pool does: at near, the predictor reaches $0.30$ against $1.03$ for the random set ($p = 0.004$). It does not target for ConvNeXt, where it loses to Always-Maha++ and to the random set. Maha++ and NCI dominate at this scale: Maha++ is the oracle on $34\%$ of the rows and NCI on $17\%$ (base Mahalanobis on $6\%$), and Maha++ is the best fixed CSF for ConvNeXt, Swin, and ViT-B/16. Maha++ has a lower regret than base Mahalanobis on $83\%$ of the rows (mean $7.2$ vs $16.0$). In NC terms, the ImageNet families lie between the CIFAR CNNs and the fine-tuned ViTs: their median self-duality is $0.51$--$0.89$, against $0.085$ for VGG-13 (maximum $0.68$) and about $1.05$ for the fine-tuned ViTs and the self-supervised probes. The per-family standardization removes these level differences, so the heads see only the variation within each family.

\begin{table}[h]
    \centering
    \small
    \caption{ImageNet transfer per family (exploratory): mean set-regret of the NC-only predictor, the best fixed CSF of the cell (chosen with hindsight), and the random set of the same size, with raw one-sided mean-test $p$-values (predictor better).}
    \label{tab:imagenet_family}
    \begin{tabular}{@{}llccccc@{}}
    \toprule
    Family & Cell & NC only & Best fixed CSF & Random set & $p$ vs best fixed & $p$ vs random \\
    \midrule
    ResNet-50 (16) & pooled & $0.83$ & $4.93$ (fDBD) & $1.42$ & $<10^{-4}$ & $0.054$ \\
     & near & $0.30$ & $1.07$ (REN) & $1.03$ & $0.006$ & $0.004$ \\
     & far & $1.01$ & $2.31$ (NCI) & $1.55$ & $0.019$ & $0.10$ \\
    ViT-B/16 (8) & pooled & $0.60$ & $1.12$ (Maha++) & $1.03$ & $0.086$ & $0.074$ \\
     & near & $1.00$ & $1.99$ (Maha++) & $1.43$ & $0.093$ & $0.17$ \\
     & far & $0.47$ & $0.83$ (Maha++) & $0.90$ & $0.15$ & $0.066$ \\
    Swin (6) & pooled & $1.07$ & $2.16$ (Maha++) & $1.26$ & $0.031$ & $0.36$ \\
     & near & $0.98$ & $5.15$ (NCI) & $1.87$ & $0.016$ & $0.077$ \\
     & far & $1.10$ & $1.07$ (Maha++) & $1.06$ & $0.59$ & $0.54$ \\
    ConvNeXt (6) & pooled & $1.44$ & $0.91$ (Maha++) & $0.89$ & $0.79$ & $0.79$ \\
     & near & $1.63$ & $1.08$ (Maha++) & $1.44$ & $0.87$ & $0.60$ \\
     & far & $1.38$ & $0.85$ (Maha++) & $0.71$ & $0.79$ & $0.83$ \\
    Other (11) & pooled & $1.48$ & $4.34$ (REN) & $1.37$ & $<10^{-3}$ & $0.60$ \\
     & near & $1.69$ & $1.02$ (REN) & $1.05$ & $0.74$ & $0.75$ \\
     & far & $1.41$ & $3.49$ (NCI) & $1.47$ & $0.072$ & $0.42$ \\
    \bottomrule
    \end{tabular}
\end{table}

\newpage
\section*{NeurIPS Paper Checklist}

\begin{enumerate}

\item {\bf Claims}
    \item[] Question: Do the main claims made in the abstract and introduction accurately reflect the paper's contributions and scope?
    \item[] Answer: \answerYes{}
    \item[] Justification: The introduction's four contributions (NC-based detector selection under cross-architecture transfer, a factorial benchmark with CLIP near/mid/far stratification, a rank-based clique pipeline, and PCA projection filtering) each map to a results section: §\ref{results:top_cliques}, §\ref{results:projection_gains}, and §\ref{results:neural_collapse}. The headline numbers in the introduction and conclusion (Mantel $r = 0.37$--$0.70$ with $p \leq 0.002$, $55$--$80\%$ set-regret reduction across near/mid/far, and 65 of 66 Holm-corrected Wilcoxon comparisons significant) come directly from §\ref{results:neural_collapse} and Appendix~\ref{appendix:mantel}. The introduction and conclusion limit the predictor's claim to VGG-13 $\rightarrow$ ResNet-18 transfer; §\ref{results:limitations} and Appendices~\ref{appendix:vit_transfer}--\ref{appendix:imagenet} report the weaker results on other architecture families.

\item {\bf Limitations}
    \item[] Question: Does the paper discuss the limitations of the work performed by the authors?
    \item[] Answer: \answerYes{}
    \item[] Justification: Section~\ref{results:limitations} is a dedicated Limitations section with five paragraphs covering (i) which claims are established and which are exploratory, (ii) scope (datasets, backbones, paradigms, uncovered training methods, untested hypotheses on class imbalance and multi-label training), (iii) shift stratification (dependence on the CLIP encoder and the clustering, fine-grained and industrial domains), (iv) cross-architecture transfer (the predictor beats fixed CSFs on other families, but a same-size random shortlist reaches about the same regret on ImageNet models and a lower one on fine-tuned ViTs, and it fails at far OOD on self-supervised probes; one ResNet-18 run per cell), and (v) statistical methodology (correlational Mantel association, rank-pipeline assumptions, pairwise tests not adjusted for multiplicity, dependence on the NC1 normalization, AUGRC-specific hyperparameter tuning, metric sensitivity of Deep Gamblers, no compute/latency evaluation). Appendix~\ref{appendix:prediction} adds further limitations on the 8-Papyan-NC vector's coarseness and the shortlist-versus-singleton tradeoff.

\item {\bf Theory assumptions and proofs}
    \item[] Question: For each theoretical result, does the paper provide the full set of assumptions and a complete (and correct) proof?
    \item[] Answer: \answerNA{}
    \item[] Justification: The paper is an empirical systematic analysis of OOD detection; it contains no formal theorems, lemmas, or proofs. The mechanistic discussions in Section~\ref{methods:neural_collapse} and Appendix~\ref{appendix:projection_full} are interpretive and reference existing definitions (e.g., Papyan Neural Collapse metrics) without introducing new theoretical claims requiring proof.

    \item {\bf Experimental result reproducibility}
    \item[] Question: Does the paper fully disclose all the information needed to reproduce the main experimental results of the paper to the extent that it affects the main claims and/or conclusions of the paper (regardless of whether the code and data are provided or not)?
    \item[] Answer: \answerYes{}
    \item[] Justification: Section~\ref{methods:def_notations}--\ref{methods:prediction} describes the full experimental factorization (backbone, training paradigm, source dataset, dropout, Deep~Gamblers reward, OOD regime), and Section~\ref{results:experimental_setup} specifies the 22 evaluated CSFs grouped by side, the AURC/AUGRC ranking metrics, and the use of 5 random seeds per (backbone, paradigm) cell. Appendix~\ref{appendix:hyperparameter_selection} reports the per-cell training-time dropout setting and Deep~Gamblers reward selected by minimizing validation AUGRC; Appendix~\ref{appendix:cfs_variations} defines every CSF and projection variant; Appendix~\ref{appendix:clip} fixes the CLIP encoder, kernel parameters ($c_0=1, q=3$ following~\citet{binkowski2018demystifying}), and prompt count ($L=80$); Appendix~\ref{appendix:friedman} fully specifies the Friedman/Conover/Bron-Kerbosch pipeline; Appendices~\ref{appendix:mantel} and~\ref{appendix:prediction} document the Mantel ($9{,}999$ permutations, fixed seed) and bootstrap ($n_{\text{boot}}=2000$, fixed seed) protocols; Appendix~\ref{appendix:mean_regret} documents the sign-flip test of the mean ($100{,}000$ draws, fixed seed). The code is released at the public URL given in the introduction.

\item {\bf Open access to data and code}
    \item[] Question: Does the paper provide open access to the data and code, with sufficient instructions to faithfully reproduce the main experimental results, as described in supplemental material?
    \item[] Answer: \answerYes{}
    \item[] Justification: The public code repository given in the introduction (MIT license) contains the analysis pipeline (training, evaluation, Friedman/Conover clique pipeline, NC-based predictor, sensitivity analyses, plotting scripts), with environment setup and step-by-step commands documented in the README. The analyses start from the stored score and NC-metric archives linked in the README. The NC extractor in the repository now computes the corrected NC1 of Appendix~\ref{appendix:neural_collapse}, so a fresh extraction from checkpoints gives those values and not the stored values that the main results use. All four source datasets (CIFAR-10/100, SuperCIFAR-100, TinyImageNet), ImageNet-1k, and the OOD evaluation sets are publicly available and the repository documents how to point dataset paths to local downloads. The VGG-13 and fine-tuned-ViT checkpoints on CIFAR-10, CIFAR-100, and SuperCIFAR-100 are inherited from the FD-Shifts project~\cite{jaeger2022call,traub2024overcoming}; the TinyImageNet checkpoints and the held-out ResNet-18 checkpoints used for cross-architecture transfer (Section~\ref{results:neural_collapse}) are released on Zenodo and linked in the README.

\item {\bf Experimental setting/details}
    \item[] Question: Does the paper specify all the training and test details (e.g., data splits, hyperparameters, how they were chosen, type of optimizer) necessary to understand the results?
    \item[] Answer: \answerYes{}
    \item[] Justification: Section~\ref{results:experimental_setup} specifies the four source datasets, the two main backbones (VGG-13 trained from scratch, ViT fine-tuned from pretrained weights), the three training paradigms, the use of 5 random seeds per (backbone, paradigm) cell, and the AURC and AUGRC ranking metrics. Validation-time choices (early stopping, CSF hyperparameter tuning such as temperature scaling and PCA dimensionality, model selection) are anchored on an ID-only validation split. The full per-cell selected values for dropout and Deep~Gamblers reward (chosen by minimizing validation AUGRC) appear in Appendix~\ref{appendix:hyperparameter_selection}, and the underlying training protocol (optimizer, schedule, augmentations) follows FD-Shifts~\cite{jaeger2022call,traub2024overcoming} as cited in Section~\ref{results:experimental_setup}. Appendices~\ref{appendix:ssl} and~\ref{appendix:imagenet} give the same details for the self-supervised probes and the ImageNet pool.

\item {\bf Experiment statistical significance}
    \item[] Question: Does the paper report error bars suitably and correctly defined or other appropriate information about the statistical significance of the experiments?
    \item[] Answer: \answerYes{}
    \item[] Justification: All quantitative claims are accompanied by statistical evidence. Section~\ref{results:top_cliques} reports Friedman/Conover top cliques at $\alpha = 0.05$ instead of single-winner summaries, with unadjusted pairwise tests and a Holm-adjusted check in Appendix~\ref{appendix:holm}; Section~\ref{results:projection_gains} reports paired Wilcoxon signed-rank tests with per-CSF win/loss counts in the tables of Appendix~\ref{appendix:projection_full}; the cross-architecture transfer of Section~\ref{results:neural_collapse} reports mean set-regret with $95\%$ bootstrap confidence intervals from $n_{\text{boot}} = 2000$ resamples (fixed seed) Holm-corrected paired Wilcoxon tests across the $66$ (regime, side, baseline) comparisons, and a model-clustered sign-flip test of the mean (Appendix~\ref{appendix:mean_regret}); the Mantel test in Appendix~\ref{appendix:mantel} uses $9{,}999$ permutations to assess NC/CSF-rank correlation. Each procedure (bootstrap variability source, permutation policy, Holm correction grouping) is described in full in the corresponding appendix.

\item {\bf Experiments compute resources}
    \item[] Question: For each experiment, does the paper provide sufficient information on the computer resources (type of compute workers, memory, time of execution) needed to reproduce the experiments?
    \item[] Answer: \answerYes{}
    \item[] Justification: Appendix~\ref{appendix:compute} reports the GPU types used for each backbone (NVIDIA T4 for VGG-13 and ResNet-18, NVIDIA A100 for ViT inference, run in parallel; NVIDIA V100 for the ImageNet CLIP regimes) and the CPU-side parallelism (12 workers for CSF inference, projection-filtering, NC, and statistical pipelines). The full checkpoint inventory ($280$ VGG-13, $40$ ViT, and $56$ ResNet-18, totalling $376$ checkpoints) is itemized per source dataset. Training compute attributable to this paper is the held-out ResNet-18 pool ($56$ trainings $\times$ approximately 2.5 hours on a single T4, around $140$ GPU-hours total); the VGG-13 and ViT checkpoints on CIFAR-10, CIFAR-100, and SuperCIFAR-100 come from FD-Shifts~\cite{jaeger2022call,traub2024overcoming} and were not retrained; the TinyImageNet models (60 VGG-13, 10 ViT) were trained with the FD-Shifts code and are not included in this total. Per-checkpoint NC and CSF feature extraction takes approximately 2 minutes (5 minutes on TinyImageNet) and per-CSF inference takes under 1 minute, with 20-minute PCA/KPCA fitting phases on CIFAR-100 and TinyImageNet. CLIP-based OOD aggregation runs once per dataset in roughly 10 minutes. Aggregated over all 376 checkpoints, the inference and analysis pipelines consume on the order of several hundred GPU-hours, with one to two weeks of total wall-clock under the parallelism described above. The self-supervised and ImageNet experiments need no backbone training: a few GPU-hours of feature extraction for the self-supervised probes and about 13 GPU-hours for the ImageNet sweep ($47$ pretrained models), with the predictor and statistical analyses on CPU.
    
\item {\bf Code of ethics}
    \item[] Question: Does the research conducted in the paper conform, in every respect, with the NeurIPS Code of Ethics \url{https://neurips.cc/public/EthicsGuidelines}?
    \item[] Answer: \answerYes{}
    \item[] Justification: The paper uses publicly released image-classification benchmarks (CIFAR-10/100, SuperCIFAR-100, TinyImageNet) and standard OOD evaluation sets (iSUN, LSUN-cropped, LSUN-resize, SVHN, Places365, Textures), with classifier checkpoints inherited from the FD-Shifts project~\cite{jaeger2022call,traub2024overcoming} or trained with its code, and public pretrained encoders and ImageNet classifiers in Appendices~\ref{appendix:ssl} and~\ref{appendix:imagenet}. The work involves no human subjects, no scraped data, no sensitive attributes, and no novel pretrained models. The contribution is a representation-aware analysis of OOD detection methodology, which raises no concerns under the NeurIPS Code of Ethics.

\item {\bf Broader impacts}
    \item[] Question: Does the paper discuss both potential positive societal impacts and negative societal impacts of the work performed?
    \item[] Answer: \answerNA{}
    \item[] Justification: The paper analyzes how the geometry of learned representations relates to the relative performance of OOD detection CSFs. It does not introduce a new model, dataset, or deployable system; the contribution is methodological (a representation-aware benchmark, a rank-based statistical pipeline, and a predictor over Neural Collapse metrics). There is no direct deployment path that would change societal impact relative to the existing OOD-detection literature on which it builds.
    
\item {\bf Safeguards}
    \item[] Question: Does the paper describe safeguards that have been put in place for responsible release of data or models that have a high risk for misuse (e.g., pre-trained language models, image generators, or scraped datasets)?
    \item[] Answer: \answerNA{}
    \item[] Justification: The paper does not release any high-risk artifacts. The released code is an analysis pipeline (CSF inference, statistical tests, NC computation, plotting); the only model checkpoints released with the paper are the TinyImageNet classifiers and the held-out ResNet-18 image classifiers used for the cross-architecture transfer test (Section~\ref{results:neural_collapse}), which carry no misuse risk beyond standard image-classification benchmarks. No generative models, language models, or scraped datasets are released or introduced.

\item {\bf Licenses for existing assets}
    \item[] Question: Are the creators or original owners of assets (e.g., code, data, models), used in the paper, properly credited and are the license and terms of use explicitly mentioned and properly respected?
    \item[] Answer: \answerYes{}
    \item[] Justification: All datasets and models used are publicly released and used under their original terms. The image-classification benchmarks (CIFAR-10, CIFAR-100, SuperCIFAR-100~\cite{krizhevsky2009learning}, TinyImageNet~\cite{le2015tiny}) and the OOD evaluation sets (iSUN~\cite{xu2015turkergaze}, LSUN-cropped/LSUN-resize~\cite{yu2015lsun}, SVHN~\cite{netzer2011reading}, Places365~\cite{zhou2017places}, Textures/DTD~\cite{cimpoi2014describing}) are accessed through the FD-Shifts data pipeline~\cite{jaeger2022call,traub2024overcoming}, which packages each dataset under its original academic-research license. Classifier checkpoints on CIFAR-10, CIFAR-100, and SuperCIFAR-100 come from the FD-Shifts repository (released by its authors), and our pipeline uses a fork of the FD-Shifts code (Apache-2.0). The CLIP image-text encoders used for OOD stratification (Appendix~\ref{appendix:clip}) are released by~\citet{radford2021learning} under the OpenAI CLIP license. The ImageNet classifiers of Appendix~\ref{appendix:imagenet} come from the \texttt{timm} library (Apache-2.0, with the weight terms on each model card), and the DINOv2 encoder of Appendix~\ref{appendix:ssl} comes from its authors' public release. Statistical tools (Friedman, Conover, Bron-Kerbosch, Mantel) are standard published procedures cited at first use.

\item {\bf New assets}
    \item[] Question: Are new assets introduced in the paper well documented and is the documentation provided alongside the assets?
    \item[] Answer: \answerYes{}
    \item[] Justification: Three new artifacts are released with the paper: (i) the analysis pipeline (CSF inference, projection filtering, NC computation, Friedman/Conover/Bron-Kerbosch pipeline, NC-based predictor, plotting scripts) at the public URL given in the introduction (MIT license), with a README documenting environment setup and step-by-step commands; (ii) the held-out ResNet-18 classifier checkpoints used for cross-architecture transfer (Section~\ref{results:neural_collapse}; \url{https://zenodo.org/records/19712370}, CC0); (iii) the TinyImageNet VGG-13 and ViT classifier checkpoints (\url{https://zenodo.org/records/17316185}, CC BY 4.0). Documentation includes the per-cell training-time hyperparameter tables (Appendix~\ref{appendix:hyperparameter_selection}), CSF and projection-variant definitions (Appendix~\ref{appendix:cfs_variations}), the full per-cell checkpoint inventory (Appendix~\ref{appendix:compute}), and the CLIP-aggregation parameters (Appendix~\ref{appendix:clip}).

\item {\bf Crowdsourcing and research with human subjects}
    \item[] Question: For crowdsourcing experiments and research with human subjects, does the paper include the full text of instructions given to participants and screenshots, if applicable, as well as details about compensation (if any)?
    \item[] Answer: \answerNA{}
    \item[] Justification: The paper does not involve crowdsourcing or research with human subjects. All experiments use publicly released image-classification benchmarks and standard OOD evaluation sets (Section~\ref{results:experimental_setup}).

\item {\bf Institutional review board (IRB) approvals or equivalent for research with human subjects}
    \item[] Question: Does the paper describe potential risks incurred by study participants, whether such risks were disclosed to the subjects, and whether Institutional Review Board (IRB) approvals (or an equivalent approval/review based on the requirements of your country or institution) were obtained?
    \item[] Answer: \answerNA{}
    \item[] Justification: The paper does not involve crowdsourcing or research with human subjects, so IRB approval (or equivalent) is not applicable.

\item {\bf Declaration of LLM usage}
    \item[] Question: Does the paper describe the usage of LLMs if it is an important, original, or non-standard component of the core methods in this research? Note that if the LLM is used only for writing, editing, or formatting purposes and does \emph{not} impact the core methodology, scientific rigor, or originality of the research, declaration is not required.
    \item[] Answer: \answerNA{}
    \item[] Justification: Large language models are not part of the core methodology of this paper. The CLIP image-text encoder~\cite{radford2021learning} is used purely as a fixed embedding extractor for OOD-set proximity (Appendix~\ref{appendix:clip}); no language generation, prompting, or LLM-driven reasoning is involved in the analysis. LLMs were not used as a core component of the research.

\end{enumerate}

\end{document}